\documentclass{article}
\usepackage{arxiv}

\usepackage{xcolor}
\usepackage[hidelinks]{hyperref}
\usepackage{graphicx}
\usepackage{amsmath}
\usepackage{amssymb}
\usepackage{bm}
\usepackage{array}
\usepackage{caption}
\usepackage{url}
\usepackage[ruled, lined, linesnumbered, commentsnumbered, longend]{algorithm2e}
\SetKwComment{Comment}{$\triangleright$\ }{}

\usepackage{multirow}
\usepackage{hyperref}
\usepackage{subcaption}
\usepackage{latexsym}

\title{DPA-WNO: A gray box model for a class of stochastic mechanics problem}
\author{ \hspace{1mm}Tushar\\
	Department of Applied Mechanics\\
	Indian Institute of Technology (IIT) Delhi\\
	Hauz Khas - 110 016, New Delhi, India \\
	\texttt{amz218314.iitd@gmail.com} \\
	\And
	\hspace{1mm}Souvik~Chakraborty \\
	Department of Applied Mechanics\\
	Yardi School of Artificial Intelligence (joint appointment)\\
	Indian Institute of Technology (IIT) Delhi\\
	Hauz Khas - 110 016, New Delhi, India \\
	\texttt{souvik@am.iitd.ac.in} \\
}
\begin{document}
\maketitle

\begin{abstract}
The well-known governing physics in science and engineering is often based on certain assumptions and approximations. Therefore, analyses and designs carried out based on these equations are also approximate. The emergence of data-driven models has, to a certain degree, addressed this challenge; however, the purely data-driven models often (a) lack interpretability, (b) are data-hungry, and (c) do not generalize beyond the training window. Operator learning has recently been proposed as a potential alternative to address the aforementioned challenges; however, the challenges are still persistent. We here argue that one of the possible solutions resides in data-physics fusion, where the data-driven model is used to correct/identify the missing physics. To that end, we propose a novel Differentiable Physics Augmented Wavelet Neural Operator (DPA-WNO). The proposed DPA-WNO blends a differentiable physics solver with the Wavelet Neural Operator (WNO), where the role of WNO is to model the missing physics. This empowers the proposed framework to exploit the capability of WNO to learn from data while retaining the interpretability and generalizability associated with physics-based solvers. We illustrate the applicability of the proposed approach in solving time-dependent uncertainty quantification problems due to randomness in the initial condition. Four benchmark uncertainty quantification and reliability analysis examples from various fields of science and engineering are solved using the proposed approach. The results presented illustrate interesting features of the proposed approach.
\end{abstract}



\section{Introduction}
Physical systems are driven by the laws of physics, often represented as Partial Differential Equations (PDEs) \cite{sommerfeld1949partial, debnath2005nonlinear, jones2009differential}. The literature on PDEs is also quite mature with techniques such as Finite Element Methods (FEM) \cite{kang1996finite, cottrell2009isogeometric, yin2022interfacing}, Finite Difference Methods (FDM) \cite{ozicsik2017finite}, and Finite Volume Methods (FVM) \cite{eymard2000finite} readily available at our fingertip. However, the well-known physical laws are often based on certain assumptions and approximations \cite{jones2019investigation, swartz1985concept}. For example, the effect of the environment is often neglected {\cite{leighton1965feynman,auyang1998foundations}}, the geometry of structures is often simplified by ignoring joints \cite{di2023nonlinear}, and nonlinearities in systems are sometimes neglected \cite{rocha2014undesirable}. Henceforth, even the most accurate solver only provides an approximate solution. Additionally, dynamical systems exhibit altered behaviour over time due to the development of defects like cracks and deformities, and it is necessary to track these changes so as to ensure the safety of the system. One potential alternative to physics-based models is to rely on experimental data; however, experiments are expensive and time-consuming and hence, we can only afford a limited number of experiments (usually in the order of tens). Generally, so few experiments are often not adequate, specifically when dealing with problems involving uncertainty quantification {\cite{ kobayashi2023uncertainty, soize2017uncertainty, gres2022uncertainty, hu2023simplified, psaros2023uncertainty}, optimization \cite{chakraborty2017surrogate,gill2019practical, wang2022time}, and reliability analysis \cite{goswami2016reliability, roy2023support, ding2023real, afshari2022machine} }. Therefore, it is necessary to conduct research and develop methods and techniques that can exploit approximate physics and high-fidelity data to build accurate predictive models.


Data-driven approaches for solving PDEs have been proposed as a potential alternative to numerical methods discussed before \cite{karniadakis2021physics}. Physics-Informed Neural Network (PINN) \cite{cai2021physics,chakraborty2021transfer,raissi2019physics, rezaei2022mixed} is perhaps the most popular framework in this category. In PINN, the solution field is represented using a neural network and the inductive bias, in the form of residual loss, is introduced into the loss function. PINN provides the flexibility to train the model by using both physics and data; however, the data and the physics should have similar fidelity; unfortunately, this setup does not help as the physics is often low-fidelity owing to the assumptions and approximations inherent to it. To address this issue, multi-fidelity PINN \cite{goswami2020transfer} has been proposed where transfer learning is used to map from one fidelity level to another fidelity level. A number of extensions of this idea can also be found in the literature \cite{tripura2023physics, navaneeth2023stochastic}. Although this approach yields satisfactory results in the vicinity of the training regime, the framework lacks out-of-distribution generalization and extrapolation capabilities; this is because this method is extrusive in nature and does not attempt in-equation correction. 

Some work on in-line augmentation of governing physics by using machine learning can also be found in the literature. One of the first works along this line can be found in {\cite{garg2022physics}, where duel Kalman filter (DKF) \cite{dertimanis2019input} and Gaussian process \cite{mackay1998introduction} were used for in-equation augmentation. A two-level strategy was adopted; in the first level, the residual and the state variables were estimated by using DKF and in the second step, GP was used to learn the mapping between the estimated state variables and residual, and the same was used to augment the known-physics. This method yields accurate results for discrete systems (i.e., systems governed by ordinary differential equations). As an improvement over these approaches, physics-integrated deep learning approaches were proposed by researchers \cite{takeishi2021physics}. Following a similar spirit as \cite{garg2022physics}, the known physics was augmented with a neural network and an end-of-end training algorithm was used to compute the neural network parameters. The idea was later extended in \cite{tripathi2023physics} for reliability analysis of dynamical systems. Other work along this line includes \cite{li2021data}. However, these works are also applicable to discrete dynamical systems only. The idea was recently extended by the authors for stochastic dynamical systems governed by Gaussian white noise \cite{chakraborty2023deep}.


Operator learning \cite{li2018sequential, kovachki2021neural} as a potential alternative to numerical solvers has recently surfaced for solving parametric PDEs.  This approach involves employing neural operators, that learns the solutions to parametric Partial Differential Equations (PDEs) through functional mapping in infinite-dimensional solution spaces. The DeepONet \cite{lu2021learning} is one example of a neural operator that utilizes a neural network-based framework to learn the functional mapping between input and output datasets. It consists of two networks, the branch net and the trunk net, which enable the mapping between the input function and the corresponding output at a sensor point. Another approach, known as Graph Neural Operators (GNO) \cite{li2020neural}, focuses on learning the kernel of integral transforms through a message-passing interface between graph networks. While GNO provides a different perspective on operator learning, the stability of its architecture decreases with an increase in hidden layers. In recent work, the Fourier Neural Operator (FNO) \cite{li2020fourier,wen2022u} has been introduced as another operator learning framework. FNO learns the parameters of the integral kernel in Fourier space using spectral decomposition. It shows promising performance compared to other state-of-the-art methods for operator learning. However, FNO lacks spatial resolution information due to the frequency localization of the basis functions used in the Fast Fourier Transform (FFT). As a result, FNO may perform poorly in scenarios involving complex geometric shapes or when learning the spatial behaviour of signals. To address this, the Wavelet Neural Operator (WNO) \cite{tripura2023wavelet, thakur2022multi} was introduced. WNO leverages wavelet transformation \cite{zhang2017time,boggess2015first} to capture the spatial and frequency localization and learn the variations in input patterns over spatial coordinates. To address the pure data-driven nature of operators, physics-informed operators \cite{tripura2023physics} can also be found in the literature. Operator learning provides a massive boost in accuracy as compared to conventional neural network-based approaches; however, the challenges associated with data-driven and physics-informed approaches hold true for operator learning as well.

Based on the discussion above, the following conclusions can be drawn:
\begin{itemize}
    \item The operator learning algorithms provide a significant improvement over other machine learning models. However, operator learning algorithms are (a) data-hungry and (b) not interpretable. More importantly, it cannot handle the common scenario where the physics available is approximate (low-fidelity) and the (high-fidelity) data available is sparse.
    \item \textit{Data-physics fusion}, a.k.a. gray-box model, has emerged as a potential option for handling approximate physics (low-fidelity) and sparse (high-fidelity) data. However, this domain is still in its infancy. 
\end{itemize}

The objective of this paper is to develop a novel framework that can fuse the strength of operator learning with the grey-box model. To that end, we propose the differentiable physics-augmented neural operator, a framework that exploits the strengths of both operator learning and the grey-box model. Although the proposed framework can be used with any operator learning algorithm, we here advocate the use of WNO \cite{tripura2023wavelet} because of its already proven performance. The resulting framework referred to as Differentiable Physics Augmented WNO (DPA-WNO) has the following salient features:

\begin{itemize}
    \item \textbf{Physics-WNO fusion: } The proposed framework is a novel framework that augments the WNO with a low-fidelity physics model within the form of governing equation itself. This allows better generalization as compared to an extrusive augmentation \cite{chakraborty2021transfer}.
    \item \textbf{End-to-end training: }The proposed approach is trained in an end-to-end fashion and hence, the training phase is computationally efficient.
    \item \textbf{Differentiable physics}: The proposed approach augments differentiable physics \cite{um2020solver} solver with wavelet neural operator. The differentiable physics solver allows training of the model by using the back-propagation algorithm; in other words, no numerical approximation is involved in training the network.
\end{itemize}
The efficacy of the proposed approach is illustrated in solving a number of benchmark uncertainty quantification and reliability analysis problems.

The subsequent sections of the paper are organized as follows: Sec. \ref{sec:ps} presents a formal description of the problem statement. Sec. \ref{sec:pawno} provides an overview of the proposed approach, including a basic description of WNO and the algorithm. Sec. \ref{sec:ns} showcases the results obtained from the numerical examples considered. Finally, Sec. \ref{sec:concl} concludes the paper with closing remarks.
\section{Some preliminary definitions}\label{sec:def}
%
We begin by defining certain definitions that are essential for the discussion ahead.

\noindent \textit{Definition 1: Stochastic field} 

\noindent Consider $\left(\bm \Theta, \mathcal F, \mathcal P \right)$ to be the probability space and 
$\mathcal L_2\left(\bm \Theta, \mathcal F, \mathcal P \right)$ to be the Hilbert space of random variables having a finite 
second-order moment. A \textit{stochastic field} $\mathcal H \left(\bm x, \bm \theta \right), \bm x \in \mathbb R^N, \bm \theta \in \bm \Theta$
is a curve in  $\mathcal L_2\left(\bm \Theta, \mathcal F, \mathcal P \right)$. Therefore, $\mathcal H \left(\bm x, \bm \theta_0 \right)$ is a realization of the \textit{stochastic field} and $\mathcal H \left(\bm x_0, \bm \theta \right)$ is a random variable.
\\
\\
\noindent \textit{Definition 2: Limit-state function and failure domain} 

\noindent Consider $\mathcal H (\bm x, \bm \theta)$ to be a stochastic field and $\mathcal O: \mathcal H(\bm x, \bm \theta^i) \mapsto \mathcal J(\bm x, \bm \theta^i)$, to be an operator that maps a realization of the stochastic field, $\mathcal H (\bm x, \bm \theta^i)$ to realization of another stochastic field, $\mathcal J(\bm x, \bm \theta^i)$. The function  $\mathcal J(\bm x, \bm \theta^i)=0$ is referred to as the \textit{limit-state function} if $\min_{\bm x^j} \mathcal J(\bm x^j, \bm \theta^i) < 0, \forall i $ denotes the \textit{failure domain} $\Omega_f$ and $\min_{\bm x^j} \mathcal J(\bm x^j, \bm \theta^i) \ge 0, \forall i $ is the safe domain. Mathematically, we define the \textit{failure domain} as collections of functions, $\mathcal H (\bm x, \bm \theta^i, \forall i)$ such that 
\begin{equation}
    \Omega_f \triangleq \left\{\mathcal H (\bm x, \bm \theta^i):  \min_{\bm x^j} \mathcal J(\bm x^j, \bm \theta^i) < 0\right\}
\end{equation}
\\
\\
\noindent \textit{Definition 3: Probability of failure and reliability index} 

\noindent Considering $\Omega_f$ to be the failure domain, the probability of failure is defined as
\begin{equation}
    P_f = \int_{\Omega_f}{dF_{\bm \Theta}(\bm \theta)} = \int_{\Omega}{\mathbb I_{\Omega_f}(\bm \theta)dF_{\bm \Theta}(\bm \theta)},
\end{equation}
where $dF_{\bm \Theta}(\bm \theta)$ represents the cumulative density function of $\bm \theta$ and $\mathbb I_{\Omega_f}$ represents the indicator function such that
\begin{equation}
    \mathbb I_{\Omega_f}(\bm \theta ) = \left\{ \begin{array}{ll}
      1,   & \text{if }\mathcal H (\bm x, \bm \theta) \in \Omega_f  \\
      0,   & \text{elsewhere}
      \end{array} \right. 
\end{equation}

\section{Problem statement}\label{sec:ps}
Consider $\mathcal D = \left[ \mathcal H (\bm x; \bm \theta^i), \bm u (\bm x, t; \bm \theta^i) \right]_{i=1}^{N_s}$ represents data available from laboratory experiments. $\mathcal H (\bm x; \bm \theta^i)$ here represents the input stochastic field and $\bm u (\bm x, t; \bm \theta^i)$ represents the response variable. Using the response variable, the limit-state function is generally represented as
\begin{equation}\label{eq:ls}
    \mathcal J(\bm x, t; \bm \theta) := g_t - g\left(\bm u (\bm x, t; \bm \theta)\right), 
\end{equation}
where $g_t$ represents the threshold. As an example, the input stochastic field can be elastic modulus, Poisson's ratio, boundary condition, forcing function, and initial condition. In this paper, we limit our discussion to uncertainty in the initial condition only. Similarly, the measured response can be displacement field, strain field, and stress field. Here, we consider that $\bm u (\bm x, t; \bm \theta)$ represents the maximum response. We also consider that the limit state function is represented in terms of maximum response over the field and hence,
\begin{equation}\label{eq:ls1}
    g\left(\bm u (\bm x, t; \bm \theta)\right) = \max_{\bm x} \bm u (\bm x, t; \bm \theta)
\end{equation}
With this setup, the limit-state function in Eq. \eqref{eq:ls} is no longer a function of space $\bm x$. To emulate a realistic setup, we impose a constraint on the number of experiments $N_s$ that one can perform in the lab. Therefore, directly using the data to perform uncertainty quantification and reliability analysis is not an option.

Consider that the governing PDE of a system along with Dirichlet boundary and initial conditions are represented as:
\begin{subequations}\label{eq:gov_eq}
\begin{equation}\label{eq:ge}
\frac{\partial \bm{u}}{\partial t} = \mathcal{G}_k\left(\bm{u},\bm{u_x},\bm{u_{xx}}\ldots, \bm{p}\right),\;\; x\in D,\; t\in[0,T],
\end{equation}
\begin{equation}\label{eq:d_bc}
    \bm u(\bm x_b, t) = u_b,\;\;\bm x_b \in dD, t\in[0,T],
\end{equation}
\begin{equation}\label{eeq:ic}
    \bm u(\bm x, t=0) = \mathcal H (\bm x; \bm \theta),\;\; \bm x \in D, \bm \theta \in \bm \Theta.
\end{equation}
\end{subequations}
$\mathcal{G}_k$ represents some general operator representing the PDE and $\bm{p}$ represents the parameters of the PDE. As stated earlier, the uncertainty is in the initial condition (See Eq. \eqref{eeq:ic}) and is parameterized by $\bm \theta$. We note that it is possible to solve the uncertainty quantification and reliability analysis problems purely based on the governing equation in Eq. \eqref{eq:gov_eq}. However, as is often the case, we consider that Eq. \eqref{eq:gov_eq} is derived based on certain assumptions and approximations. Therefore, reliability analysis and uncertainty quantification performed based on Eq. \eqref{eq:gov_eq} will also be approximate. 
With this setup, the objective of this paper is two-fold:
\begin{itemize}
    \item First, we want to develop an end-to-end framework that can utilize the sparse data $D$ to correct the governing physics in Eq. \eqref{eq:gov_eq}. We are particularly interested in in-equation correction as that is expected to yield superior results.
    \item The second objective is to illustrate that the developed model can be used for solving uncertainty quantification and reliability analysis problems. The scope of this paper is limited to uncertainty quantification and reliability analysis due to stochastic initial conditions. Stochasticity in other parameters is not considered in this paper.
\end{itemize}

\section{Proposed approach}\label{sec:pawno}
In this section, we describe the basic principles of the proposed model, along with the architecture design of the model, the training process and the representative diagram for the overall framework. However, before proceeding with the overall algorithm, we briefly discuss the Wavelet Neural Operator (WNO), which is a key component in the algorithm.

\subsection{Wavelet Neural operator (WNO)}
\begin{figure}[t]
    \centering
\includegraphics[width=1\textwidth]{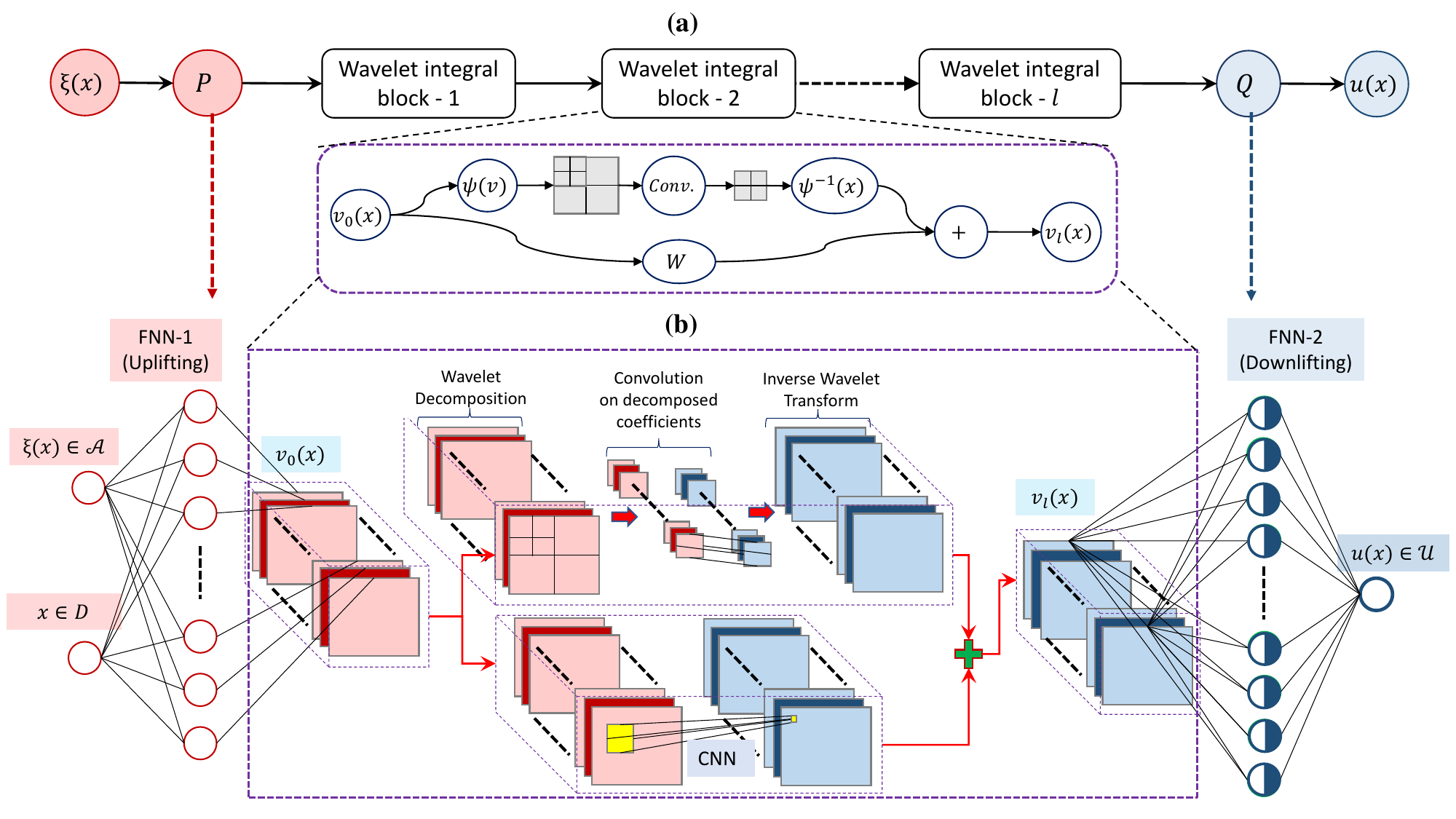}
    \caption{Wavelet Neural Operator(WNO) architecture: Panel (a) of the figure presents a schematic representation of WNO. The inputs undergo a local transformation (P(·)) to be lifted to a higher dimension, followed by processing through multiple wavelet kernel integral layers. The final output is transformed back to the original dimension using another local transformation (Q(·)) and activated to produce the solution u(x). The wavelet kernel integral layer performs multilevel wavelet decomposition, convolution with neural network weights, and inverse wavelet transformation. In panel (b), a simplified version of the WNO with a single wavelet kernel integral layer is depicted. It includes initial parameters and spatial information as inputs, shallow fully connected neural networks for local transformations (P(·) and Q(·)), and separate branches for wavelet decomposition and convolutional neural network processing. The outputs are summed, activated, and transformed to obtain the target solution u(x). The approach allows for the construction of WNOs with any number of wavelet integral layers}
    \label{fig:WNO}
\end{figure}

The Wavelet Neural Operator (WNO) is a neural operator that learns the integral operator for parametric partial differential equations (PDEs) by combining an integral kernel with wavelet transformation. WNO takes advantage of wavelets' ability to localize functions in both time/space and frequency domains, allowing for precise pattern tracking and accurate learning of functional mappings. This localization property enables WNO to provide high spatial and frequency resolution. Here, we briefly discuss the fundamental working principal of the WNO.

Consider a $d$ dimensional smooth domain $D$ in the Banach functional space. We establish an input-output pair ${\xi(x) \in \mathcal{A}, u(x) \in \mathcal{U}}$, where $u(x)$ is obtained by applying the solution operator $\mathcal{D}$ to the input parameter field $\xi(x)$, as $u(x) = \mathcal{D}(\xi(x))$. For the sake of simplicity, let us consider a straightforward example with a constant $\xi$ as illustrated below.

\begin{equation}
\begin{aligned}
& \frac{{\partial^2 u(x)}}{{\partial x^2}} + \xi u(x) = f(x, u(x)); \quad x \in D\\
& u(x) = 0; \quad x \in \partial D
\end{aligned}
\end{equation}
where the function $f: D \times C \rightarrow C$, is continuous. The aforementioned differential equation can be transformed into an integral equation represented by the following form,
\begin{equation}
    u(x) = \int_{D} k(x, \psi)f(\psi, u(\psi)) d\psi; \quad x \in D
\end{equation}
where $k(.)$ is the Green's function. The above equation can be generalised to Urysohn-type integral equation, in case of non linear PDEs, represented as:
\begin{equation}\label{eq:Urysohn}
    u(x) = \int_{D} k(x, \psi)f(\psi, u(\psi)) d\psi + g(x); \quad x \in D 
\end{equation}
where $g(x)$ is some linear function and k(.) in this equation is the kernel function, which represents the non linear counterpart of the Green's function. WNO learns this nonlinear solution operator $\mathcal{D}$: $(\xi(x))\mapsto u(x)$ mapping the input function space $\xi(x)\in \mathcal{A}$ to the output solution space $u(x) \in \mathcal{U}$. To train such a model, it requires $N$ samples of the input-output $\left\{\xi_j, u_j(x)\right\}^{N}_{j=1}$ pair.

Given that the integral in equation \ref{eq:Urysohn} cannot be defined in an infinite-dimensional space, we need to obtain a finite-dimensional parameterization space by discretizing the solution domain $D \in \mathbb{R}^n$. To enable a multi-dimensional kernel convolution, the input $\xi(x)$ is transformed into a higher-dimensional space, using a local transformation denoted as P: $\xi(x) \mapsto v_0(x)$. This local transformation can be represented by a shallow fully connected neural network (FNN) or a 1-D convolution. In this lifted space, a series of iterations, described by Eq. \ref{eq:Urysohn}, are performed. These iterations involve the transformation $\mathcal{G}: v_{j+1}\mapsto \mathcal{G}(v_j)$ and resemble the concept of hidden layers in neural networks, but they follow the principles of operator theory in functional analysis \cite{hutson2005applications}. After $l$ iterations, a second local transformation denoted as $Q: v_l(x) \mapsto u(x)$ is applied to obtain the final solution space. Inspired by equation \ref{eq:Urysohn}, the step-wise updates G(·) are defined as follows:
\begin{equation}
  G(v_j)(x) := \phi((K(\xi; \varphi) * v_j)(x) + \mathcal{W} v_j(x)); \quad x \in D, \quad j \in [1, l]  
\end{equation}
where $\phi(.)$ represents a non-linear activation function, $\varphi \in \theta_{NN}$ denotes the kernel parameters, $W:$ represents a linear transformation, and K is the nonlinear integral operator which is defined as follows:
\begin{equation}\label{eq:wvkernel}
    (K(\xi; \varphi) * v_j)(x) := \int_{D} k(\xi(x), x, \psi; \varphi) v_j(\psi)d\psi; \quad x \in D, \quad j \in [1, l]
\end{equation}
The main idea in WNO is to parameterize the neural network in the wavelet domain and learn this kernel $k(\xi(x), x, \psi; \phi)$. To enable this parameterization, the lifted input $v_j(x)$ undergoes a wavelet transform. The forward and inverse wavelet transforms, denoted as $\mathcal{W}(·)$ and $\mathcal{W}^{-1}(·)$, respectively, are defined as follows \cite{daubechies1992ten}:
\begin{equation}
\begin{aligned}
    (\mathcal{W}v_j)(s, \tau) &= \int_{D} \frac{\Gamma(x)}{|s|^{1/2}} \zeta\left(\frac{x - \tau}{s}\right) dx \\
    (\mathcal{W}^{-1}(v_j)_w)(x) &= \frac{1}{C_{\zeta}} \int\limits_{0}^{\infty} \int_{D} (v_j)_w(s, \tau) \frac{1}{|s|^{1/2}} \tilde{\zeta}\left(\frac{x - \tau}{s}\right) d\tau \frac{ds}{s^2}
\end{aligned}
\end{equation}
where $\Gamma(x)$ represents the orthonormal mother wavelet. The scaling and translational parameters for wavelet decomposition are denoted as $s$ and $\tau$, respectively. The wavelet decomposed coefficients of $v_j(x)$ are indicated as $(v_j)_w$. The function $\zeta(·)$ corresponds to the scaled and shifted mother wavelet. Additionally, $0< C_{\zeta}< \infty$ represents the admissible constant \cite{daubechies1992ten}.

To achieve the objective of learning the kernel integration in the wavelet domain, the kernel $k$ is explicitly defined in the wavelet space, referred to as $R_{\varphi} = \mathcal{W}(k)$. Utilizing the convolution theorem, the integral in Eq. \ref{eq:wvkernel} over the wavelet domain can be expressed as follows:

\begin{equation}
(K(\varphi) * v_j)(x) = \mathcal{W}^{-1}(R_{\varphi} \cdot \mathcal{W}(v_j))(x); \quad x \in D
\end{equation}
A schematic representation of WNO is shown in Fig.\ref{fig:WNO}.

\subsection{Physics augmented WNO}
WNO discussed in the previous section is designed to learn the solutions operator from data. However, in a realistic setting, we generally have access to both physics (low fidelity) and data (sparse). In this section, we propose a variant of WNO that allows integration of both physics and data. The fundamental idea is to augment WNO to the low fidelity physics, which can be shown as:
\begin{equation}\label{aug_eqn}
\frac{\partial \mathbf{u}}{\partial t} = \mathcal{G}_{k}(\mathbf{u},\mathbf{u_x},\mathbf{u_{xx}}…,\mathbf{p}_k) + \mathcal{W}(\mathbf{u};\bm{\theta});
\end{equation}
where,  $\mathcal{W}(\mathbf{u};\bm{\theta})$ represents the WNO. The hypothesis here is that WNO will learn the missing physics. Since we augment the WNO with the low fidelity physics we refer to the proposed approach as the physics augmented WNO. For using PA-WNO in practice, one needs to train the WNO in Eq.\eqref{aug_eqn}. However, we don't have access to the samples/measurements of missing physics. Two different approaches can be adopted for training $\mathcal{W}(\mathbf{u};\bm{\theta})$; (a) Sequential approach and (b) end-to-end approach. Sequential approaches are generally expensive as it involves a two stage process; estimating the residual by employing filtering and then training the model \cite{garg2022physics}. Therefore a better alternative is to employ an end-to-end approach. However, this will involve integration of differentiable physics solver with WNO. We here present differential physics augmented WNO (DPA-WNO). A schematic  representation of DPA-WNO is a=shown in Fig.\ref{fig:overall}  

\subsubsection{Training}\label{subsubsec:training}
Our objective is to train our model such that the Wavelet Neural Operator (WNO), is able to learn the missing part in the known physics, through end to end training. As mentioned earlier, we possess high-fidelity solution data for the problem at hand. In real-world scenarios, this data may correspond to accurately recorded sensor measurements. However, for the purpose of this work, we generated synthetic data by solving the complete physics partial differential equation (PDE). The loss function used to train the model involves calculating the mean square error between the ground truth and the solution obtained by numerically solving the augmented equation (Eq.\eqref{aug_eqn}). To ensure propagation of information through the solver, we developed a differentiable physics based finite difference solver. This allows us to compute the gradients by using the automatic differentiation functionality. During the training process, the model is progressively exposed to the system and acquires knowledge about its dynamics by gradually unrolling the number of predicted time-steps as the training advances. At the initial stages of training, the model predicts only a limited number of time-steps ($10$ in most cases) starting from its initial state. We update the parameters for $N$ epochs ($N =100$ in most cases). However, as the training progresses, this prediction horizon extends up to $50$ time-steps and a total of $500$ epochs. This progressive expansion enables the model to uncover and grasp the underlying dynamics that may not be evident from the provided initial states alone. Furthermore, since the model's output prediction becomes the input for the subsequent time-step, we can back propagate the gradients through multiple time-steps, without the latent variable recurrent connections \cite{geneva2020modeling}.
\begin{figure}[t]
    \centering
\includegraphics[width=1\textwidth]{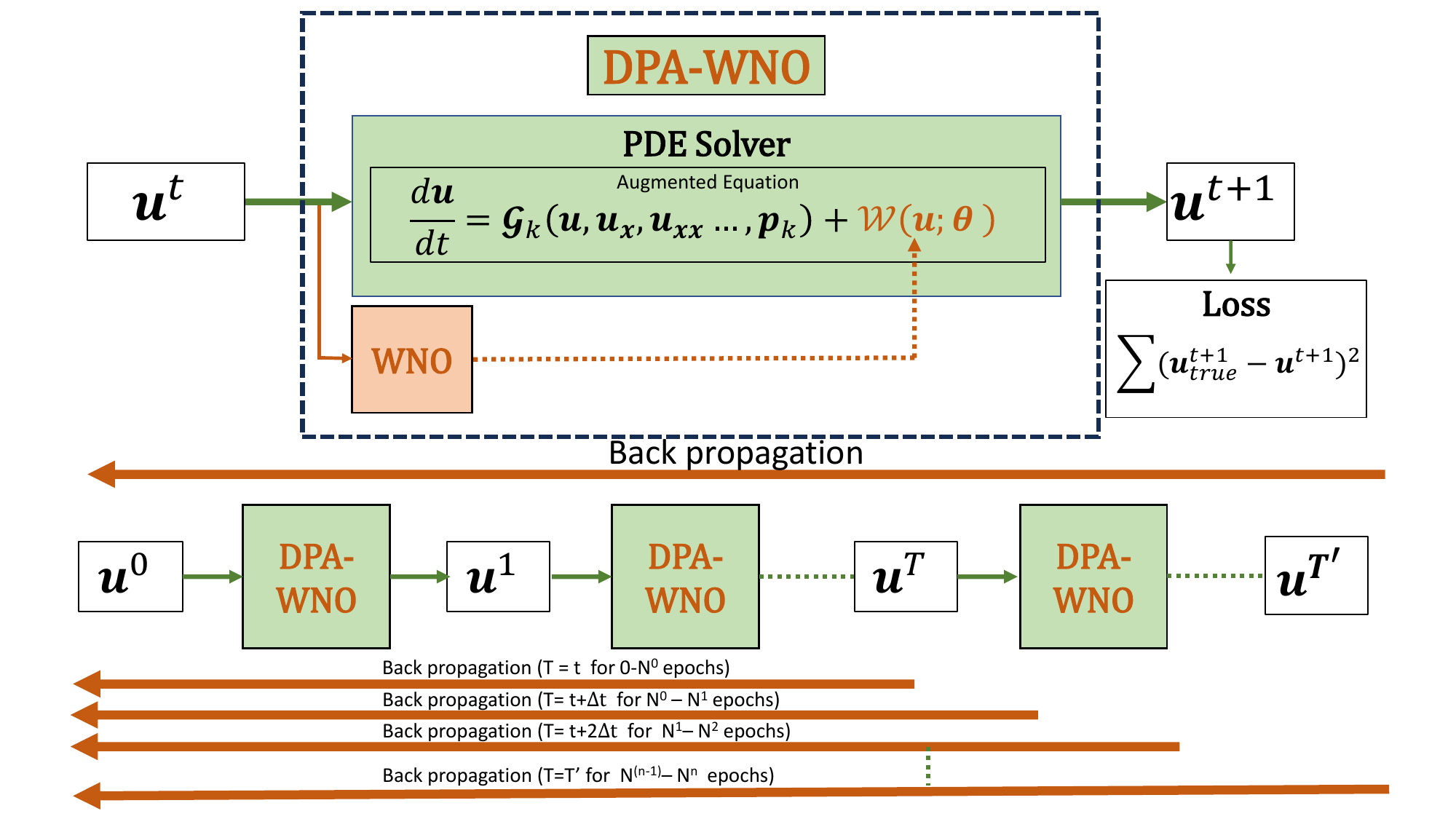}
    \caption{Schematic representation of DPA-WNO}
    \label{fig:overall}
\end{figure}

\subsubsection{Algorithm}
Having discussed the network architecture and the loss function, we proceed with discussing the proposed algorithm. The code accompanying the proposed approach is developed using \texttt{PyTorch} and Adam optimizer is used to train the model. Details on the algorithm are discussed in Algorithm \ref{alg:framework}.
\begin{algorithm}[h]
\caption{Algorithm for the 1-D model}\label{alg:framework}
\SetKwInOut{KwIn}{Input} 
\SetKwInOut{KwOut}{Output}
\KwIn{Initial conditions: $\bm{u}_{0,1:N_x}^{(i)}, $training data: $\mathcal{D} = \left\{\bm{u}_{0,1:N_x}^{(i)}, \bm{u}^{(i)}_{1:N_t,1:N_x} \right\}_{i=1}^N$, learning rate: $\eta$, time-step: $\Delta t$, grid points spacing $\Delta x$ }
Initialize network parameters $\bm \theta$.\\
\For{e = 1 to number of epochs}{
     \For{b = 1 to number of batches}{
          {Draw a mini-batch $\mathcal{A} = \left\{\bm{u}_{0,1:N_x}^{(i)}, \bm{u}^{(i)}_{1:N_t,1:N_x} \right\}_{i=1}^n$ from $\mathcal D$}\\
          {T $\longleftarrow$ Time steps to unroll[epochs] \Comment*[r]{\ref{subsubsec:training}}}
          \For{t = $0$ to $T$}{ 
            {$\left[\mathcal{G}_{uk}(\bm{u}(t),\bm{u_x}(t),\bm{u_{xx}}(t)…,\bm{p}_{uk})\right] \longleftarrow \mathcal{W}(\bm u(t); \bm \theta)$}\\
            {$\bm u_{pred}(t+1) = \bm u_{pred}(t) +
            \Delta t\cdot\left\{ \mathcal{G}_{k}(\bm{u}(t),\bm{u_x}(t),\bm{u_{xx}}(t)…,\bm{p}_k)  + \mathcal{W} (\bm u(t); \bm \theta)\right\}$ \Comment*[r]{Differentiable Physics}}
            {Compute the training loss, $\widehat{\mathcal{L}}^{(t+1)}_{\mathrm{MSE}}$: $\sum_{i=1}^{n}(\bm u^{(i)}_{pred; 1:N_x}(t+1) - \bm u^{(i)}_{true; 1:N_x}(t+1))^2$}}
      {Total loss per batch = $\sum_{t=0}^{T}\widehat{\mathcal{L}}^{(total)}_{\mathrm{MSE}}$}\\
      {Calculate gradient $\frac{\partial{\widehat{\mathcal{L}}}_{\mathrm{MSE}}^{2}}{\partial \bm \theta}$ and update $\bm \theta$ using gradient based Adam optimizer}
    }}
\KwOut{Learnt the parameters $\bm \theta$}
\end{algorithm}

\section{Numerical Examples}\label{sec:ns}
In this section, we evaluate the performance of the proposed framework with the help of four benchmark problems from literature. For illustrative purpose, we have synthetically created a low fidelity physics model by omitting some portion of the PDE. For each example problem, two such cases have been created by omitting different parts of the equation, in order to evaluate the performance of the proposed framework in learning variety of situations and complexities. We compare the solutions obtained using the proposed framework with the ground truth. In particular, we illustrate the extrapolation and generalisation capabilitiies of the proposed approach. To provide a comparative and comprehensive evaluation of the framework in solving uncertainty quantification problems, probability density function (PDF) plots of response are presented corresponding to uncertainty in the initial conditions. These plots compare the performance of the proposed framework DPA-WNO with purely data driven WNO and known physics models. The comparison is done with reference to the ground truth. Lastly, we illustrate the capability of the proposed approach in solving the time-dependent reliability problem.
For quantitative assessment, we employed two error metrics to evaluate the accuracy of predictions. The first metric utilized is the mean square error, while the second metric is the Hellinger distance. The Hellinger distance, denoted as $H(P(u(x,t), Q(u(x,t))$, is defined based on two probability density functions, $P(u(x,t)$ and $Q(u(x,t)$, and can be expressed mathematically as follows:

\begin{equation}\label{eq:hellinger}
H(P(u(x,t), Q(u(x,t)) = \frac{1}{\sqrt{2}} \left|\sqrt{P(u(x,t))}-\sqrt{Q(u(x,t))}\right|_2
\end{equation}
here, $u=u(x,t)$ represents the variable of interest, and the Hellinger distance calculates the dissimilarity between the square roots of $P(u)$ and $Q(u)$ using the Euclidean norm. By utilizing these error metrics, we aimed to quantitatively measure the predictive error. 

Additionally, reliability analysis is conducted by sampling 5000 initial conditions from a zero mean Gaussian random field with the Exponential Sine Squared kernel which can be represented as follows:
\begin{equation}\label{eq:kernel}
 k(x, x') = \alpha \exp \left( -\frac{2}{l^2} \sin^2 \left( \frac{\pi}{p} \|x - x'\| \right) \right)
\end{equation}
where, $k(x, x')$ represents the kernel function f, $\alpha$ represents the process variance; $l$ is a parameter representing the length scale; $p$ is a parameter representing the periodicity of the kernel and $\|x - x'\|$ represents the Euclidean distance between $x$ and $x'$. In all the examples, the limit state function is expressed followings Eqs. \eqref{eq:ls} and \eqref{eq:ls1}. 


For all the subsequent numerical examples, the chosen architecture for the Wavelet Neural Operator (WNO) remains consistent, including the case of purely data-driven WNO. The architecture consists of four layers of WNO blocks, each activated using the GeLU activation function \cite{tripura2023wavelet}. A uniform approach is followed across all examples, employing four levels of wavelet decomposition. The specific wavelet utilized is the $db6$ wavelet, belonging to the Daubechies family \cite{daubechies1992ten, meyer1993wavelets}. The input comprises two nodes, one for $\bm{u(x,t)}$ and the other for the grid $\bm{x}$, with net dimension $(N\times N_x\times 2)$. This input is initially processed by a fully connected layer $fc_0$ with a dimension of 64. Subsequently, it passes through four layers of CNNs, each with a width of 64. The output is then downlifted by fully connected layer $fc_1$ with a dimension of 128, followed by $fc_2$ with a dimension of 1, with final output dimension $(N\times N_x\times 1)$.

After integrating the WNO framework with the partially known physics model (PDE), the augmented PDE is solved using the differentiable physics-based Finite Difference Method (FDM). Across all examples, the model is trained for a total of 500 epochs. Additional specific details pertaining to each example are provided accordingly.

\subsection{Example 1: Burgers' equation}\label{subsec:eg1}

In this first example, we consider the well-known Burgers' equation, which is frequently employed to mathematically represent various phenomena such as the formation of waves, turbulence, fluid mechanics, gas dynamics, traffic flow, and more \cite{kutluay1999numerical}. The considered equation is a time-dependent PDE, which is represented as:
\begin{equation}\label{eq:burg_comp}
    \frac{\partial u}{\partial t} + u \frac{\partial u}{\partial x} = \nu \frac{\partial^2 u}{\partial x^2}
\end{equation}
where, $t$ represents time, $x$ represents the spatial variable, $\nu$ represents the kinematic viscosity and $u=u(x,t)$ represents the velocity at location $x$ and time $t$.
\subsubsection{Missing advection term in the known physics}
We first consider the case, where we only have the diffusion term in the known physics and the advection term is missing. The equation for partial physics takes the following form:
\begin{equation}\label{eq:burg_ukadv}
    \frac{\partial u}{\partial t} = \nu \frac{\partial^2 u}{\partial x^2}
\end{equation}
where $\nu=0.005/\pi$ is considered for this case. This low value is taken in order to highlight the difference between partial physics and complete physics. The objective here is to employ the proposed DPA-WNO as a surrogate to learn the temporal evolution of the field, quantify the propagation of uncertainty from the stochastic initial condition to the response, and compute the reliability of the system.

\begin{figure}[t]
\captionsetup[subfigure]{labelformat=empty}
  \centering
  \begin{tabular}{ccc}
    \begin{subfigure}{0.32\textwidth}
      \includegraphics[width=\linewidth]{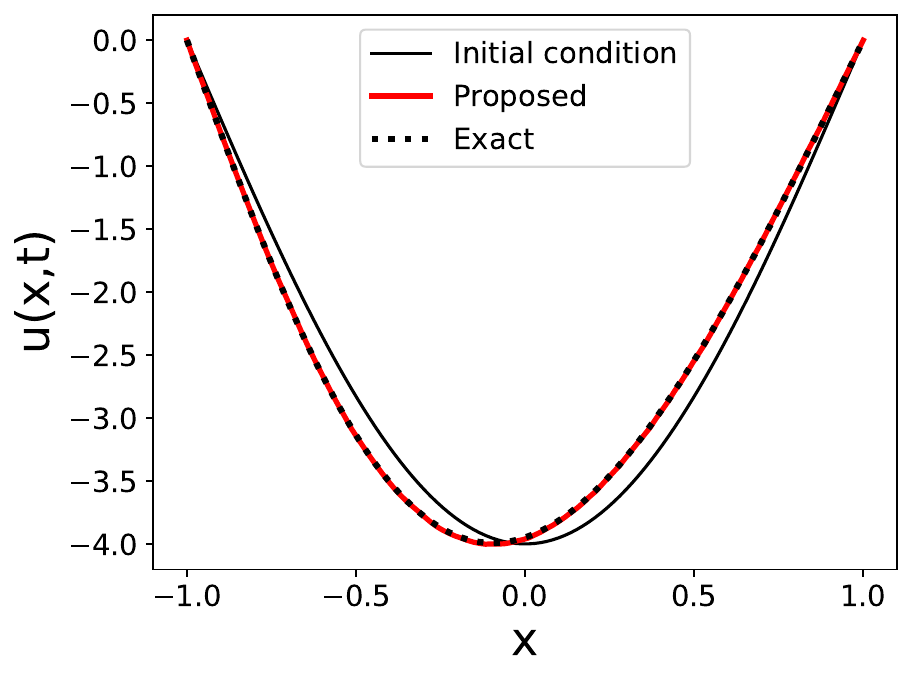}
    \end{subfigure} &
    \begin{subfigure}{0.32\textwidth}
      \includegraphics[width=\linewidth]{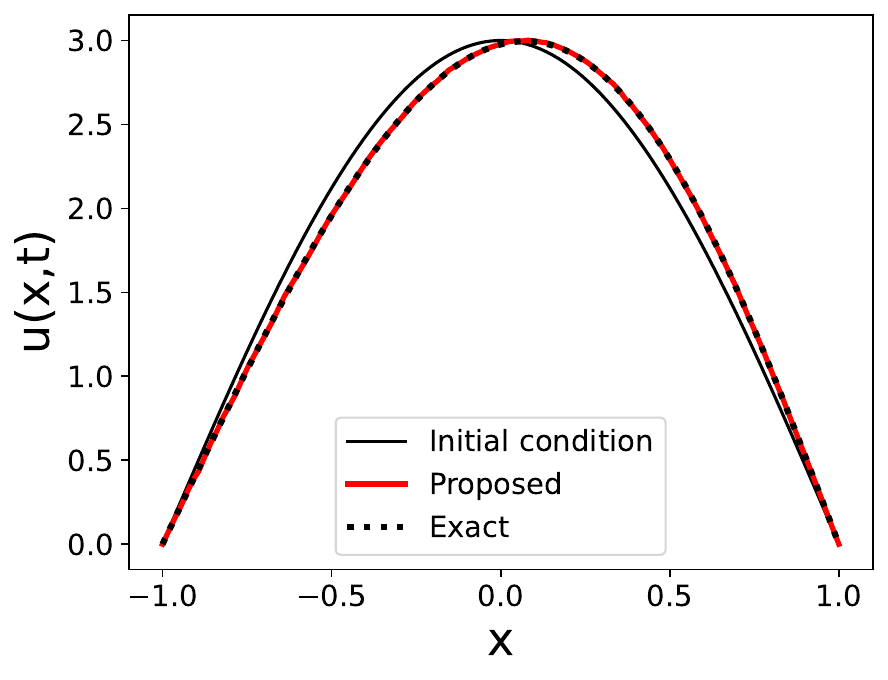}
    \end{subfigure} &
    \begin{subfigure}{0.32\textwidth}
      \includegraphics[width=\linewidth]{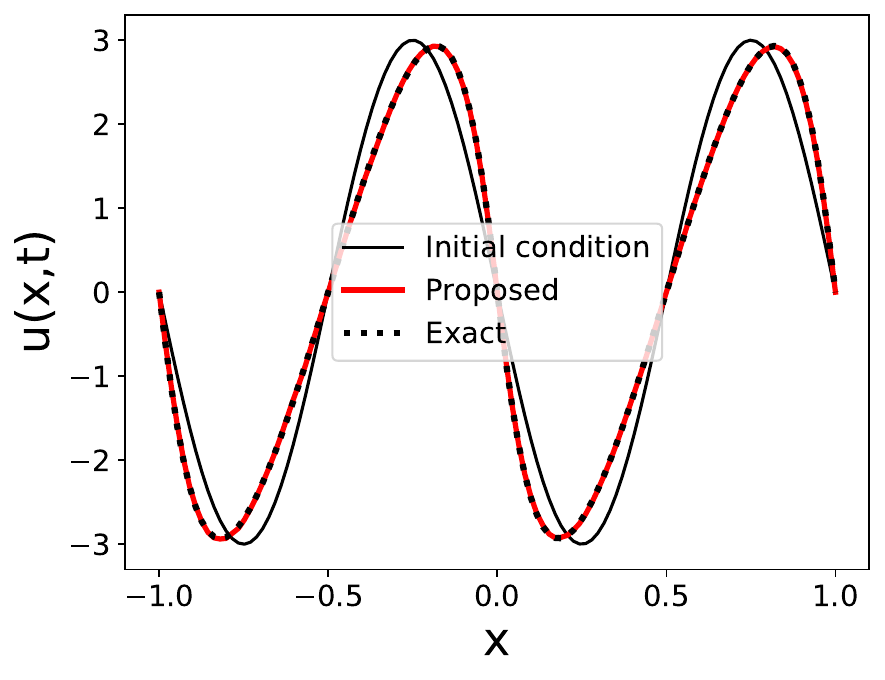}
    \end{subfigure} \\
    
    \begin{subfigure}{0.32\textwidth}
      \includegraphics[width=\linewidth]{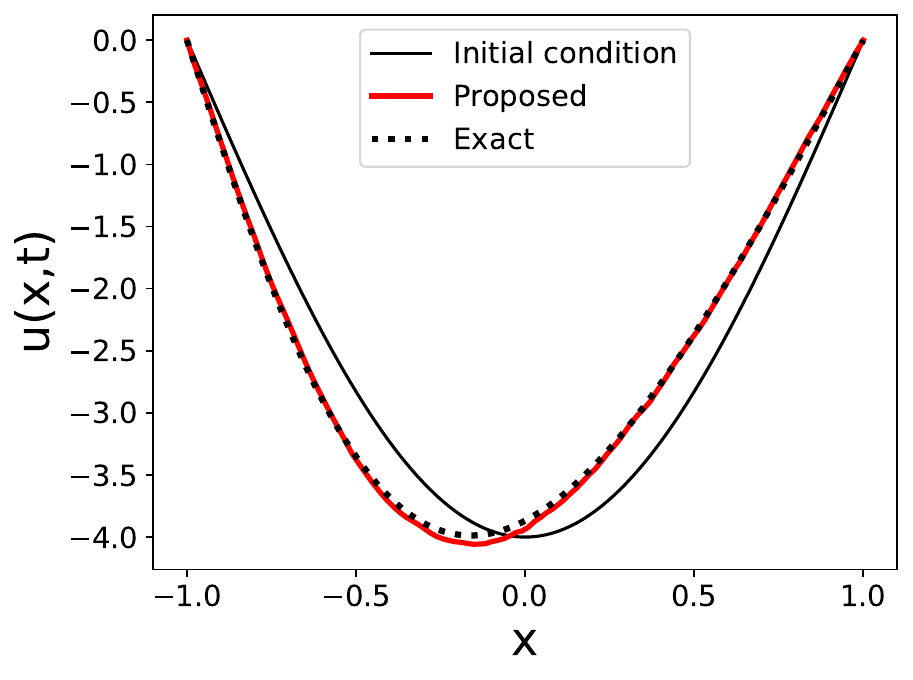}
    \end{subfigure} &
    \begin{subfigure}{0.32\textwidth}
      \includegraphics[width=\linewidth]{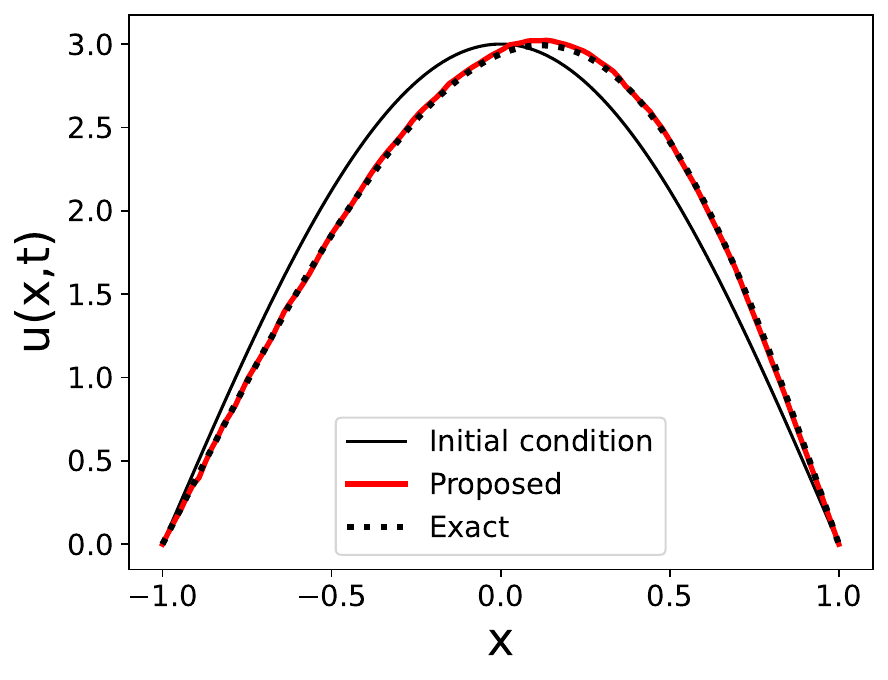}
    \end{subfigure} &
    \begin{subfigure}{0.32\textwidth}
      \includegraphics[width=\linewidth]{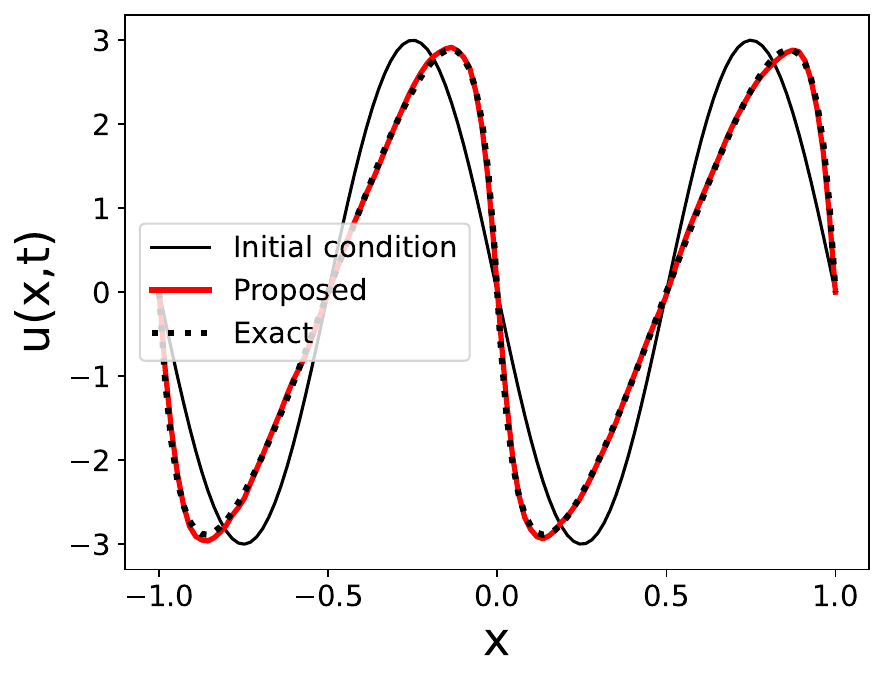}
    \end{subfigure}
  \end{tabular}
  \caption{Comparison of predicted output $u(x,t)$ obtained using DPA-WNO, with the ground truth for Burgers' equation with missing advection term. The comparison is shown for three different initial conditions taken from the trained samples. The top row corresponds to predictions within the training window ($t=48\Delta t$) while the bottom row corresponds to predictions outside the training window ($t= 101\Delta t$)}
  \label{fig:Burgmadv_pred}
\end{figure} 

 \begin{figure}[t]
\captionsetup[subfigure]{labelformat=empty}
  \centering
  \begin{tabular}{ccc}
    \begin{subfigure}{0.32\textwidth}
      \includegraphics[width=\linewidth]{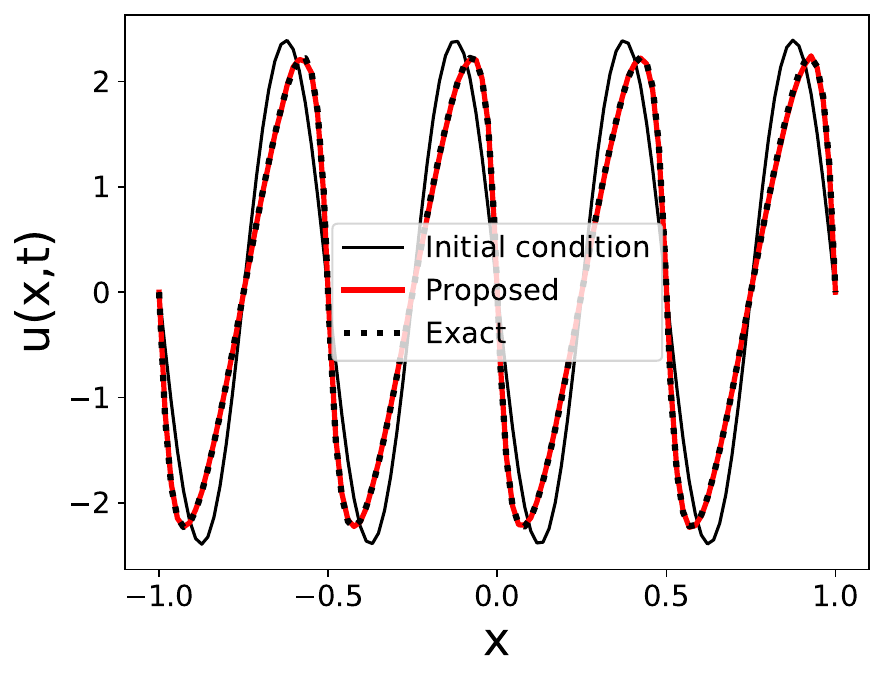}
    \end{subfigure} &
    \begin{subfigure}{0.32\textwidth}
      \includegraphics[width=\linewidth]{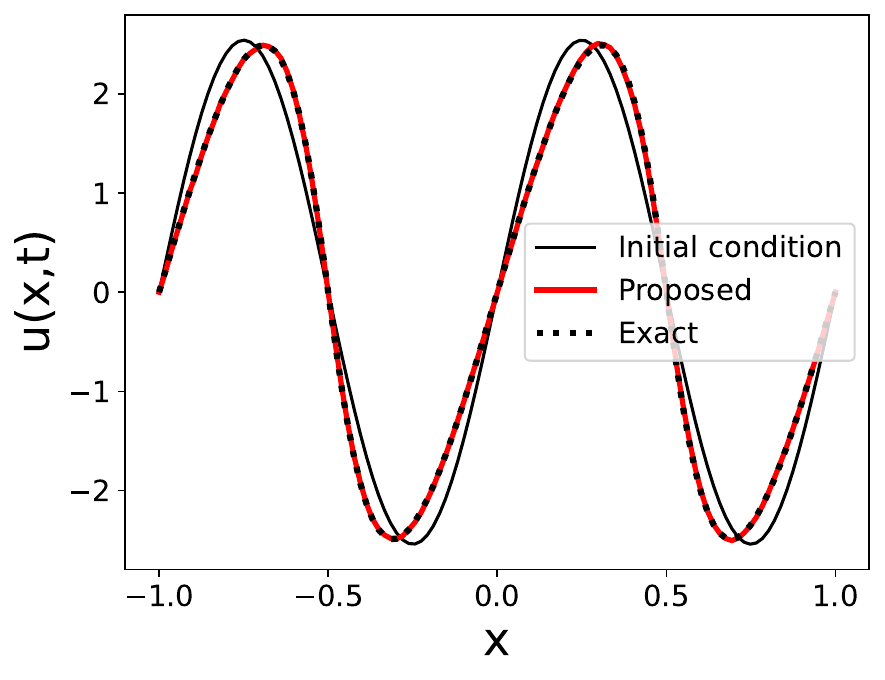}
    \end{subfigure} &
    \begin{subfigure}{0.32\textwidth}
      \includegraphics[width=\linewidth]{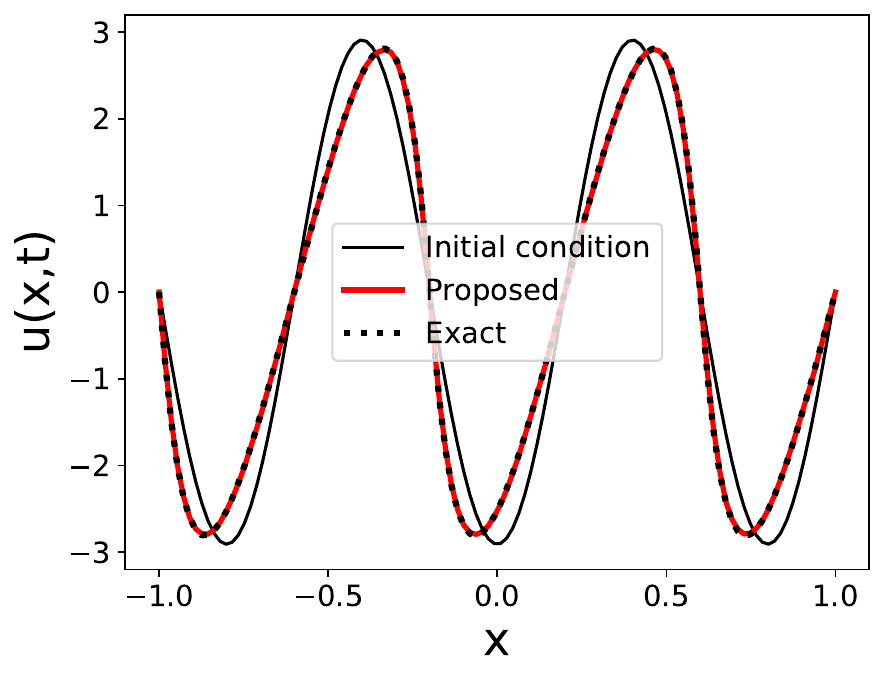}
    \end{subfigure} \\
    
    \begin{subfigure}{0.32\textwidth}
      \includegraphics[width=\linewidth]{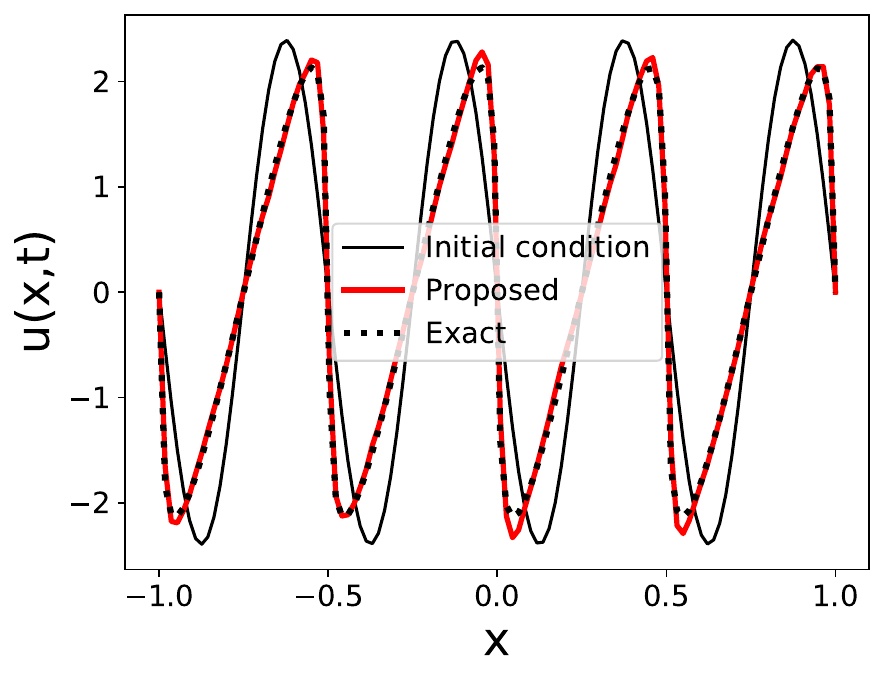}
    \end{subfigure} &
    \begin{subfigure}{0.32\textwidth}
      \includegraphics[width=\linewidth]{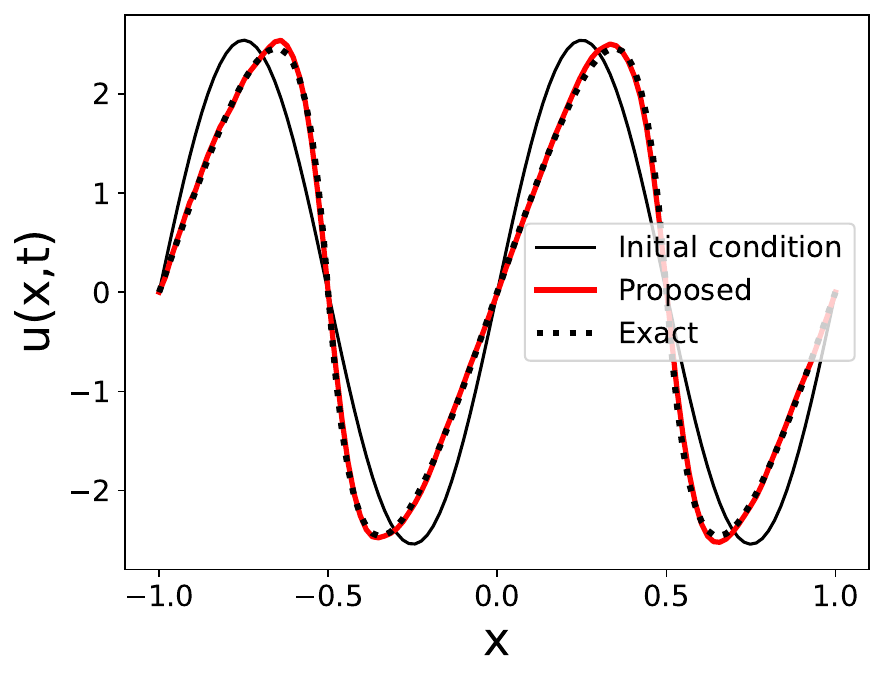}
    \end{subfigure} &
    \begin{subfigure}{0.32\textwidth}
      \includegraphics[width=\linewidth]{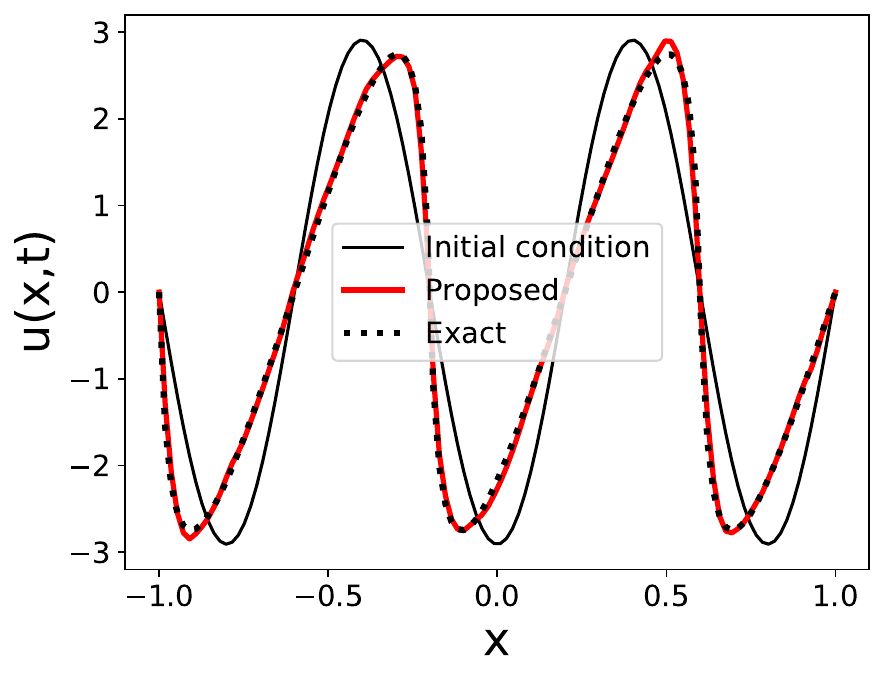}
    \end{subfigure}
  \end{tabular}
\caption{Comparison of predicted output $u(x,t)$ obtained using DPA-WNO, with the ground truth for Burgers' equation with missing advection term. The comparison is shown for three different initial conditions other than the trained samples. The top row corresponds to predictions within the training window ($t=48\Delta t$) while the bottom row corresponds to predictions outside the training window ($t=101\Delta t$)}
  \label{fig:Burgmadv_gen}
\end{figure}

In order to demonstrate the applicability of model, we have synthetically generated the ground truth data, by solving the complete physics model i.e. the complete Burgers' equation. In the generated high fidelity data $\mathcal{D} = \left\{\bm{u}_{0,1:N_x}^{(i)},\bm{u}^{(i)}_{1:N_t,1:N_x} \right\}_{i=1}^N $, we have taken $N_x = 112$ and $N_t = 50$ for training data set. We created $N=32$ sample solution sets by considering different initial conditions $\bm{u}_0(x)$. Half of the initial conditions are of the form $\bm{u}_0(x) = \alpha cos(0.5\zeta \pi x)$ and the other half of them are of the form $\bm{u}_0(x) = \beta sin(\eta \pi x)$, where $\alpha$, $\beta$ $\in$ $\left\{-8,-7,-6...,-1,1,2..8\right\}$ and $\zeta \in \left\{1,5\right\}, \eta \in \left\{2,4\right\}$.

\begin{figure}
\captionsetup[subfigure]{labelformat=empty}
  \centering
  \begin{tabular}{cc}
    \begin{subfigure}{0.48\textwidth}
      \includegraphics[width=\linewidth]{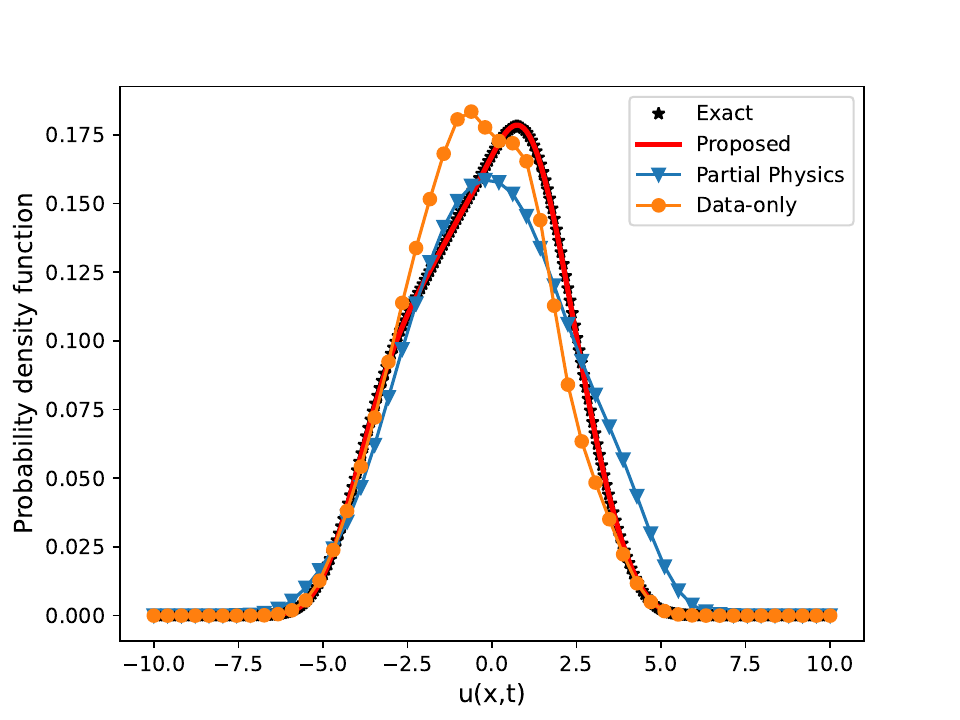}
      \caption{PDF when $t < t_{train}$}
      \label{fig:Burgmadv_pdfs_1}
    \end{subfigure} &
    \begin{subfigure}{0.48\textwidth}
      \includegraphics[width=\linewidth]{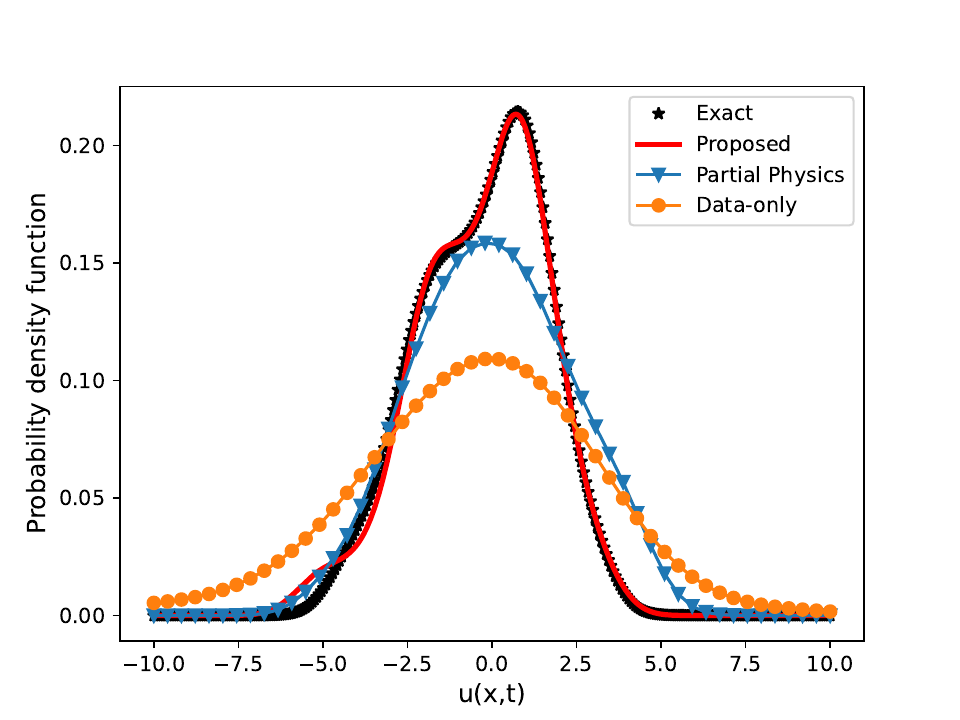}
      \caption{PDF when $t > t_{train}$}
      \label{fig:Burgmadv_pdfs_2}
    \end{subfigure}
  \end{tabular}
 \caption{Comparison of the probability density functions (PDFs) of the output $u(x=-0.4107, t=48\Delta t)$ (left) and $u(x=-0.4107, t=96\Delta t)$ (right) obtained using four different models: DPA-WNO (Proposed), Known physics (Partial physics), data-based WNO (Data only), and the ground truth (Exact). The results correspond to the missing advection case for Burger's equation}
  \label{fig:Burgmadv_pdfs}
\end{figure}

The model is trained till $50$ time steps, with $\Delta t = 0.0005$ sec, using the adaptive training strategy discussed earlier in Sec.\ref{subsubsec:training}. Gradient-based ADAM optimizer with a constant learning rate of $0.005$ is used. The testing phase involves randomly selecting 100 initial conditions for evaluation. These are divided into two equal groups: one group consists of initial conditions of the form $\bm{u}_0(x) = \alpha \cos(0.5\zeta \pi x)$, and the other group consists of initial conditions of the form $\bm{u}_0(x) = \beta \sin(\eta \pi x)$. The parameters $\alpha$ and $\beta$ are sampled from uniform distributions $\mathcal{U}(-10,10)$, ensuring a wide range of values. While, the parameters $\zeta \in \left\{1,5\right\}, \eta \in \left\{2,4\right\}$ are unchanged.

Fig.\ref{fig:Burgmadv_pred} presents a comparison between the predictions generated by DPA-WNO and the ground truth. The comparison is shown using three distinct initial conditions derived from the training samples. The results are accurate for the predictions extended to twice the time of the training window, demonstrating the extrapolative ability of the proposed framework. Fig.\ref{fig:Burgmadv_gen} presents the predictions obtained for three distinct initial conditions that are not included in the training data. The precise and accurate results displayed in the figure showcase the model's ability to generalize well to diverse initial conditions.

Fig.\ref{fig:Burgmadv_pdfs} showcases the comparison of probability density functions (PDFs) obtained using the DPA-WNO, data-driven WNO, known physics model, and the true physics model (ground truth). The results obtained using the proposed model demonstrate excellent match with the ground truth, both within and beyond the training time window, demonstrating its strong predictive and generalization capabilities for a diverse set of stochastic initial conditions. In contrast, the known physics model exhibits a noticeable deviation from the ground truth due to the absence of the advection term. This deviation becomes more pronounced outside the training window, as accumulated errors contribute to the offset. The purely data-driven WNO model could not perform as well as compared to the proposed approach. It displays a more significant offset in the extrapolation region, highlighting its limited ability to accurately predict way beyond the training time window. Additionally, the Mean Squared Error (MSE) and Hellinger distance values provide further evidence of the superiority of the proposed model. The proposed DPA-WNO model achieves a remarkably low MSE of $0.4624$, significantly outperforming the known physics model (MSE=$2.916$) and purely data-driven WNO model (MSE=$11.0227$)  as shown in the Table \ref{tab:MSEHD}. DPA-WNO  outperformed the other two models in terms of Helinger Distance  as well, with a value of 0.0659, as opposed to 0.2867 for data-driven WNO and 0.5968 for known physics model. These results demonstrate the exceptional accuracy of the DPA-WNO model in generalizing to a wide set of initial conditions.

As previously mentioned, for the reliability analysis, we have sampled 5000 initial conditions from the Gaussian random field, using the kernel described in Eq. \ref{eq:kernel}. For this example, the scaling factor $\alpha$ is set to $4$ the length scale value $l$ is set to $0.5$, and the periodicity value $p$ is set to 1. The threshold $g_t$ for the maximum magnitude of the output is defined as 7 units. The application of DPA-WNO yields a reliability value of 95.2\%, which closely aligns with the actual reliability of 95.72\% derived from the ground truth, as presented in Table\ref{tab:Reliability}.

\subsubsection{Missing diffusion in the known physics}
We now consider the second case where the known physics takes the following form,
\begin{equation}
    \frac{\partial u}{\partial t} + u \frac{\partial u}{\partial x} = 0
\end{equation}

 \begin{figure}[t]
\captionsetup[subfigure]{labelformat=empty}
  \centering
  \begin{tabular}{ccc}
    \begin{subfigure}{0.32\textwidth}
      \includegraphics[width=\linewidth]{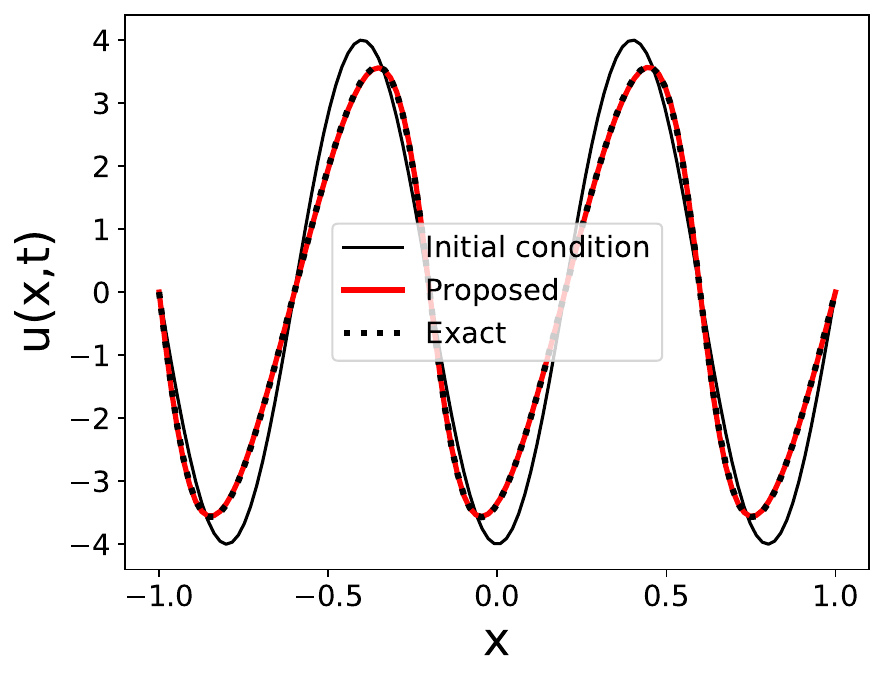}
      \caption{}
    \end{subfigure} &
    \begin{subfigure}{0.32\textwidth}
      \includegraphics[width=\linewidth]{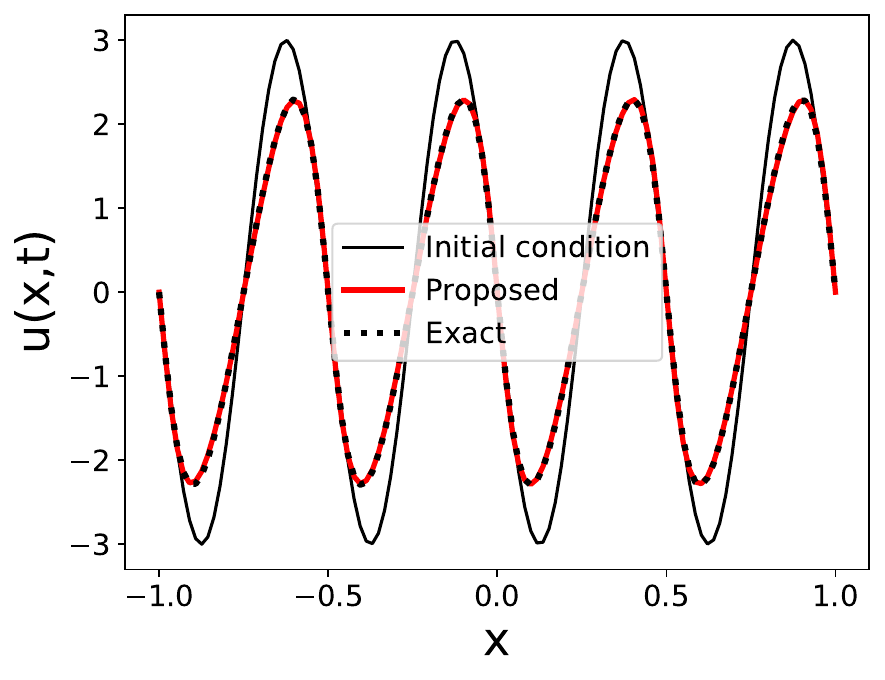}
      \caption{}
    \end{subfigure} &
    \begin{subfigure}{0.32\textwidth}
      \includegraphics[width=\linewidth]{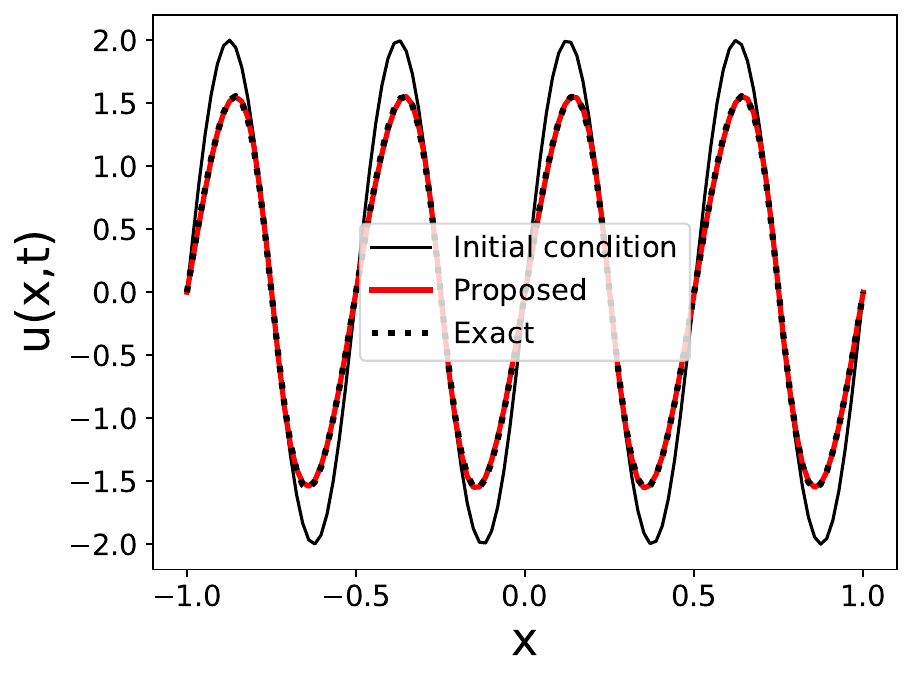}
      \caption{}
    \end{subfigure} \\
    
    \begin{subfigure}{0.32\textwidth}
      \includegraphics[width=\linewidth]{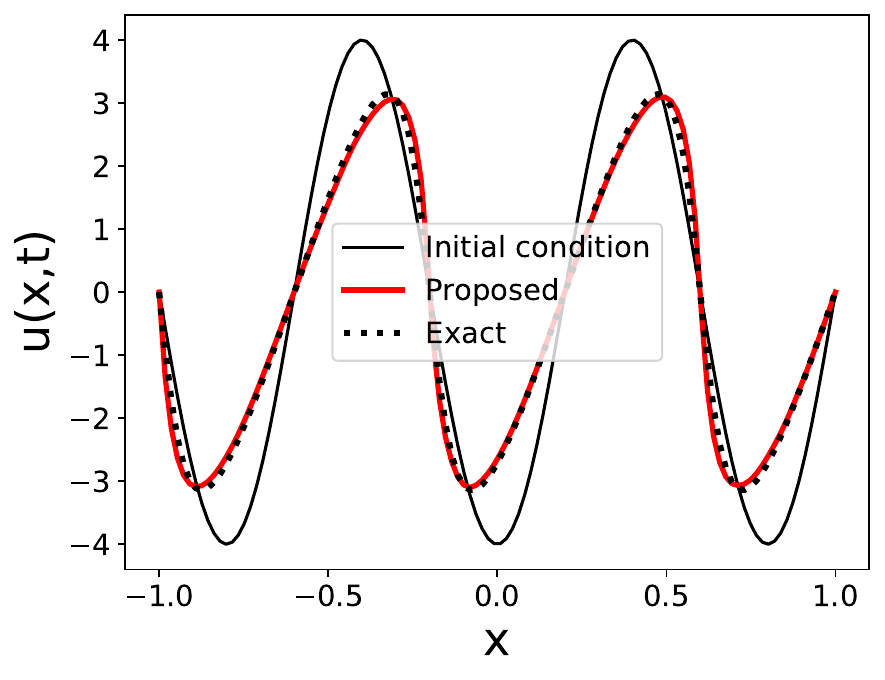}
      \caption{}
    \end{subfigure} &
    \begin{subfigure}{0.32\textwidth}
      \includegraphics[width=\linewidth]{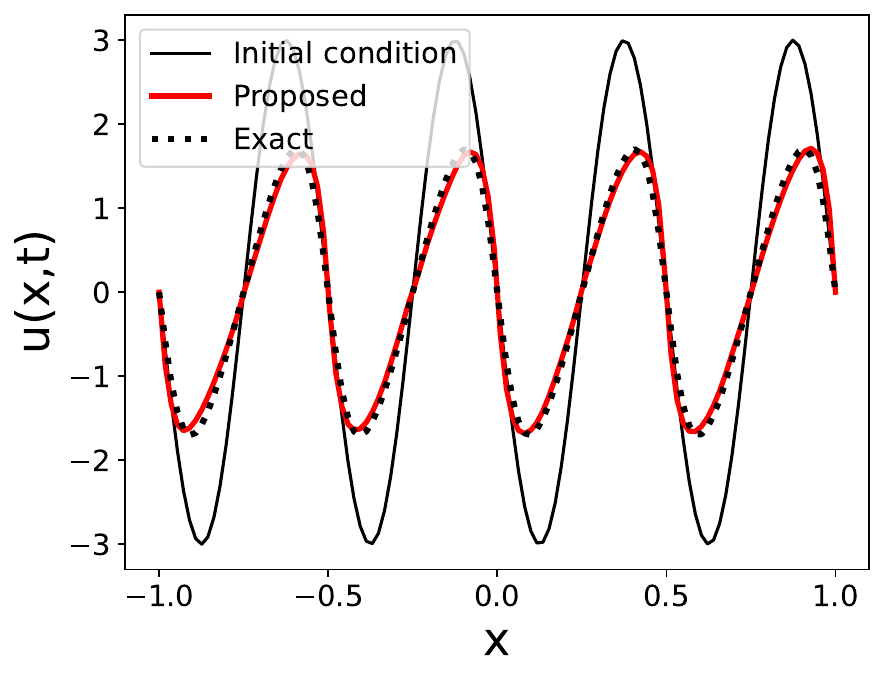}
      \caption{}
    \end{subfigure} &
    \begin{subfigure}{0.32\textwidth}
      \includegraphics[width=\linewidth]{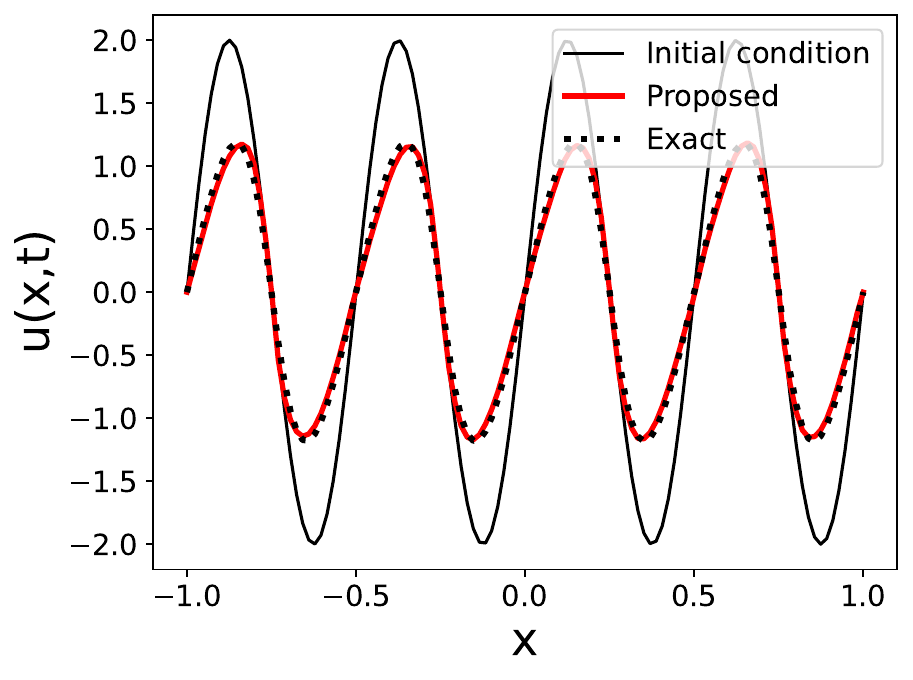}
      \caption{}
    \end{subfigure}
  \end{tabular}
  \caption{Comparison of predicted output $u(x,t)$ obtained using DPA-WNO, with the ground truth for Burgers' equation with missing diffusion term. The comparison is shown for three different initial conditions taken from the trained samples. The top row corresponds to predictions within the training window ($t=48\Delta t$) while the bottom row corresponds to predictions outside the training window ($t=101\Delta t$)}
  \label{fig:Burgmdiff_pred}
\end{figure}

 \begin{figure}[t]
\captionsetup[subfigure]{labelformat=empty}
  \centering
  \begin{tabular}{ccc}
    \begin{subfigure}{0.32\textwidth}
      \includegraphics[width=\linewidth]{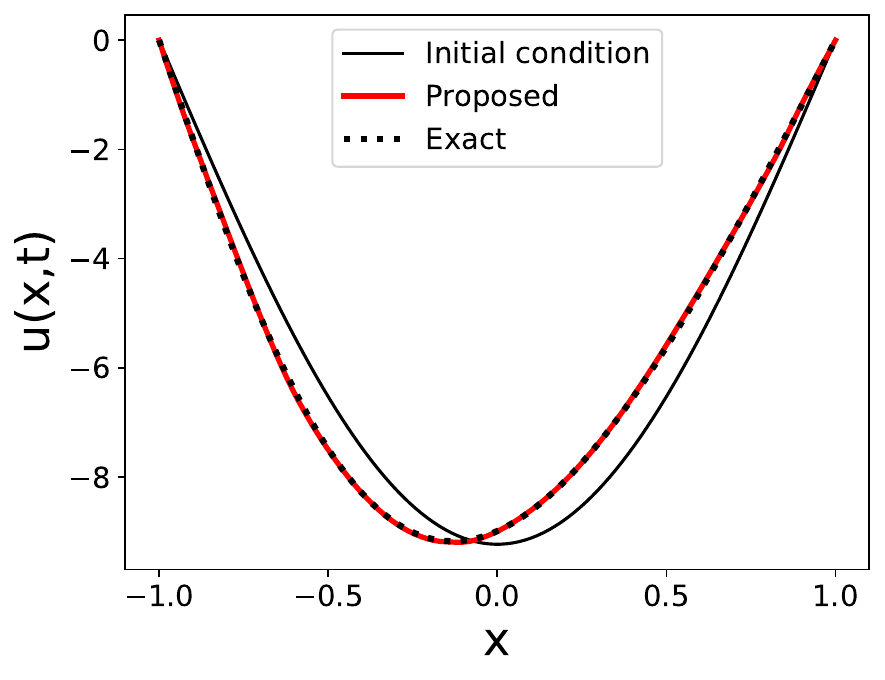}
      \caption{}
    \end{subfigure} &
    \begin{subfigure}{0.32\textwidth}
      \includegraphics[width=\linewidth]{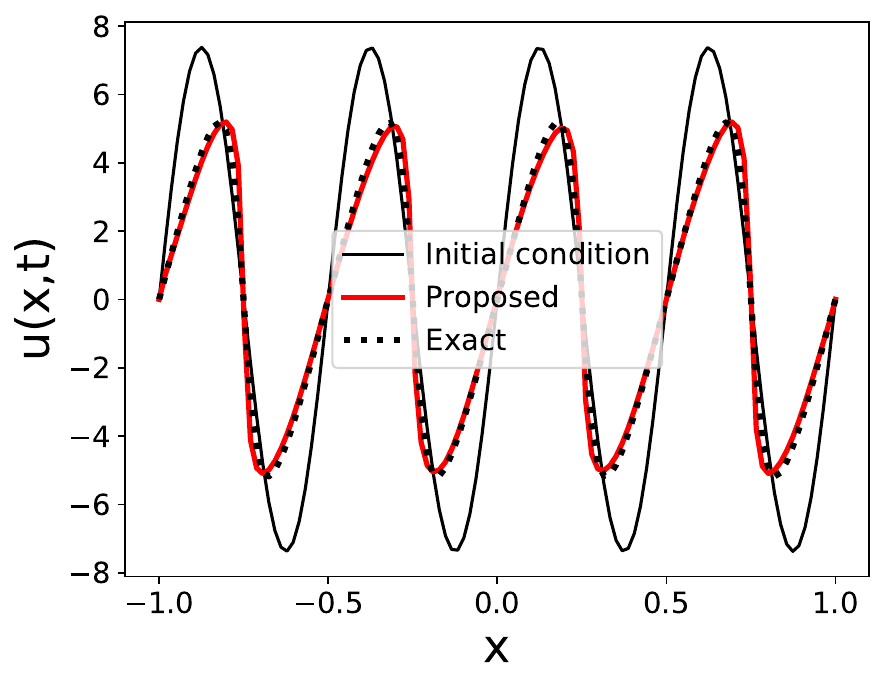}
      \caption{}
    \end{subfigure} &
    \begin{subfigure}{0.32\textwidth}
      \includegraphics[width=\linewidth]{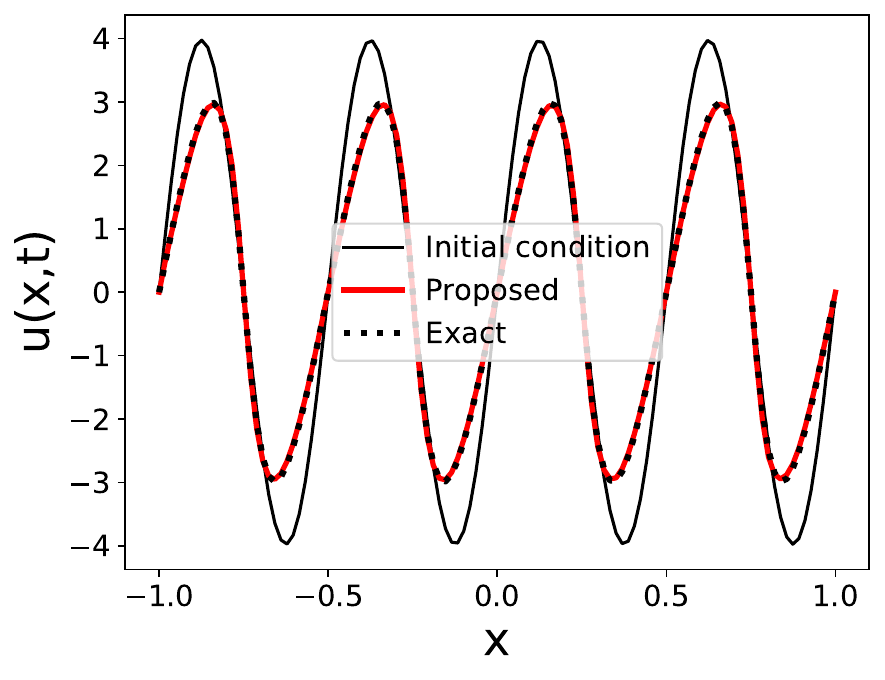}
      \caption{}
    \end{subfigure} \\
    
    \begin{subfigure}{0.32\textwidth}
      \includegraphics[width=\linewidth]{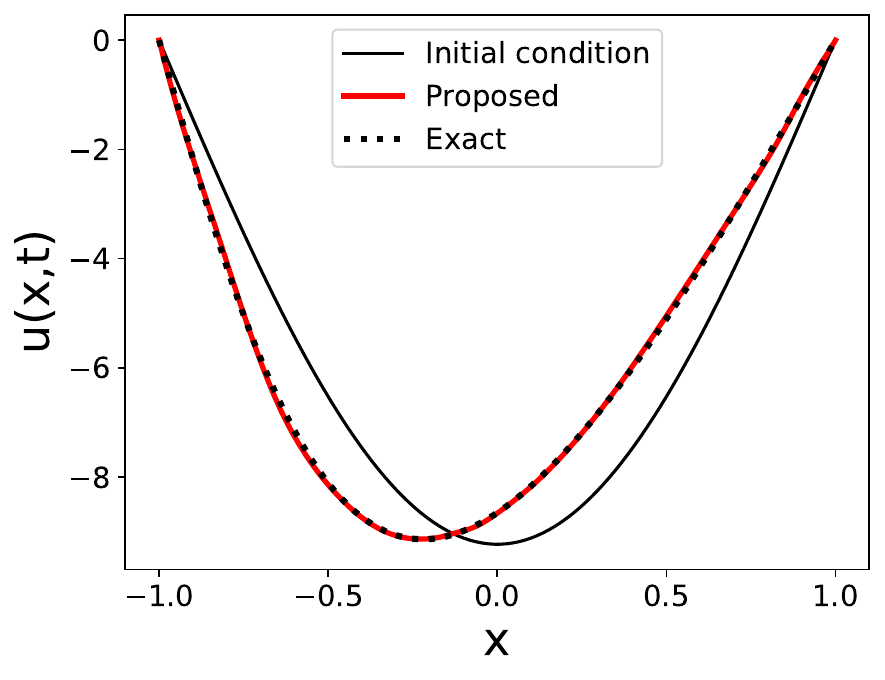}
      \caption{}
    \end{subfigure} &
    \begin{subfigure}{0.32\textwidth}
      \includegraphics[width=\linewidth]{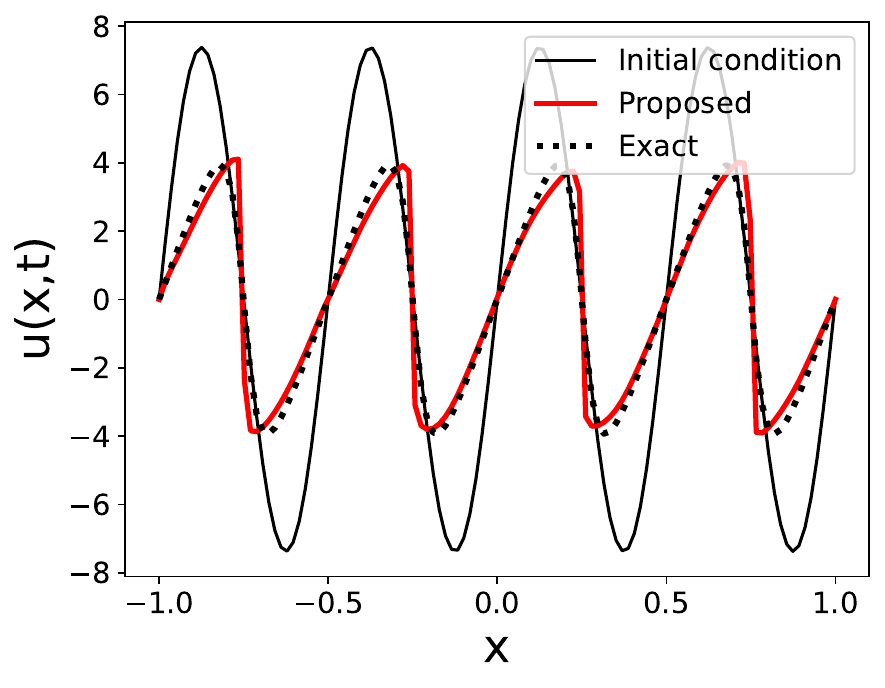}
      \caption{}
    \end{subfigure} &
    \begin{subfigure}{0.32\textwidth}
      \includegraphics[width=\linewidth]{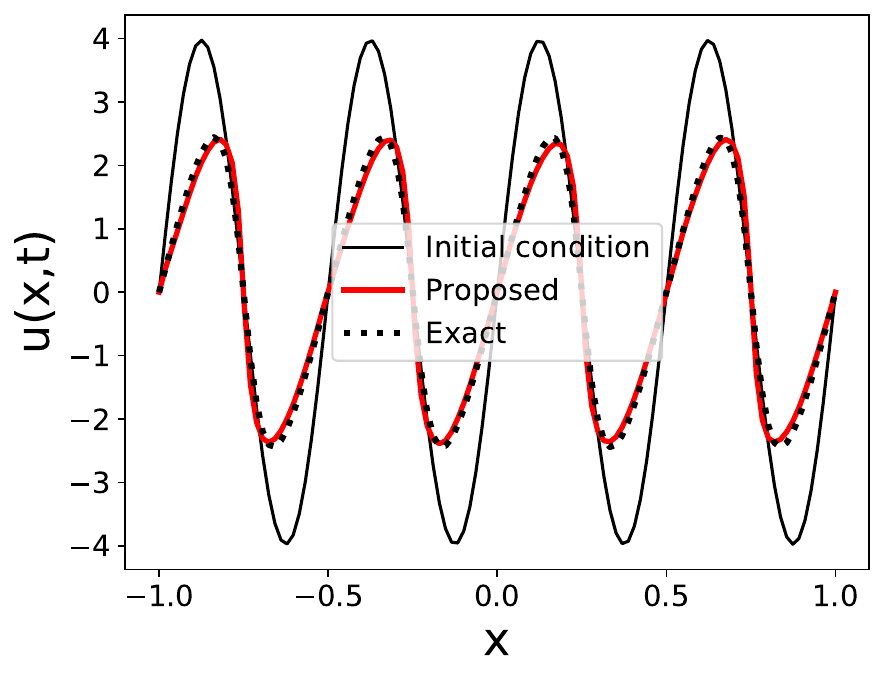}
      \caption{}
    \end{subfigure}
  \end{tabular}
  \caption{Comparison of predicted output $u(x,t)$ obtained using DPA-WNO, with the ground truth for Burgers equation with missing diffusion term. The comparison is shown for three different initial conditions other than the trained samples.  The top row corresponds to predictions within the training window ($t=48\Delta t$) while the bottom row corresponds to predictions outside the training window ($t=101\Delta t$)}
  \label{fig:Burgmdiff_gen}
\end{figure}

\begin{figure}[t]
\captionsetup[subfigure]{labelformat=empty}
  \centering
  \begin{tabular}{cc}
    \begin{subfigure}{0.48\textwidth}
      \includegraphics[width=\linewidth]{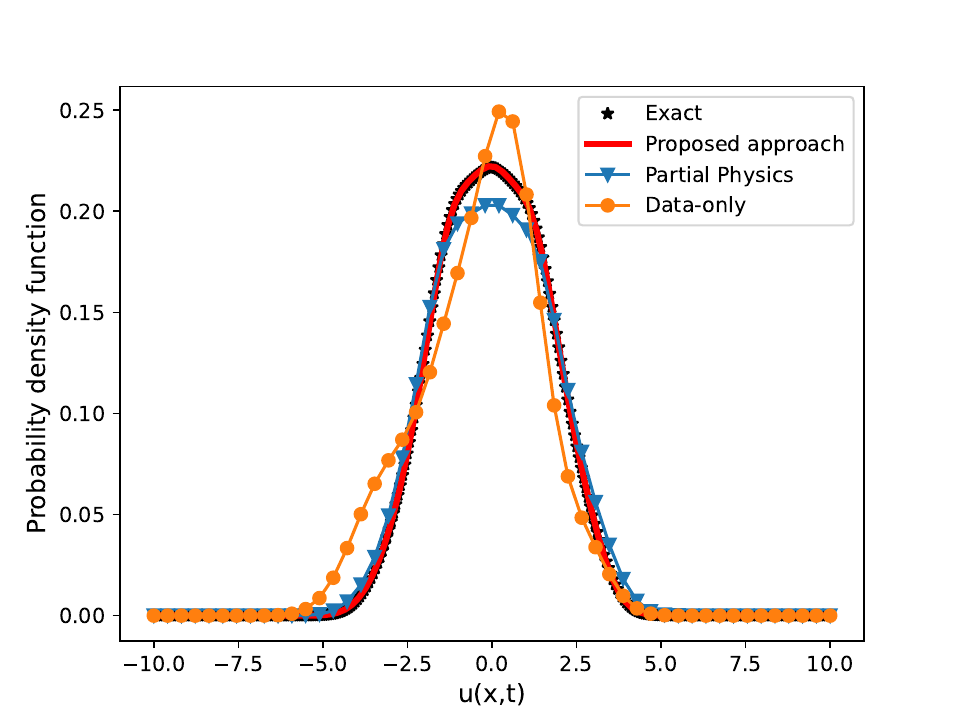}
      \caption{PDF when $t < t_{train}$}
    \end{subfigure} &
    \begin{subfigure}{0.48\textwidth}
      \includegraphics[width=\linewidth]{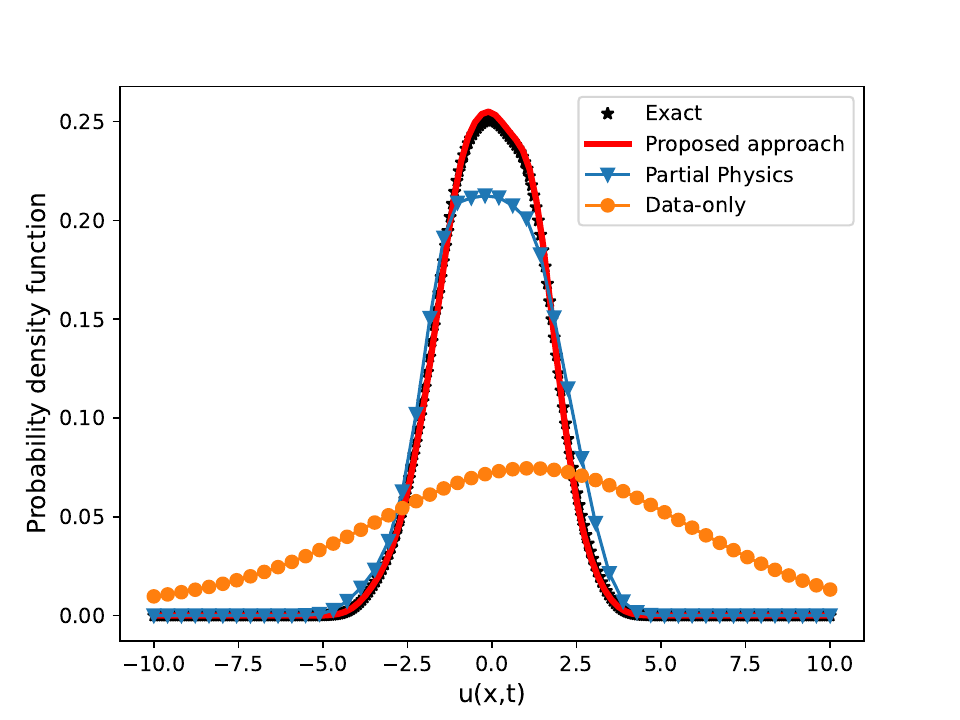}
      \caption{PDF when $t > t_{train}$}
    \end{subfigure}
  \end{tabular}
  \caption{Comparison of the probability density functions (PDFs) of the output $u(x=-0.3571,t=48\Delta t)$ (left) and $u(x=-0.3571,t=97\Delta t)$ (right) for the Burgers' equation (missing diffusion scenario) from four different models: DPA-WNO (Proposed), Known physics (Partial physics), data-based WNO (Data only), and the ground truth (Exact)}
  \label{fig:Burgmdiff_pdfs}
\end{figure}

 \noindent In generating high fidelity data $\mathcal{D} = \left\{\bm{u}_{0,1:N_x}^{(i)},\bm{u}^{(i)}_{1:N_t,1:N_x} \right\}_{i=1}^N $, we have taken $\nu = 0.3/\pi$, $N_x = 112$ and $N_t = 50$ with $\Delta t = 0.0003$. All other training and testing setup conditions remain unchanged from the previous case.

The results shown in Fig. \ref{fig:Burgmdiff_pred} demonstrate accurate predictions beyond the training window. Fig. \ref{fig:Burgmdiff_gen} shows precise predictions for diverse initial conditions not included in the training data, indicating its generalization ability. Fig. \ref{fig:Burgmdiff_pdfs} showcases the response PDF obtained using the proposed DPA-WNO, data-driven WNO, known physics and actual physics (ground truth). Results obtained using the proposed model align well with the ground truth, while the results obtained using the known physics model deviate due to missing terms. The data-driven WNO model exhibits limited accuracy, specifically in beyond the training window. The proposed model achieves a remarkably low MSE and mean Hellinger distance with values of $0.118$ and $0.0141$ respectively, outperforming the other models (see Table \ref{tab:MSEHD}).

For reliability analysis, the same setup as the previous case study is adopted. The proposed DPA-WNO yield a reliability value of 95.72\%, exactly matching the ground truth reliability (see \ref{tab:Reliability}).

\subsection{Example 2: Nagumo equation} \label{subsec:eg2}
In this example, we consider the Nagumo equation, which has various applications including modeling of wave propagation in neurons. Specifically, it is used to study the dynamics of voltage across nerve cells and the impulses in nerve fibers. The Nagumo equation, denoted by \cite{laing2009stochastic}, incorporating periodic boundary condition is expressed as follows:
\begin{equation}\label{eq: Nagumo_comp}
\begin{aligned}
  &\frac{\partial u}{\partial t} - \epsilon \frac{\partial^2 u}{\partial x^2} = u(1-u)(u-\alpha), \quad x \in (0,1), t \in (0,T)\\
   &u(x=0,t) = u(x=1,t),\quad x \in (0,1), t \in (0,T)\\
   &u_x(x=0,t) = u_x(x=1,t),\quad x \in (0,1), t \in (0,T)\\
   &u(x,t=0) = u_0(x), \quad x \in (0,1)\\ 
\end{aligned}
\end{equation}
where, the parameter $\alpha\in \mathbb{R}$ determines the wave's speed along the axon, while $\epsilon > 0$ controls the diffusion rate. 
\subsubsection{Only diffusion as the known physics}
First, we examine the scenario where only the diffusion term is present in the known physics, while the cubic term is absent. The equation of known physics takes the following form:
\begin{equation}
     \frac{\partial u}{\partial t} - \epsilon \frac{\partial^2 u}{\partial x^2} = 0, \quad x \in (0,1), t \in (0,T)
\end{equation}
The initial and boundary conditions remain the same as in Eq.\ref{eq: Nagumo_comp}.
In this case, we have considered $\epsilon=0.0001$ and $\alpha=-10$ to emphasize the disparity between partial physics and complete physics. The objective is to employ DPA-WNO as a surrogate to learn the temporal evolution, quantify the propagation of uncertainty from the stochastic initial condition to the output, and compute the reliability of the system.

\begin{figure}[t]
\captionsetup[subfigure]{labelformat=empty}
  \centering
  \begin{tabular}{ccc}
    \begin{subfigure}{0.32\textwidth}
      \includegraphics[width=\linewidth]{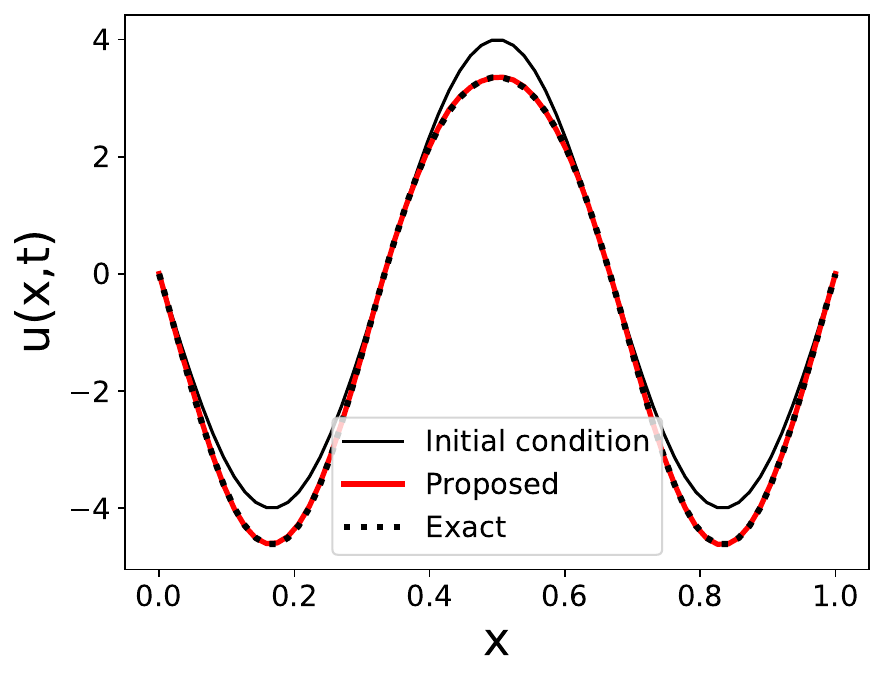}
      \caption{}
    \end{subfigure} &
    \begin{subfigure}{0.32\textwidth}
      \includegraphics[width=\linewidth]{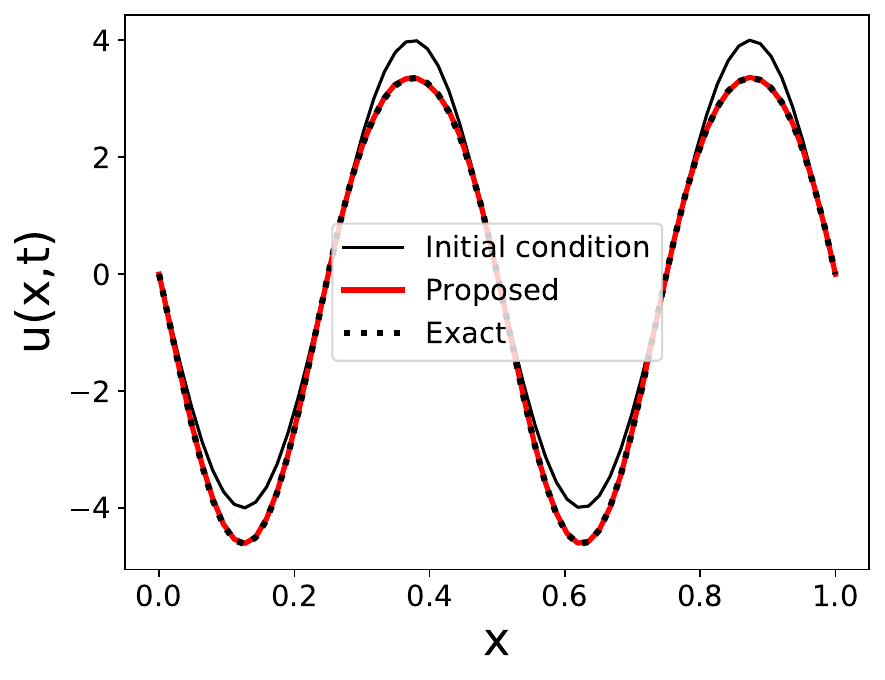}
      \caption{}
    \end{subfigure} &
    \begin{subfigure}{0.32\textwidth}
      \includegraphics[width=\linewidth]{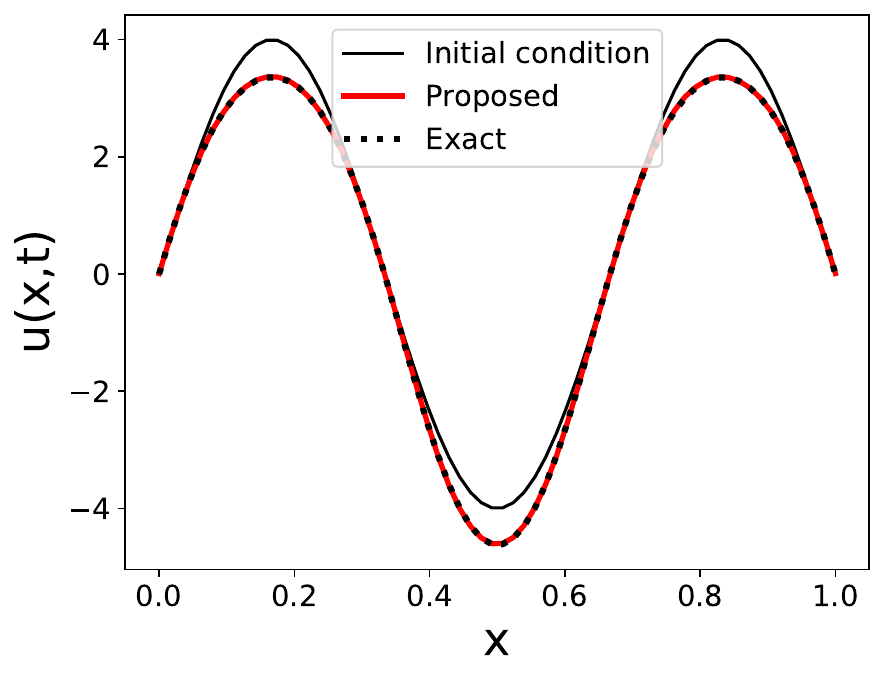}
      \caption{}
    \end{subfigure} \\
    
    \begin{subfigure}{0.32\textwidth}
      \includegraphics[width=\linewidth]{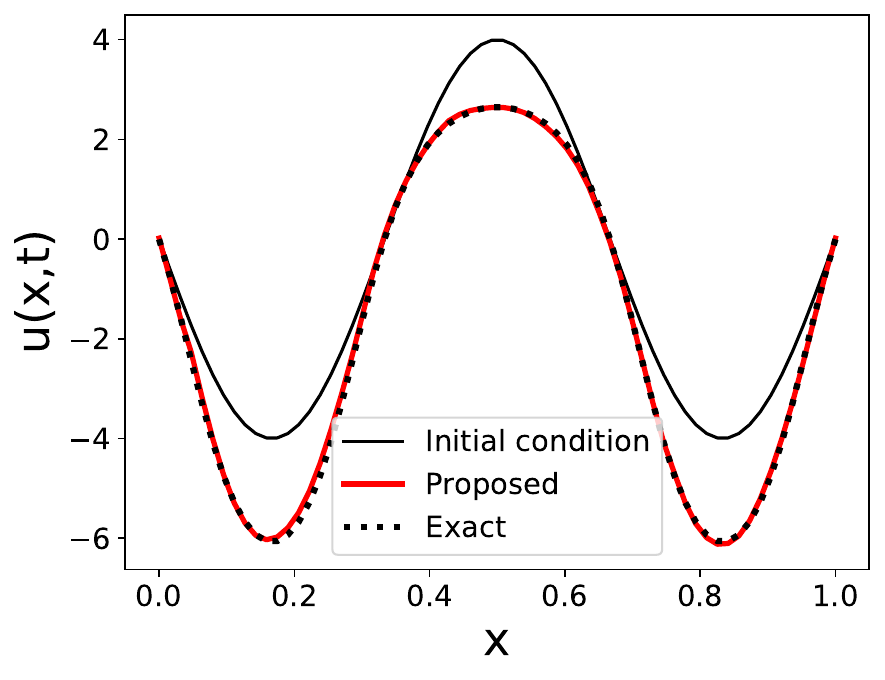}
      \caption{}
    \end{subfigure} &
    \begin{subfigure}{0.32\textwidth}
      \includegraphics[width=\linewidth]{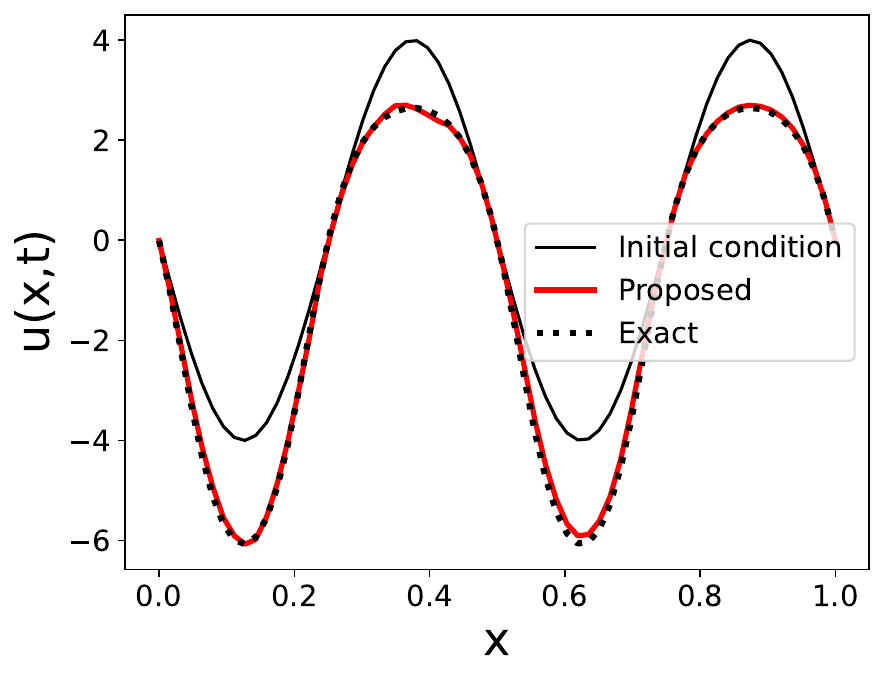}
      \caption{}
    \end{subfigure} &
    \begin{subfigure}{0.32\textwidth}
      \includegraphics[width=\linewidth]{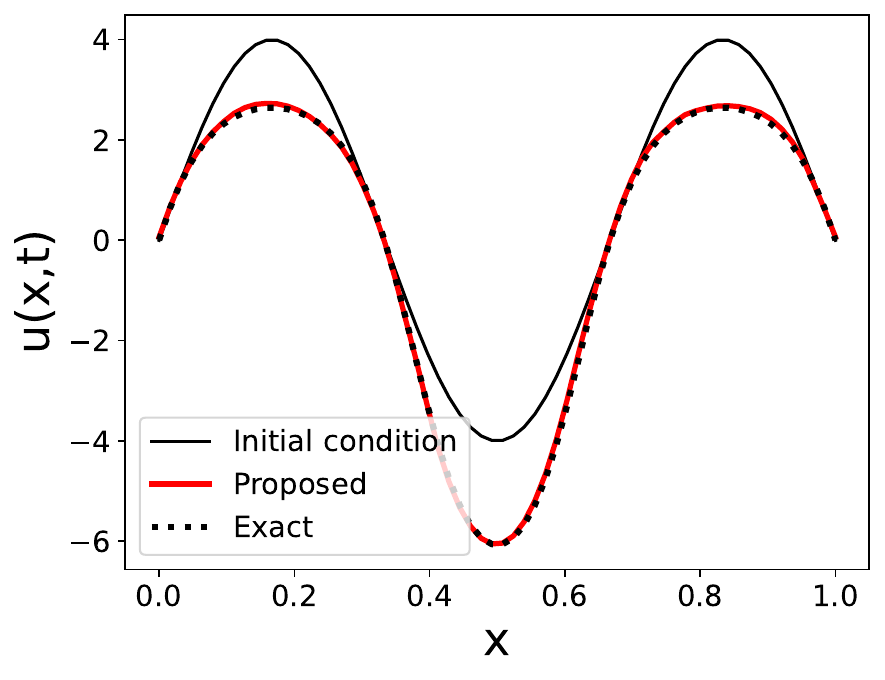}
      \caption{}
    \end{subfigure}
  \end{tabular}
  \caption{Comparison of predicted output $u(x,t)$ obtained using DPA-WNO, with the ground truth for Nagumo equation with missing advection term. The comparison is shown for three different initial conditions taken from the trained samples. The top row corresponds to predictions within the training window ($t=48\Delta t$) while the bottom row corresponds to predictions outside the training window ($t=141\Delta t$)}
  \label{fig:Nagmcub_pred}
\end{figure}

To illustrate the performance of the proposed approach, we generated synthetic data by solving the complete Nagumo equation. The high fidelity data  $\mathcal{D} = \left\{\bm{u}_{0,1:N_x}^{(i)},\bm{u}^{(i)}_{1:N_t,1:N_x} \right\}_{i=1}^N$, consists of $N=32$ sample solutions, with different initial conditions $\bm{u}_{0,1:N_x}^{(i)}$ of the form $\alpha \sin(\eta\pi x)$ where $\alpha \in \left\{-4, -3,-2, -1, 1, 2,3,4\right\}$ and $\eta \in \left\{2, 3, 4, 5\right\}$. For this case, we have considered $N_x = 64$. The model is trained for $N_t = 50$ time steps, with $\Delta t =0.0001$ seconds. We employ an adaptive training strategy discussed earlier in Sec. \ref{subsubsec:training}. The training utilizes the ADAM optimizer with a constant learning rate of $0.002$.

During the testing phase, a total of 100 initial conditions were randomly selected for evaluation. The initial conditions are of the form $\bm{u}_0(x) = \alpha \sin(\eta \pi x)$. The parameter $\alpha$ is sampled from the uniform distributions $\mathcal{U}(-10,10)$ (different from the training data) and the parameter $\eta$, can take values of $\left\{2, 3, 4, 5\right\}$ (same as training)

 \begin{figure}[t]
\captionsetup[subfigure]{labelformat=empty}
  \centering
  \begin{tabular}{ccc}
    \begin{subfigure}{0.32\textwidth}
      \includegraphics[width=\linewidth]{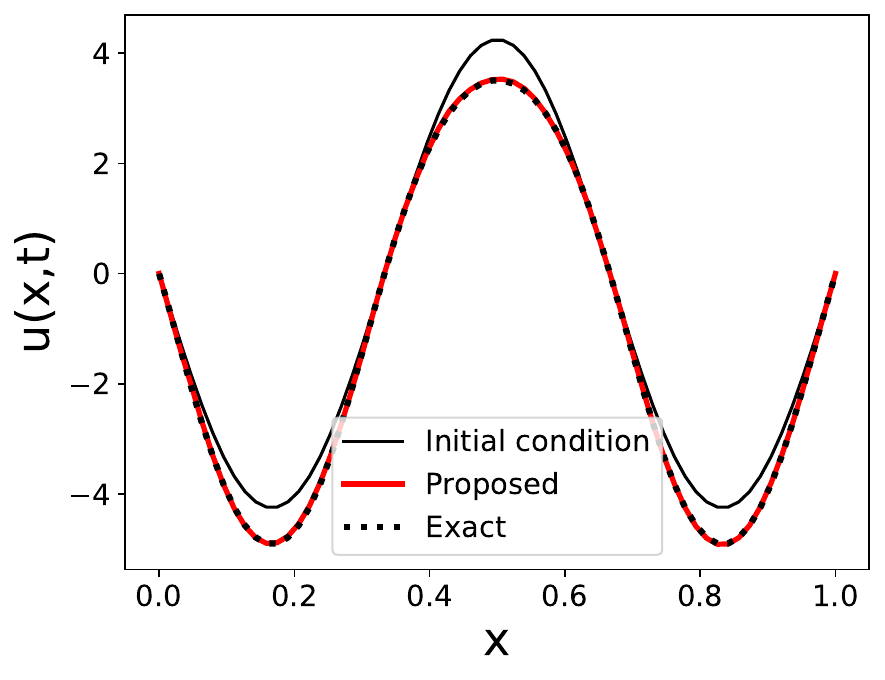}
      \caption{}
    \end{subfigure} &
    \begin{subfigure}{0.32\textwidth}
      \includegraphics[width=\linewidth]{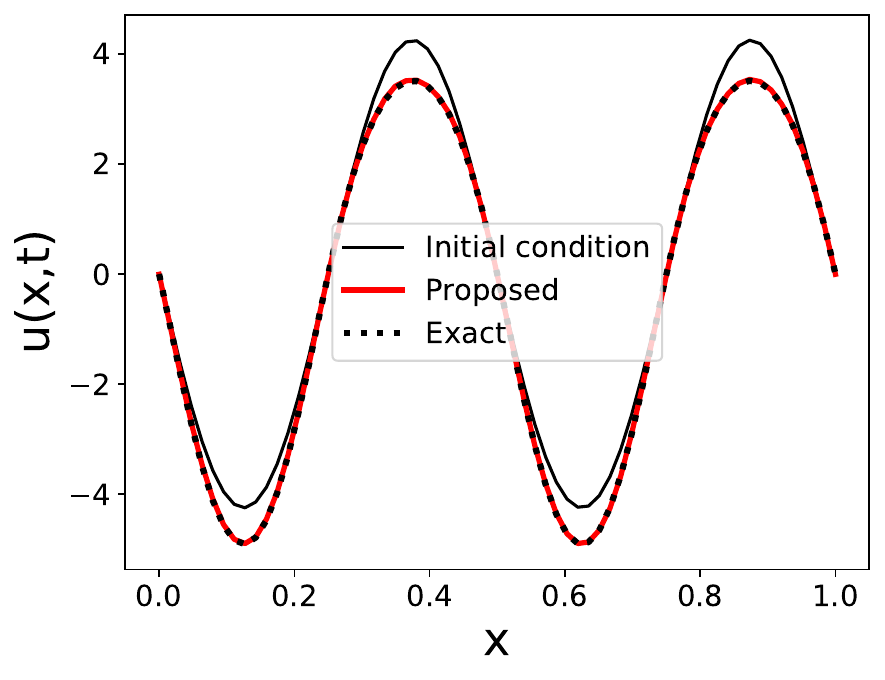}
      \caption{}
    \end{subfigure} &
    \begin{subfigure}{0.32\textwidth}
      \includegraphics[width=\linewidth]{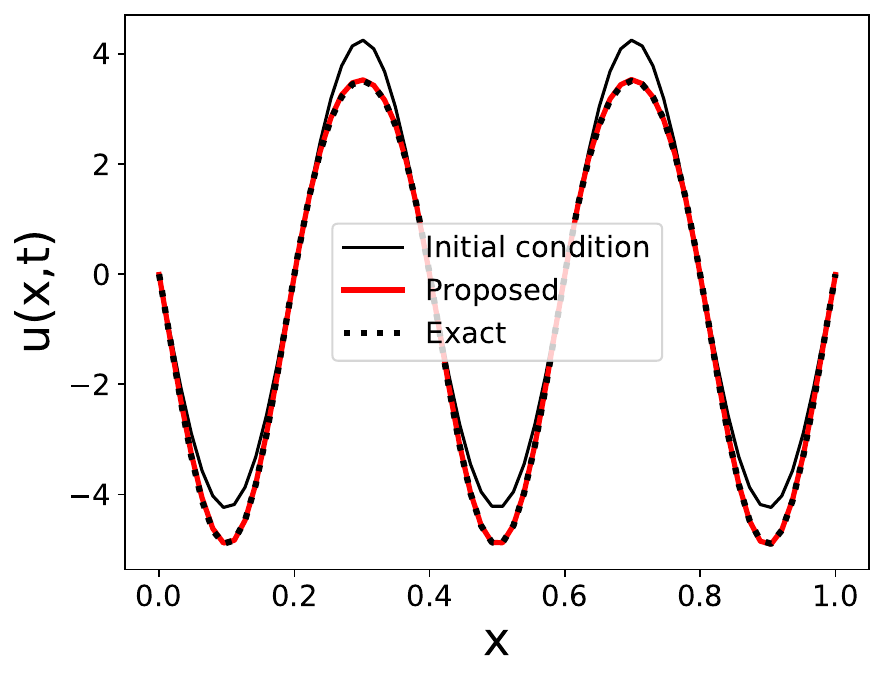}
      \caption{}
    \end{subfigure} \\
    
    \begin{subfigure}{0.32\textwidth}
      \includegraphics[width=\linewidth]{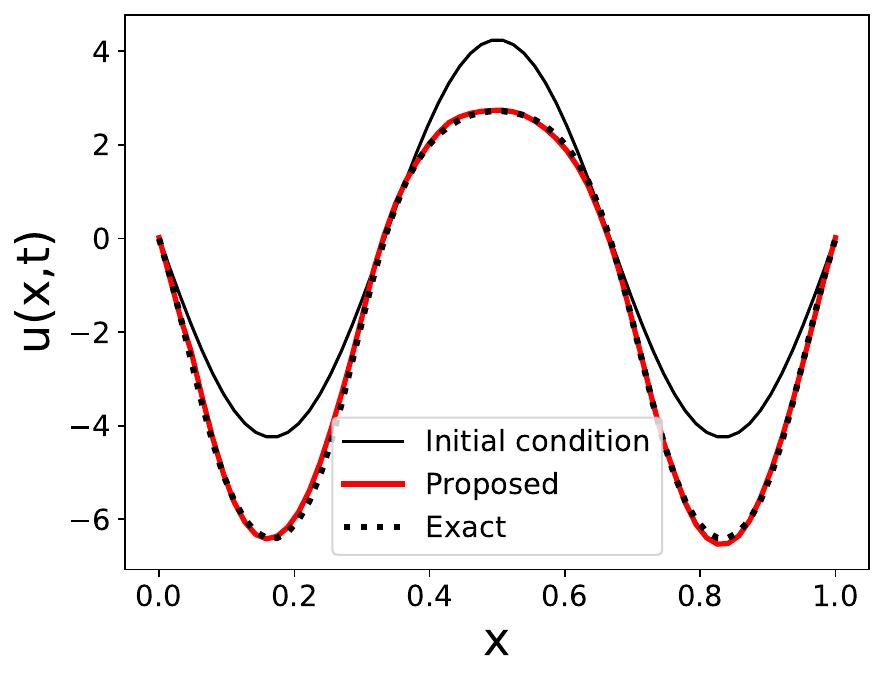}
      \caption{}
    \end{subfigure} &
    \begin{subfigure}{0.32\textwidth}
      \includegraphics[width=\linewidth]{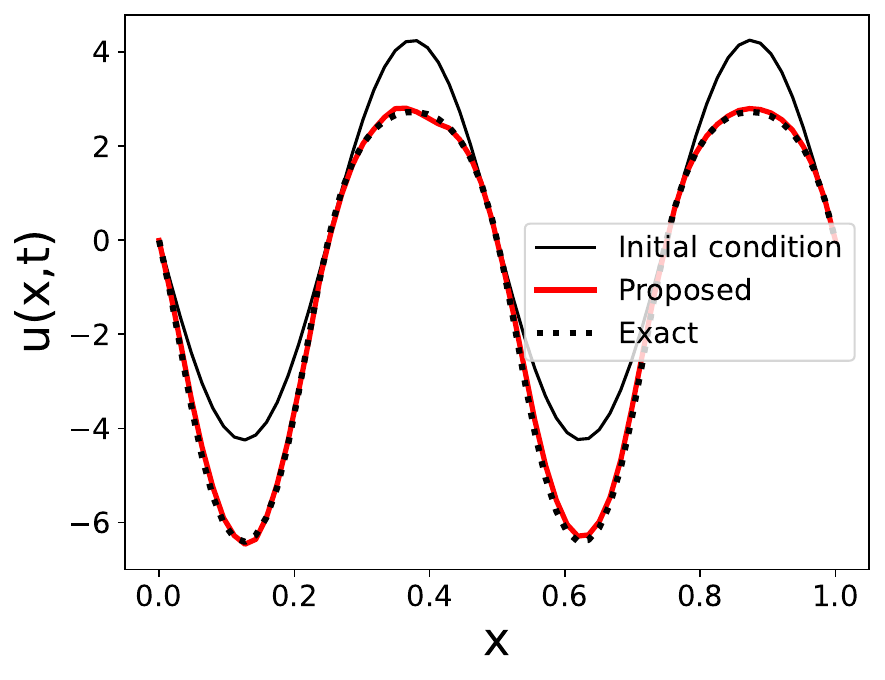}
      \caption{}
    \end{subfigure} &
    \begin{subfigure}{0.32\textwidth}
      \includegraphics[width=\linewidth]{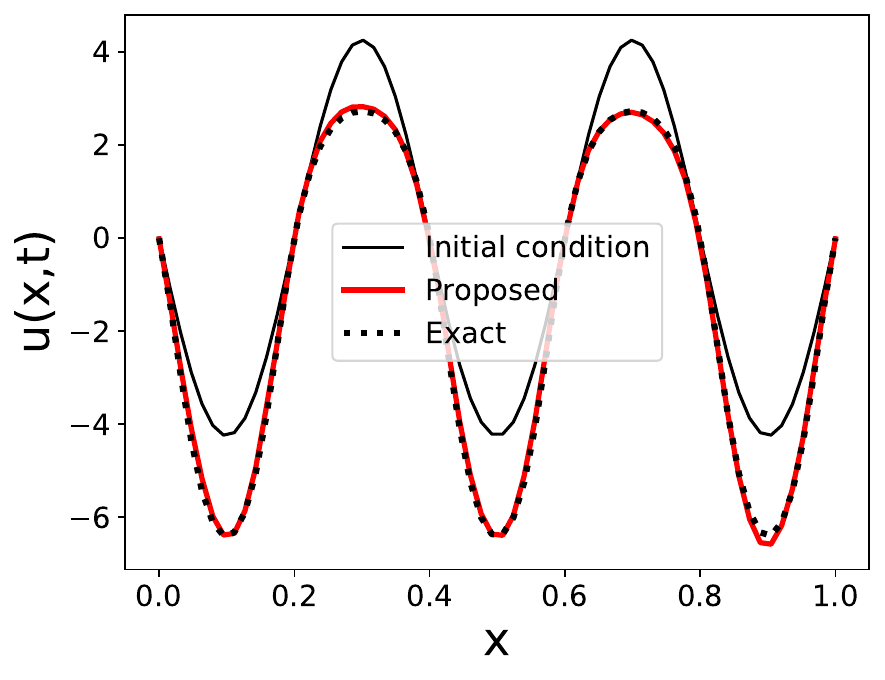}
      \caption{}
    \end{subfigure}
  \end{tabular}
  \caption{Comparison of predicted output $u(x,t)$ obtained using DPA-WNO, with the ground truth for Nagumo equation with missing advection term. The comparison is shown for three different initial conditions other than the trained samples. The top row corresponds to predictions within the training window ($t=48\Delta t$) while the bottom row corresponds to predictions outside the training window ($t=141\Delta t$)}
  \label{fig:Nagmcub_gen}
\end{figure}

The accurate predictions in Fig. \ref{fig:Nagmcub_pred} demonstrate the extrapolative ability of the proposed framework, extending to twice the training window. Fig. \ref{fig:Nagmcub_gen} showcases precise predictions for three distinct initial conditions not included in the training data, highlighting the model's strong generalization capability.

In Fig. \ref{fig:Nagmcub_pdfs}, a comparison of PDFs obtained from different models reveals the proposed model's excellent alignment with the ground truth within and beyond the training time window; this indicates DPA-WNO's robust predictive and generalization capabilities for diverse stochastic initial conditions. Conversely, the known physics model deviates noticeably from the ground truth due to the absence of the advection term, with increasing deviation outside the training window due to the accumulation of errors. The purely data-driven WNO model exhibits inferior performance compared to the proposed approach, displaying more significant offset in the extrapolation region and limited predictive ability far beyond the training time window.
The MSE and Hellinger distance reported om Table \ref{tab:MSEHD} further support the superiority of the proposed model. The proposed DPA-WNO model achieves an exceptionally low relative MSE of $0.1873$ and Hellinger distance of $0.056$, significantly outperforming the known physics model and purely data-driven WNO model. These results confirm the excellent performance of the proposed DPA-WNO model.

For reliability analysis, the kernel function as represented in Eq. \ref{eq:kernel}, incorporates the following parameter settings: a scaling factor $\alpha$ of $4$, a length scale value $l$ of $0.5$, and a periodicity value $p$ of 1. The limit-state function of the form shown in Eqs. \ref{eq:ls} and \eqref{eq:ls1}, and a threshold of $7$ units are considered. The proposed approach yields a reliability of 99.50\%, which exactly matches with that obtained from the true physics model (see Table \ref{tab:Reliability}).

\begin{figure}\label{fig}
\captionsetup[subfigure]{labelformat=empty}
  \centering
  \begin{tabular}{cc}
    \begin{subfigure}{0.48\textwidth}
      \includegraphics[width=\linewidth]{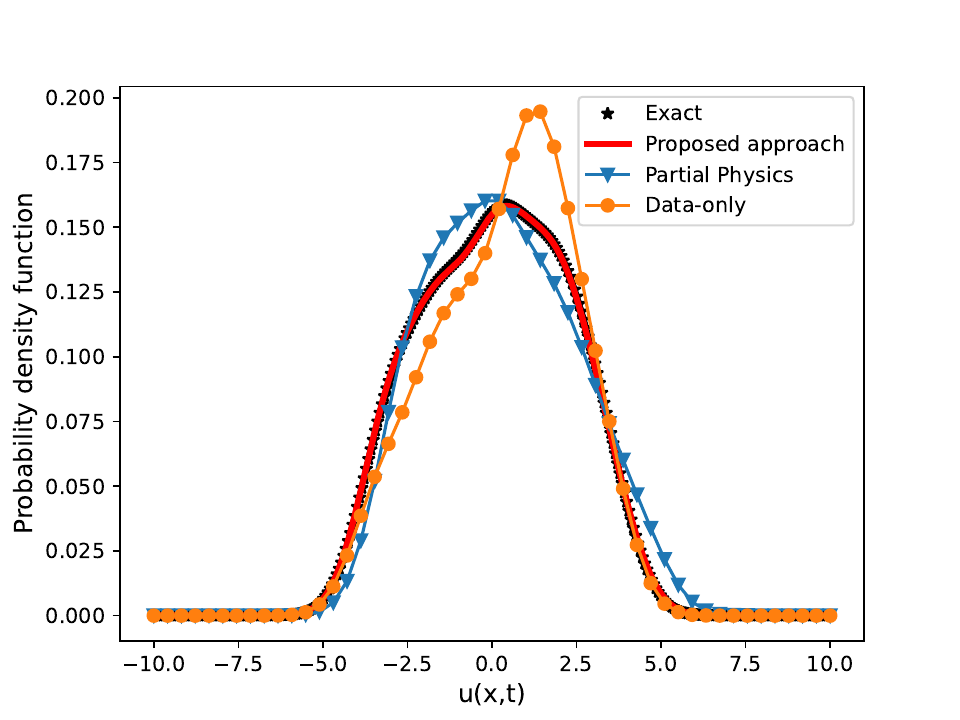}
      \caption{PDF when $t < t_{train}$}
      \label{fig:Nagmcub_pdfs_1}
    \end{subfigure} &
    \begin{subfigure}{0.48\textwidth}
      \includegraphics[width=\linewidth]{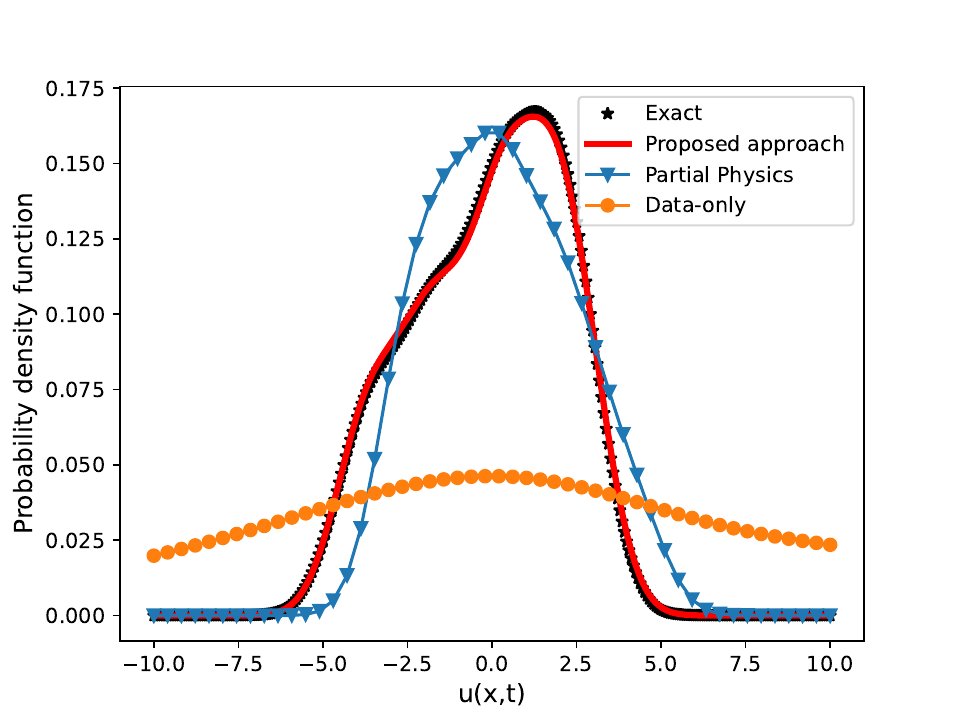}
      \caption{PDF when $t > t_{train}$}
      \label{fig:Nagmcub_pdfs_2}
    \end{subfigure}
  \end{tabular}
  \caption{Comparison of the probability density functions (PDFs) of the output $u(x=0.1093,t=48\Delta t)$ (left) and $u(x=0.1093,t=101\Delta t)$ (right) for Nagumo equation from four different models: DPA-WNO (Proposed), Known physics (Partial physics), data-based WNO (Data only), and the ground truth (Exact). The results correspond to the case where only diffusion term in the known in the known-physics model}
  \label{fig:Nagmcub_pdfs}
\end{figure}

\subsubsection{Missing diffusion term in the known physics}
Next, we consider the case where the diffusion term is missing, and the known physics is of the following form:
\begin{equation}
       \frac{\partial u}{\partial t} = u(1-u)(u-\alpha), \quad x \in (0,1), t \in (0,T)
\end{equation}
\begin{figure}
\captionsetup[subfigure]{labelformat=empty}
  \centering
  \begin{tabular}{ccc}
    \begin{subfigure}{0.32\textwidth}
      \includegraphics[width=\linewidth]{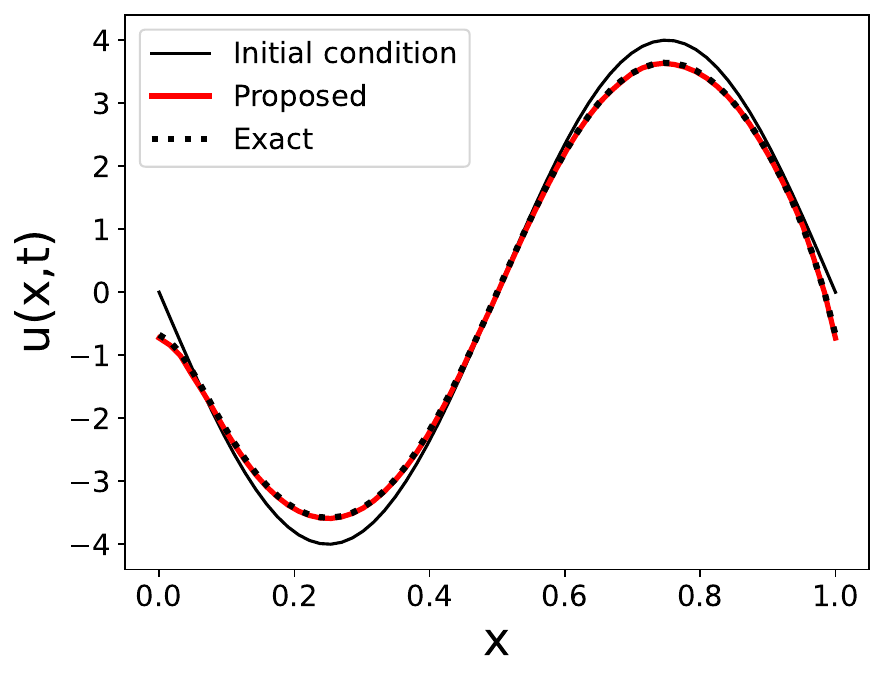}
      \caption{}
    \end{subfigure} &
    \begin{subfigure}{0.32\textwidth}
      \includegraphics[width=\linewidth]{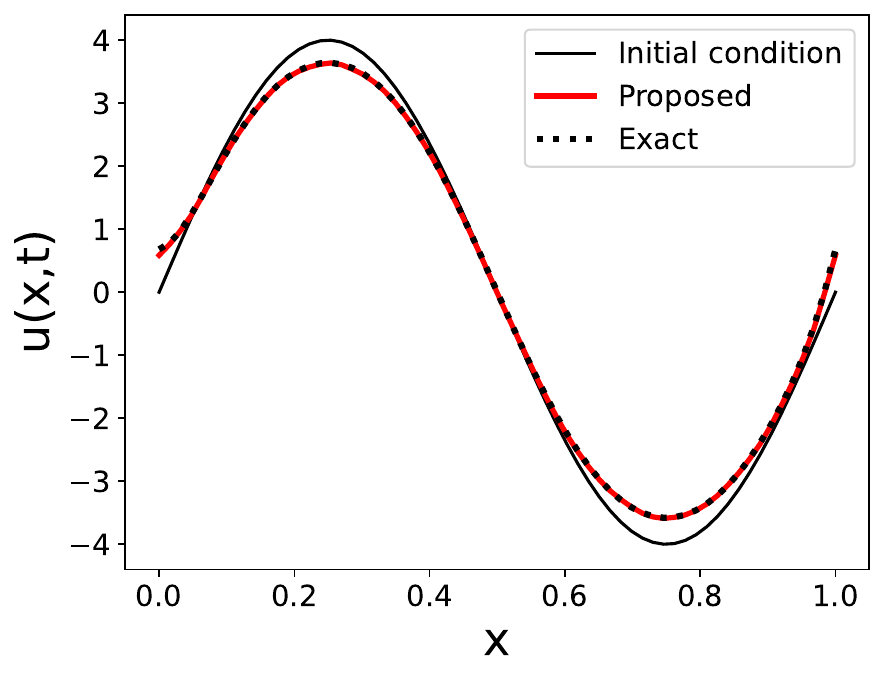}
      \caption{}
    \end{subfigure} &
    \begin{subfigure}{0.32\textwidth}
      \includegraphics[width=\linewidth]{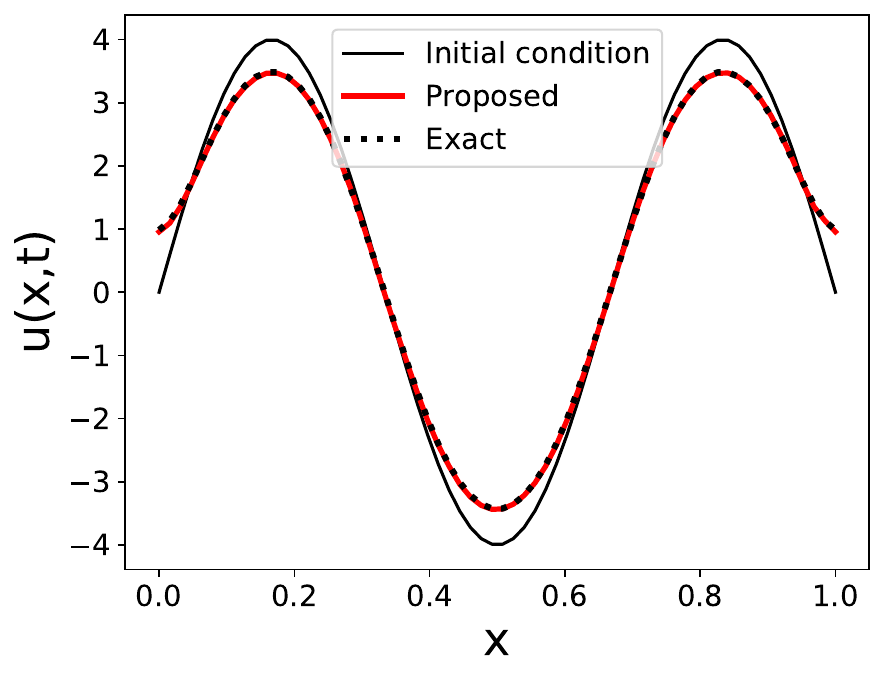}
      \caption{}
    \end{subfigure} \\
    
    \begin{subfigure}{0.32\textwidth}
      \includegraphics[width=\linewidth]{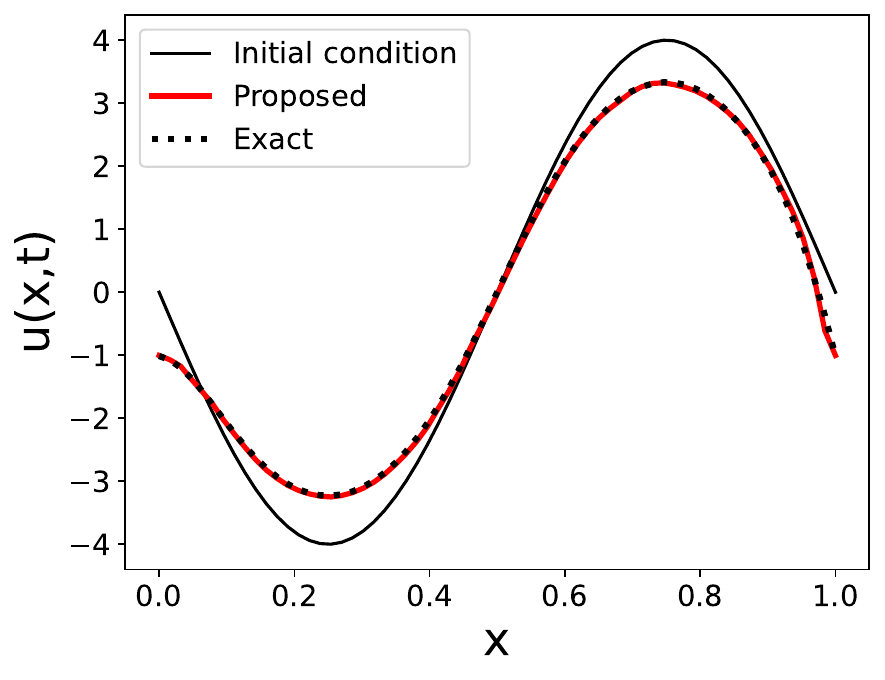}
      \caption{}
    \end{subfigure} &
    \begin{subfigure}{0.32\textwidth}
      \includegraphics[width=\linewidth]{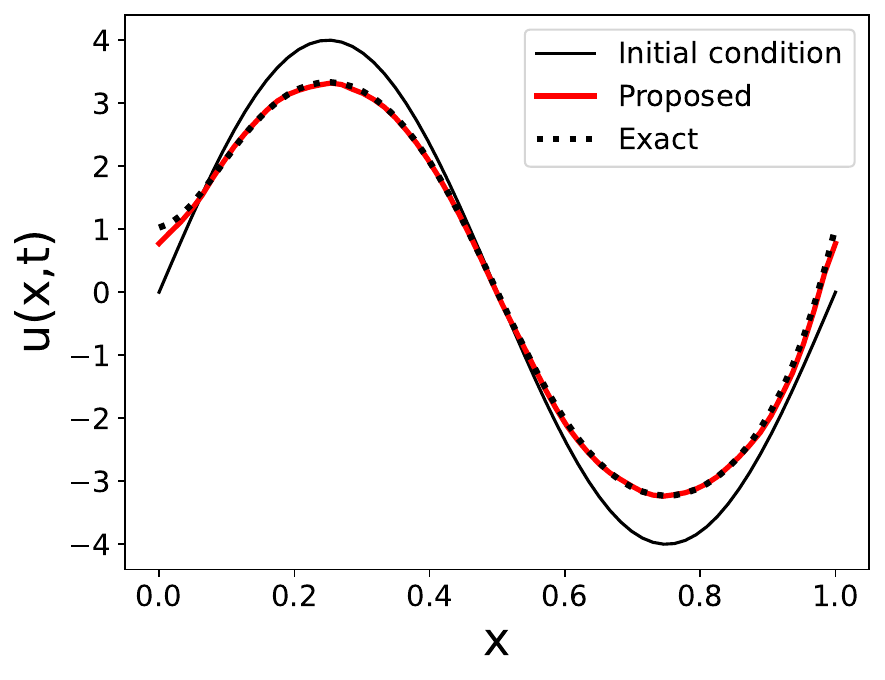}
      \caption{}
    \end{subfigure} &
    \begin{subfigure}{0.32\textwidth}
      \includegraphics[width=\linewidth]{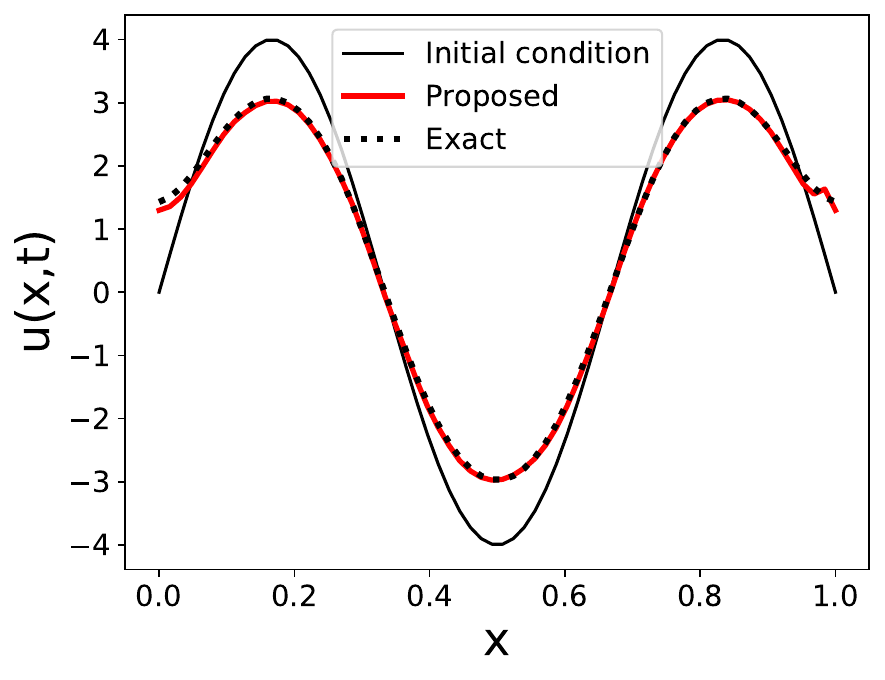}
      \caption{}
    \end{subfigure}
  \end{tabular}
  \caption{Comparison of predicted output $u(x,t)$ obtained using DPA-WNO, with the ground truth for Nagumo equation with missing diffusion term. The comparison is shown for three different initial conditions taken from the trained samples. The top row corresponds to predictions within the training window ($t=48\Delta t$) while the bottom row corresponds to predictions outside the training window ($t=101\Delta t$)}
  \label{fig:Nagmdiff_pred}
\end{figure}

\noindent The initial and boundary conditions remain same as the previous case. We have considered $\epsilon=0.2$ and $\alpha=-0.5$ to emphasize the disparity between partial physics and complete physics. The objective is to employ the proposed DPA-WNO as a surrogate to learn the temporal evolution of the solution field, quantify the propagation of the input uncertainty to the output response, and compute the reliability of the system.

\begin{figure}[t]
\captionsetup[subfigure]{labelformat=empty}
  \centering
  \begin{tabular}{ccc}
    \begin{subfigure}{0.32\textwidth}
      \includegraphics[width=\linewidth]{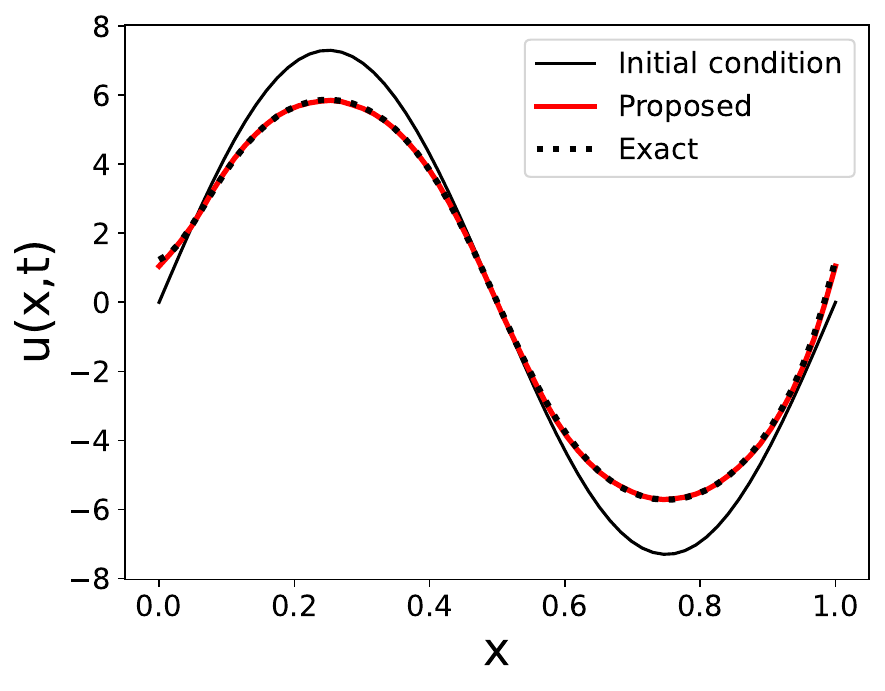}
      \caption{}
    \end{subfigure} &
    \begin{subfigure}{0.32\textwidth}
      \includegraphics[width=\linewidth]{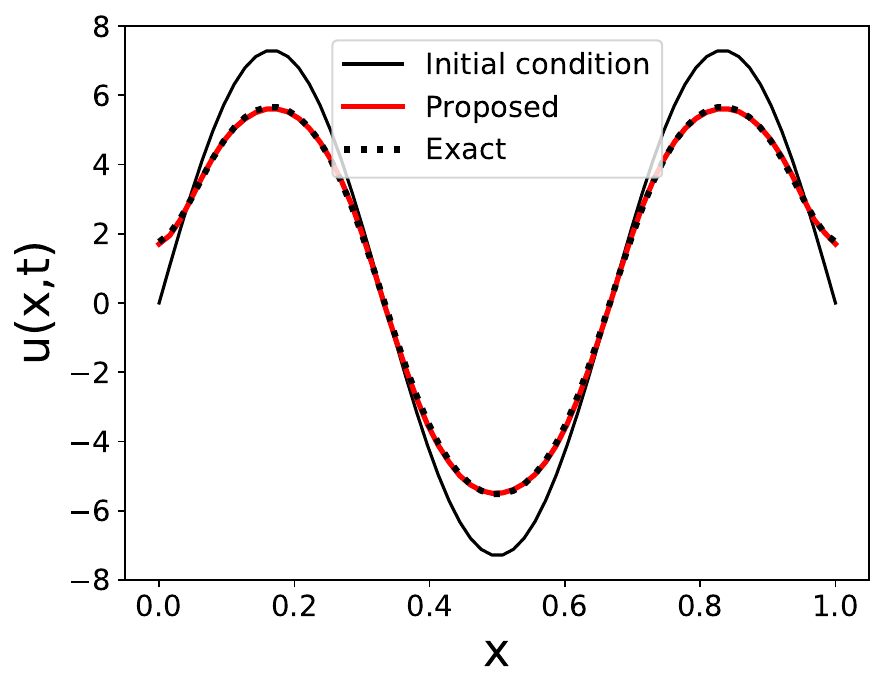}
      \caption{}
    \end{subfigure} &
    \begin{subfigure}{0.32\textwidth}
      \includegraphics[width=\linewidth]{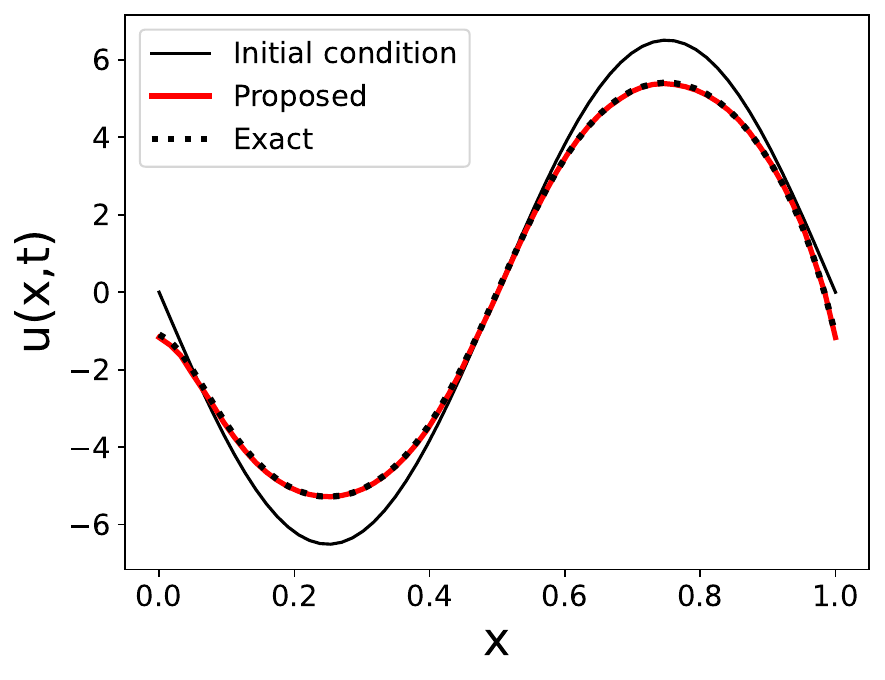}
      \caption{}
    \end{subfigure} \\
    
    \begin{subfigure}{0.32\textwidth}
      \includegraphics[width=\linewidth]{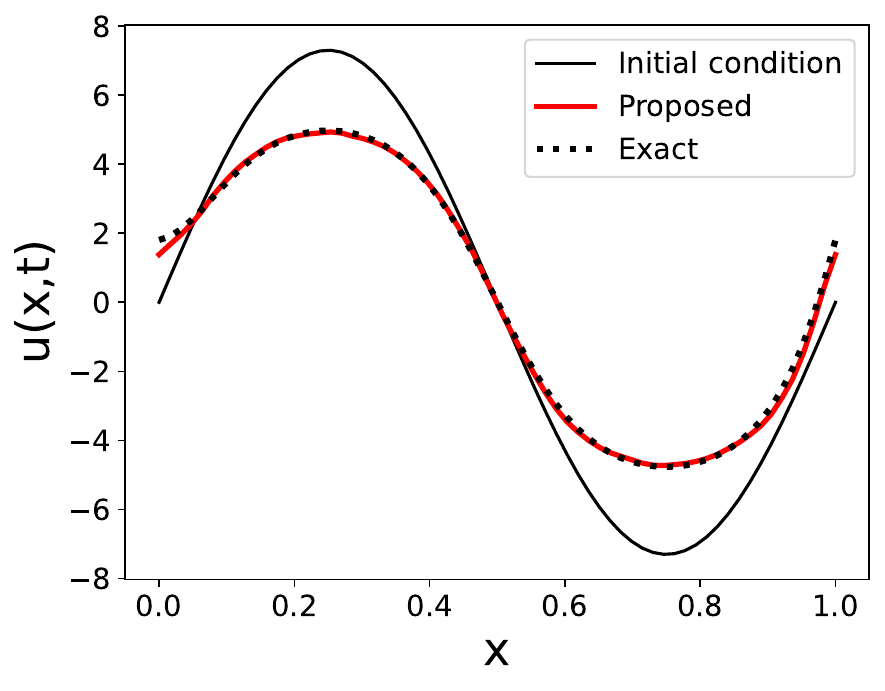}
      \caption{}
    \end{subfigure} &
    \begin{subfigure}{0.32\textwidth}
      \includegraphics[width=\linewidth]{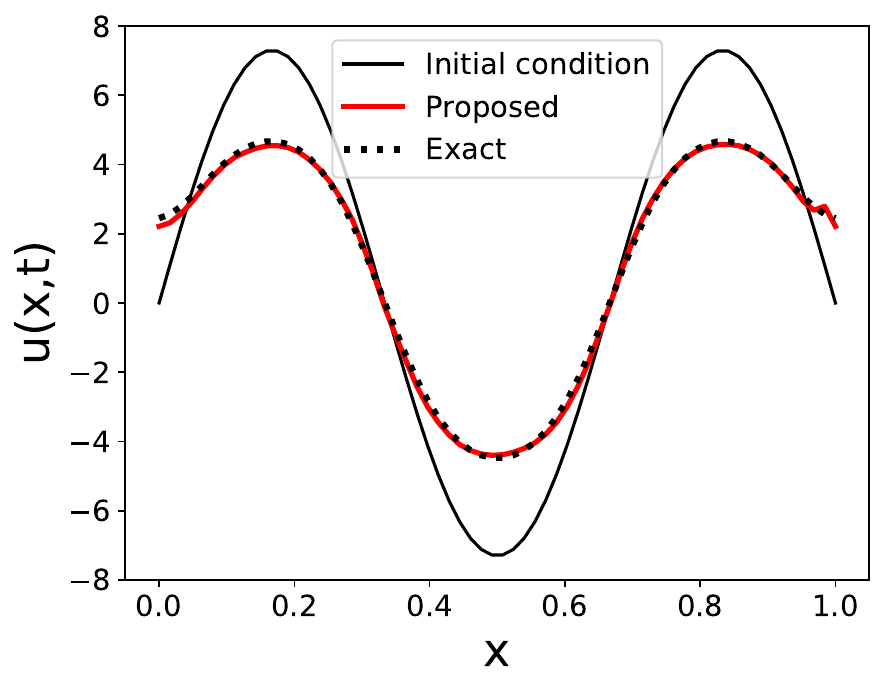}
      \caption{}
    \end{subfigure} &
    \begin{subfigure}{0.32\textwidth}
      \includegraphics[width=\linewidth]{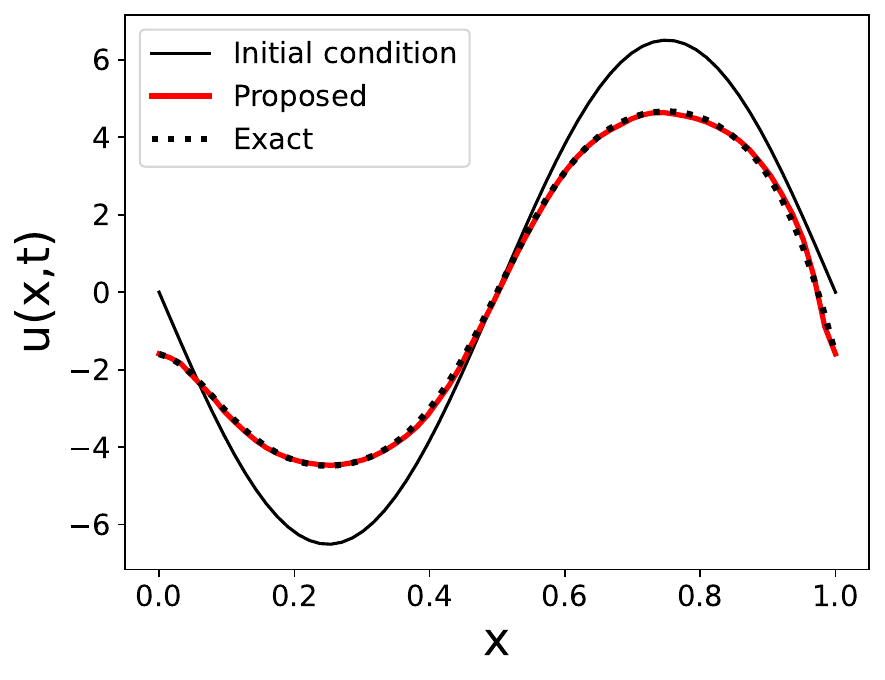}
      \caption{}
    \end{subfigure}
  \end{tabular}
  \caption{Comparison of predicted output $u(x,t)$ obtained using DPA-WNO, with the ground truth for Nagumo equation with missing diffusion term. The comparison is shown for three different initial conditions other than the trained samples. The top row corresponds to predictions within the training window ($t=48\Delta t$) while the bottom row corresponds to predictions outside the training window ($t=101\Delta t$)}
  \label{fig:Nagmdiff_gen}
\end{figure}

The results depicted in Fig. \ref{fig:Nagmdiff_pred} demonstrate accurate predictions that extend beyond the training window. Fig. \ref{fig:Nagmcub_gen} showcases precise predictions for diverse initial conditions not present in the training sample, indicating the model's strong generalization ability. In Fig. \ref{fig:Nagmdiff_pdfs}, the proposed model exhibits good alignment with the ground truth in capturing the uncertainty of the response; the known physics model, on the other hand, deviates due to missing terms. Data-driven WNO model displays limited accuracy specifically outside the training window. The DPA-WNO model yields highly accurate results as illustrated by the low MSE ($0.0042$) and low mean Hellinger distance ($0.00105$). Physics-only and data-only models yield erroneous results as shown in Table \ref{tab:MSEHD}. 

For the reliability analysis in this case, the same setup as the previous case is considered. The proposed DPA-WNO method yielded a reliability value of 99.4\%, which precisely aligns with the ground truth reliability as presented in Table \ref{tab:Reliability}.

\begin{figure}
\captionsetup[subfigure]{labelformat=empty}
  \centering
  \begin{tabular}{cc}
    \begin{subfigure}{0.48\textwidth}
      \includegraphics[width=\linewidth]{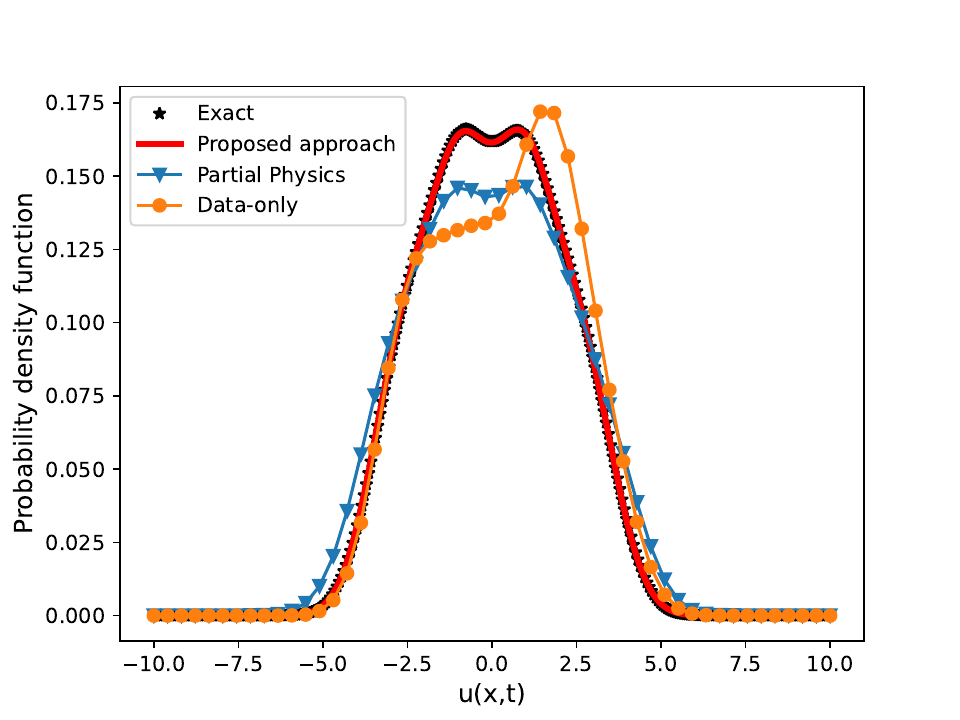}
      \caption{PDF when $t < t_{train}$}
      \label{fig: Nagmdiff_pdfs_1}
    \end{subfigure} &
    \begin{subfigure}{0.48\textwidth}
      \includegraphics[width=\linewidth]{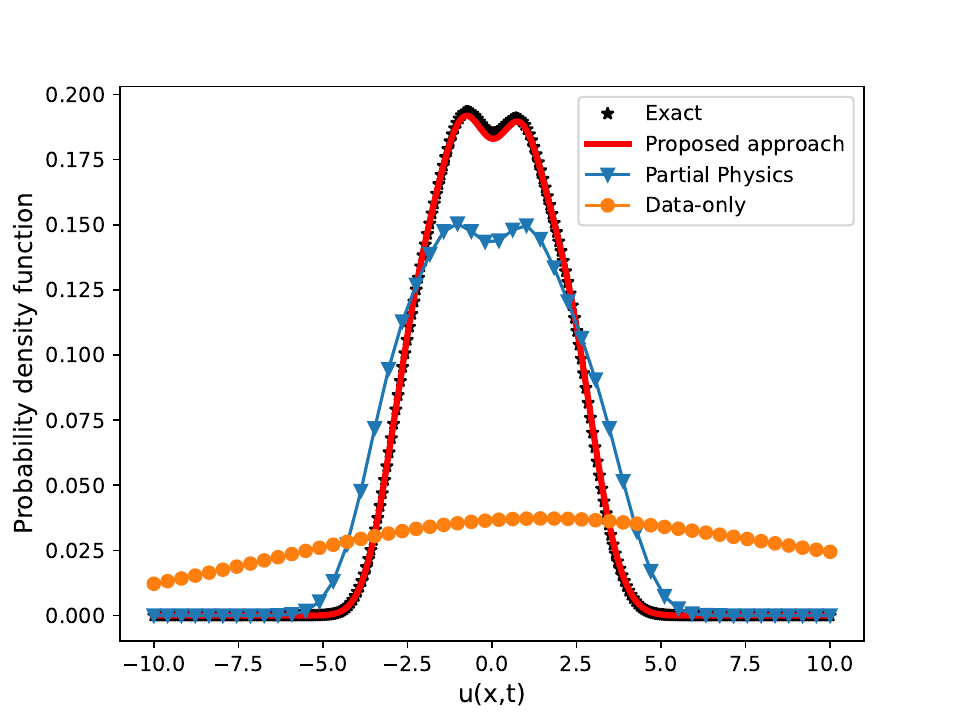}
      \caption{PDF when $t > t_{train}$}
      \label{fig:Nagmdiff_pdfs_2}
    \end{subfigure}
  \end{tabular}
  \caption{Comparison of the probability density functions (PDFs) of the output $u(x=0.8593,t=48\Delta t)$ (left) and $u(x=0.8593,t=99\Delta t)$ (right) for Nagurmo equation from four different models: DPA-WNO (Proposed), Known physics (Partial physics), data-based WNO (Data only), and the ground truth (Exact). The results correspond to the case where diffusion term is missing in the known physics}
  \label{fig:Nagmdiff_pdfs}
\end{figure}

\subsection{Example 3: Allen-Cahn equation}\label{subsec:eg3}

In this example, we consider the Allen-Cahn equation. This equation is utilized to describe the phenomenon of reaction-diffusion and has application in various areas including phase separation in multi-component alloys, chemical reactions, and crystal growth. Here, we have considered the Allen-Cahn equation with periodic boundary conditions, 
\begin{equation}\label{eq:AC_comp}
    \begin{aligned}
  &\frac{\partial u}{\partial t} - \Gamma \frac{\partial^2 u}{\partial x^2} + 5u^3 - 5u = 0, \quad x \in (-1,1), t \in (0,T)\\
   &u(x=-1,t) = u(x=1,t),\quad x \in (-1,1), t \in (0,T)\\
   &u_x(x=-1,t) = u_x(x=1,t),\quad x \in (-1,1), t \in (0,T)\\
   &u(x,t=0) = u_0(x), \quad x \in (-1,1)\\ 
\end{aligned}
\end{equation}
where $\Gamma$ is the diffusion coefficient.

\subsubsection{Only diffusion term in the known physics} 
We consider a case, where the known physics is of the following form:
\begin{equation}\label{eq:AC_mcub}
\begin{aligned}
 &\frac{\partial u}{\partial t} - \Gamma \frac{\partial^2 u}{\partial x^2} = 0, \quad x \in (-1,1), t \in (0,T)
\end{aligned}
\end{equation}
where, the $\Gamma = 0.1$. Boundary and initial conditions are kept same as in Eq.\ref{eq:AC_comp}. We have synthetically generated the high-fidelity data $\mathcal{D} = \left\{\bm{u}_{0,1:N_x}^{(i)},\bm{u}^{(i)}_{1:N_t,1:N_x} \right\}_{i=1}^N$, by solving the complete Allan Cahn equation. In this case, the spacial resolution is taken as $N_x = 112$ and $\Delta t = 0.0001$ sec. Response from first 50 times steps are considered during training. Similar to previous examples, ADAM optimizer with a constant learning rate of 0.007 is considered. Adaptive training as described in Sec.\ref{subsubsec:training} is employed.
We have considered $N = 32$ sample solutions considering different initial conditions of the form $\bm{u}_0(x) = \alpha cos(0.5\zeta \pi x)$, where  $\alpha$ $\in$ $\left\{-4,-3,-2...,-1,1,2...4\right\}$ and $\zeta \in \left\{1,3,5,7\right\}$. 

In the testing phase, 100 initial conditions were chosen for evaluation. These initial conditions followed the same form as $\bm{u}_0(x) = \alpha \cos(0.5 \zeta \pi x)$, but the parameter $\alpha$ was randomly sampled from a uniform distribution $\mathcal{U}(-10,10)$, and the parameter $\zeta$ remained constant and took values $\left\{1, 3, 5, 7\right\}$.
\begin{figure}[t]
\captionsetup[subfigure]{labelformat=empty}
  \centering
  \begin{tabular}{ccc}
    \begin{subfigure}{0.32\textwidth}
      \includegraphics[width=\linewidth]{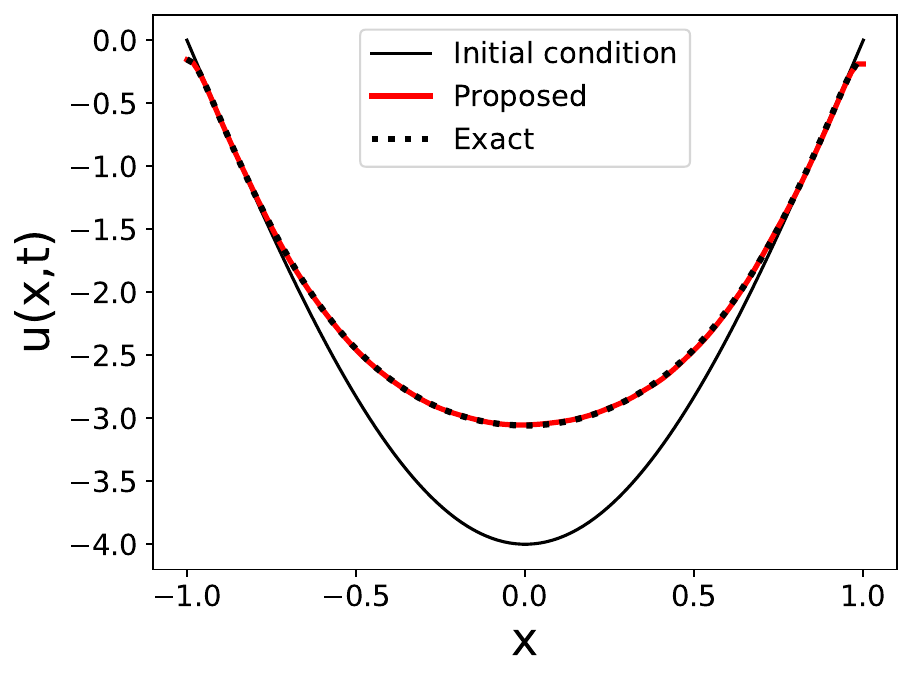}
      \caption{}
    \end{subfigure} &
    \begin{subfigure}{0.32\textwidth}
      \includegraphics[width=\linewidth]{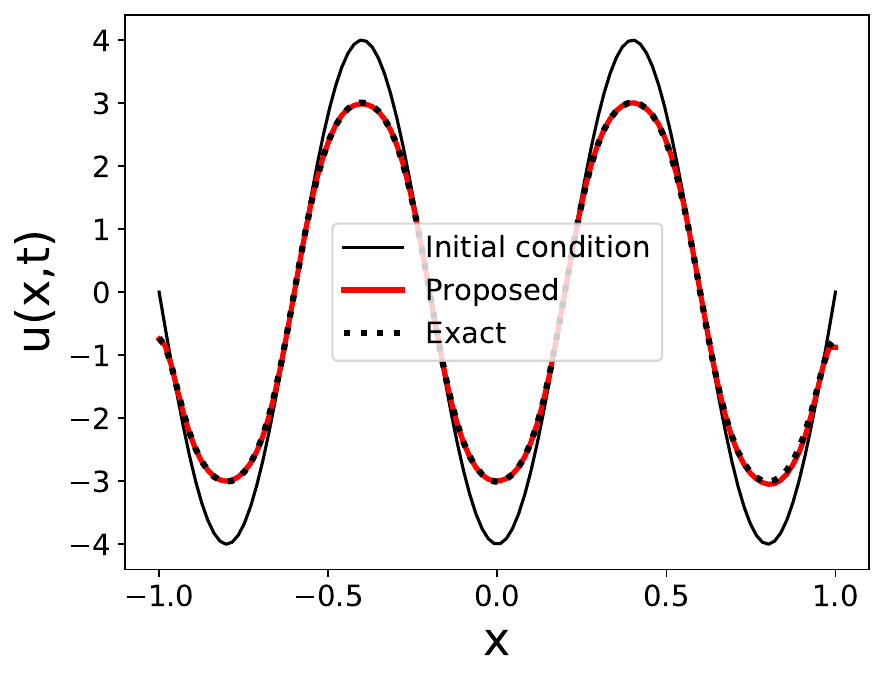}
      \caption{}
    \end{subfigure} &
    \begin{subfigure}{0.32\textwidth}
      \includegraphics[width=\linewidth]{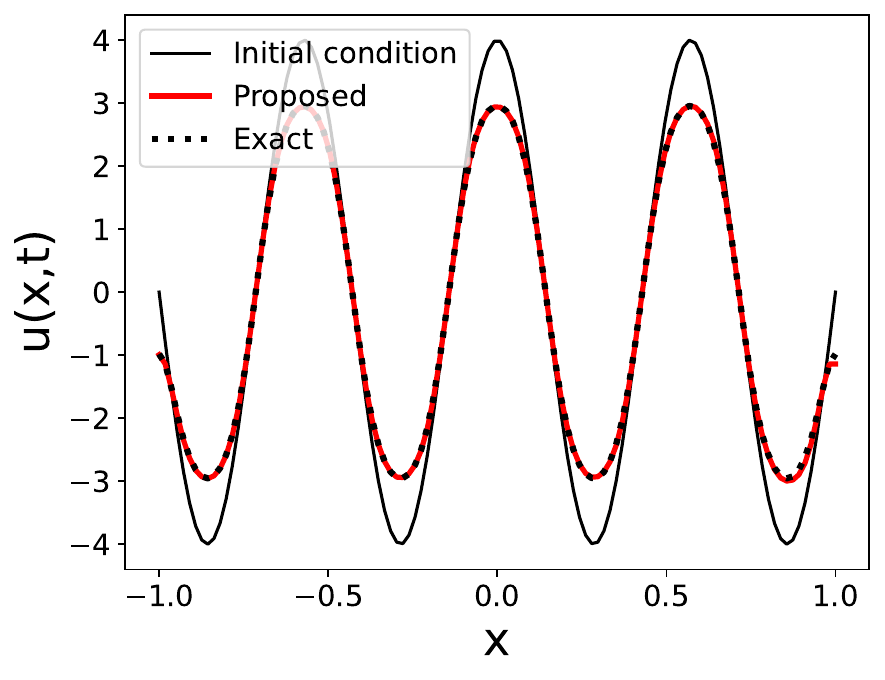}
      \caption{}
    \end{subfigure} \\
    
    \begin{subfigure}{0.32\textwidth}
      \includegraphics[width=\linewidth]{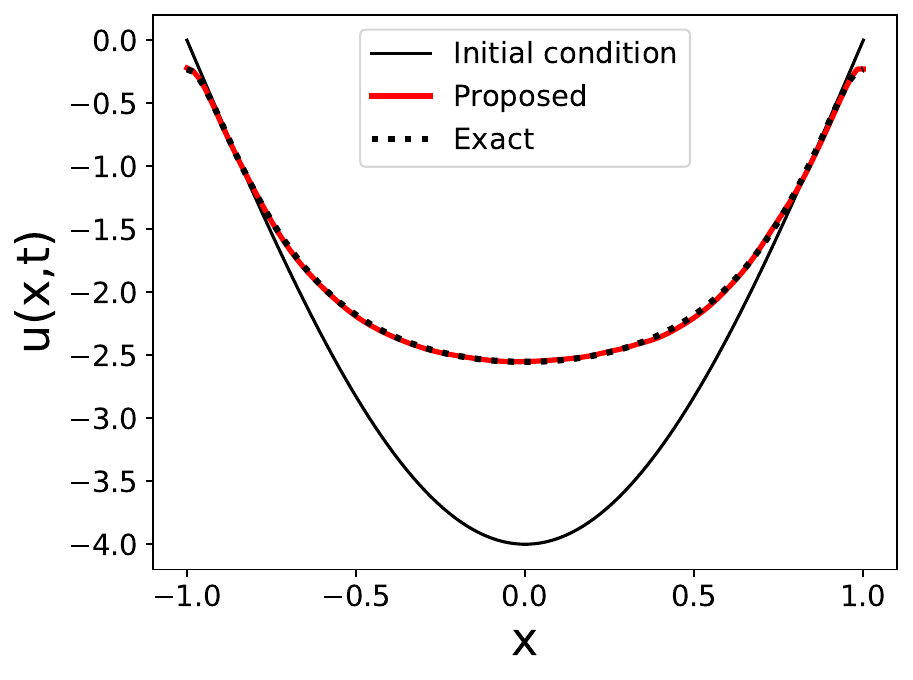}
      \caption{}
    \end{subfigure} &
    \begin{subfigure}{0.32\textwidth}
      \includegraphics[width=\linewidth]{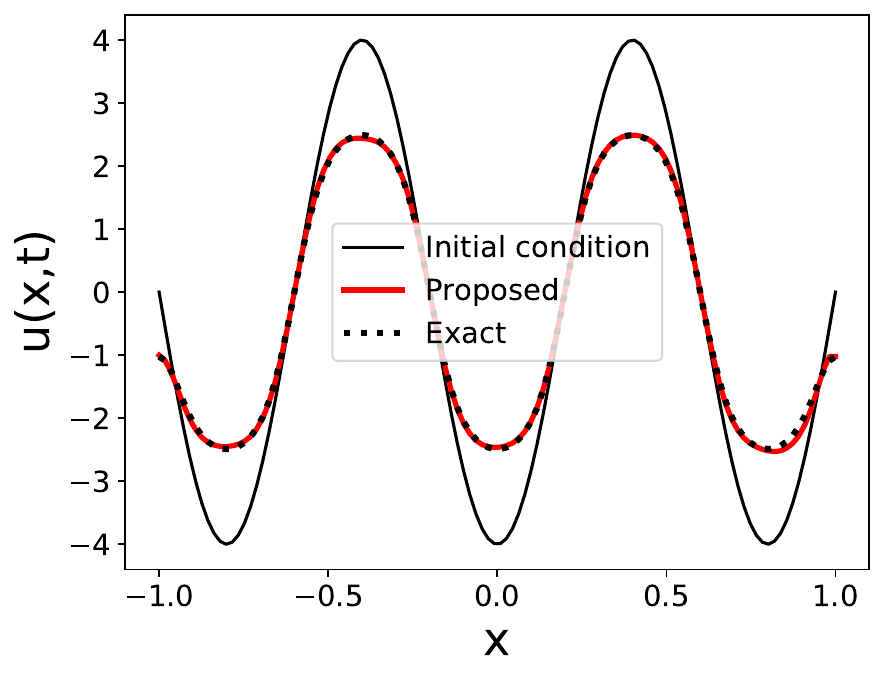}
      \caption{}
    \end{subfigure} &
    \begin{subfigure}{0.32\textwidth}
      \includegraphics[width=\linewidth]{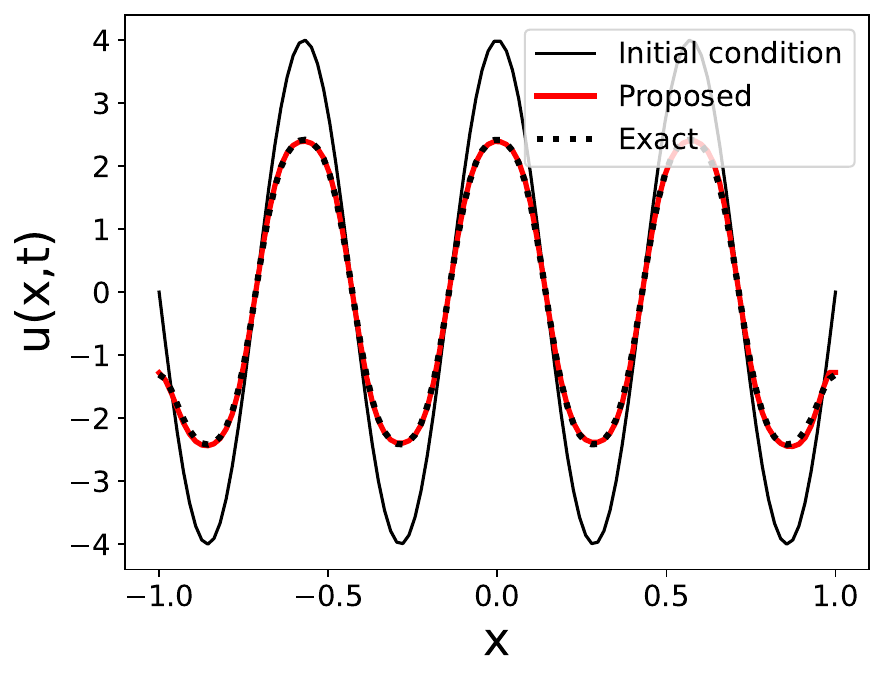}
      \caption{}
    \end{subfigure}
  \end{tabular}
  \caption{Comparison of predicted output $u(x,t)$ obtained using DPA-WNO, with the ground truth for Allen-Cahn equation with missing advection term. The comparison is shown for three different initial conditions taken from the trained samples. The top row corresponds to predictions within the training window ($t=48\Delta t$) while the bottom row corresponds to predictions outside the training window ($t=101\Delta t$)}
  \label{fig:ACmcub_pred}
\end{figure}

\begin{figure}[t]
\captionsetup[subfigure]{labelformat=empty}
  \centering
  \begin{tabular}{ccc}
    \begin{subfigure}{0.32\textwidth}
      \includegraphics[width=\linewidth]{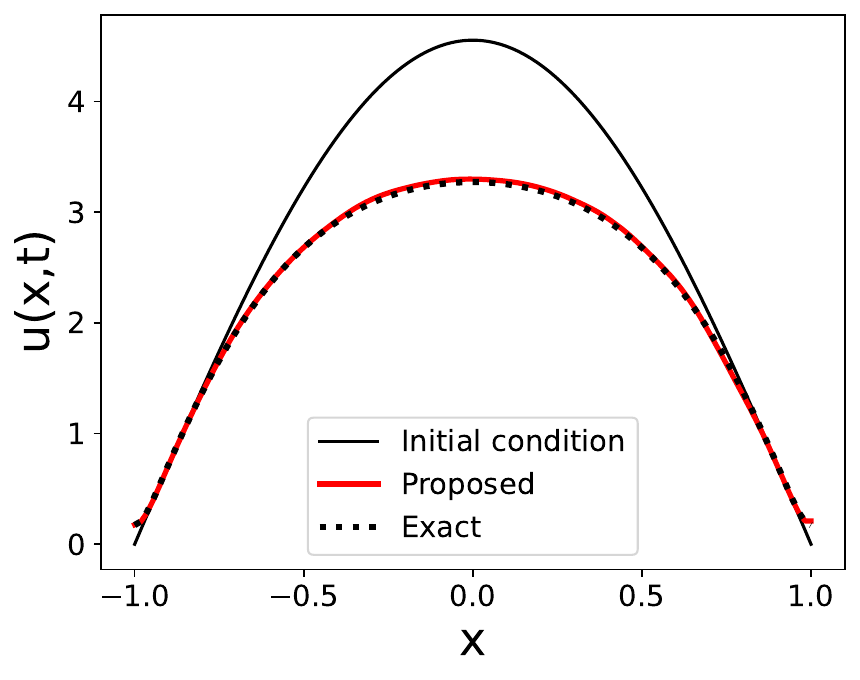}
      \caption{}
    \end{subfigure} &
    \begin{subfigure}{0.32\textwidth}
      \includegraphics[width=\linewidth]{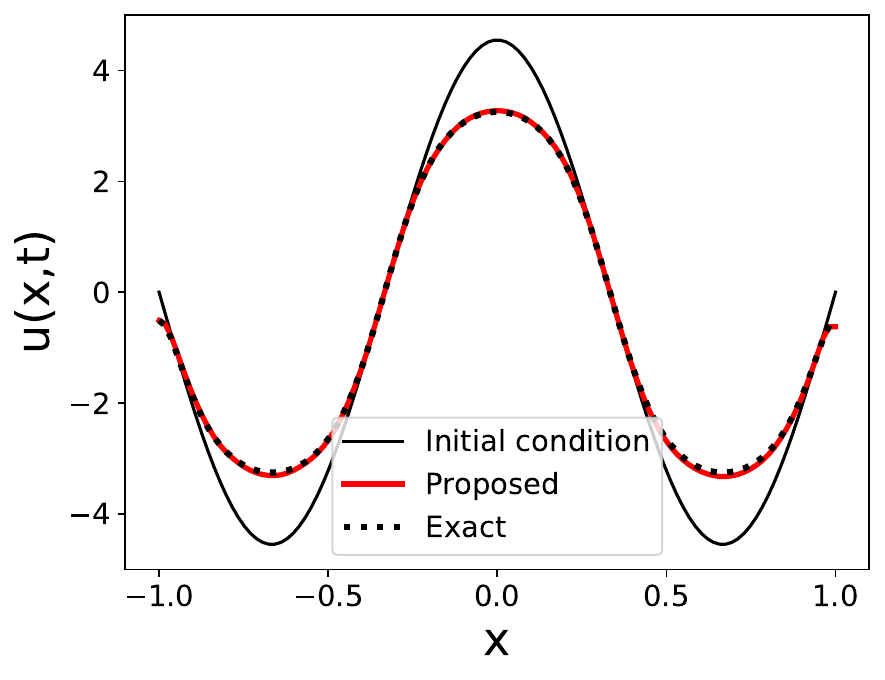}
      \caption{}
    \end{subfigure} &
    \begin{subfigure}{0.32\textwidth}
      \includegraphics[width=\linewidth]{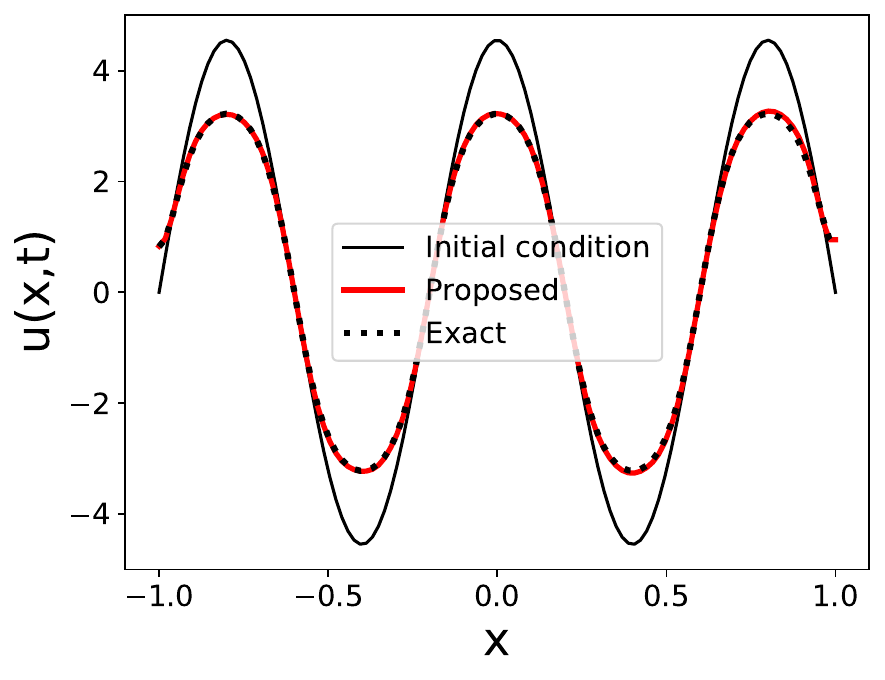}
      \caption{}
    \end{subfigure} \\
    
    \begin{subfigure}{0.32\textwidth}
      \includegraphics[width=\linewidth]{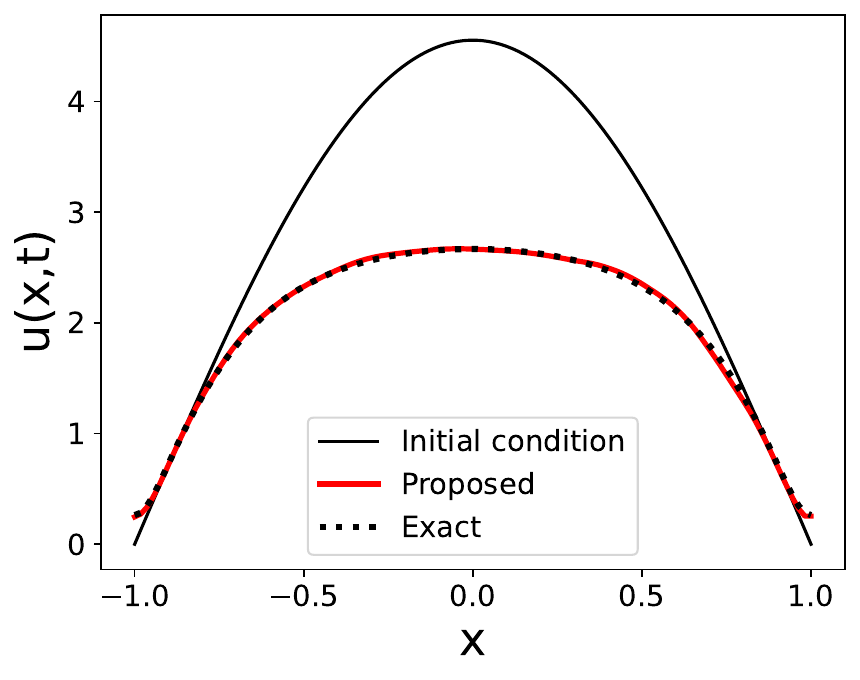}
      \caption{}
    \end{subfigure} &
    \begin{subfigure}{0.32\textwidth}
      \includegraphics[width=\linewidth]{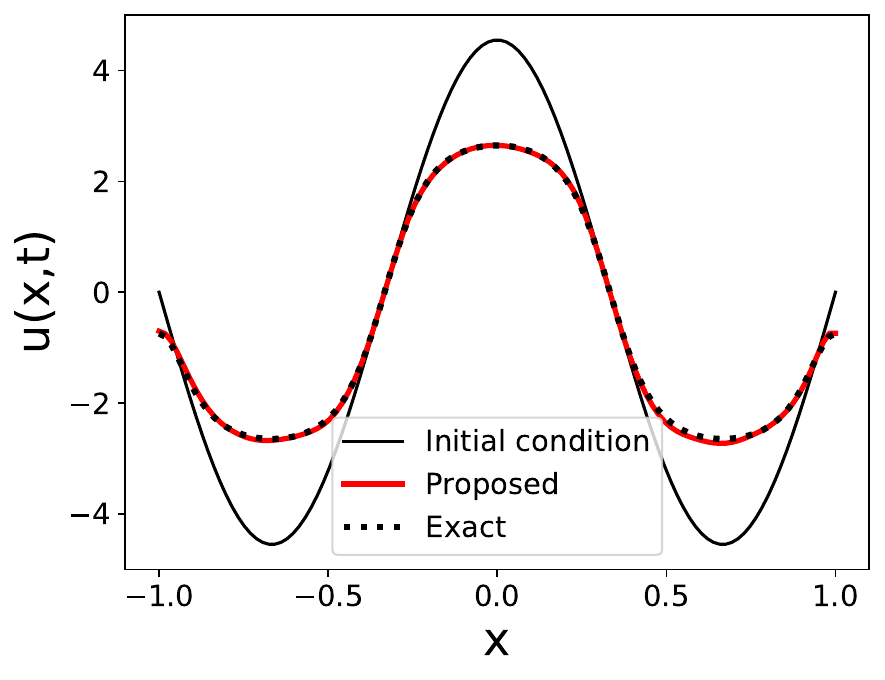}
      \caption{}
    \end{subfigure} &
    \begin{subfigure}{0.32\textwidth}
      \includegraphics[width=\linewidth]{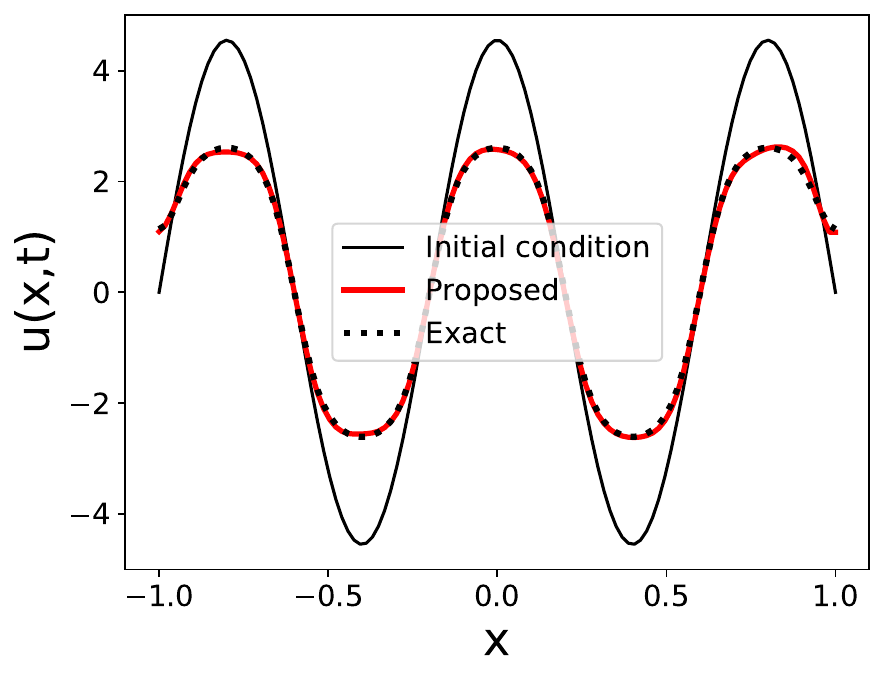}
      \caption{}
    \end{subfigure}
  \end{tabular}
  \caption{Comparison of predicted output $u(x,t)$ obtained using DPA-WNO, with the ground truth for Allen-Cahn equation with missing advection term. The comparison is shown for three different initial conditions other than the trained samples. The top row corresponds to predictions within the training window ($t=48\Delta t$) while the bottom row corresponds to predictions outside the training window ($t=101\Delta t$)}
  \label{fig:ACmcub_gen}
\end{figure}

The proposed framework demonstrates its extrapolative ability by accurately predicting beyond the training window, as shown in Fig. \ref{fig:ACmcub_pred}. Additionally, Fig. \ref{fig:ACmcub_gen} presents precise predictions for three distinct initial conditions not included in the training data, highlighting the model's generalization capability.

Fig. \ref{fig:ACmcub_pdfs} provides a comparison of the PDFs obtained using the proposed model with that of the other models (data-only and physics-only). The proposed model shows excellent alignment within and beyond the training time window, indicating its robust predictive and generalization capabilities for a diverse set of stochastic initial conditions. On the other hand, the known physics model exhibits noticeable deviation from the ground truth, particularly outside the training window due to accumulated errors. The purely data-driven WNO model yields inferior results compared to the proposed approach, with a more significant offset in the extrapolation region, suggesting its limited predictive ability far beyond the training time window. Further, the proposed DPA-WNO model achieves a low MSE of $0.1583$, significantly outperforming the known physics model and purely data-driven WNO model. It also shows superior performance in terms of mean Hellinger distance (Table \ref{tab:MSEHD}).

\begin{figure}[ht!]
\captionsetup[subfigure]{labelformat=empty}
  \centering
  \begin{tabular}{cc}
    \begin{subfigure}{0.48\textwidth}
      \includegraphics[width=\linewidth]{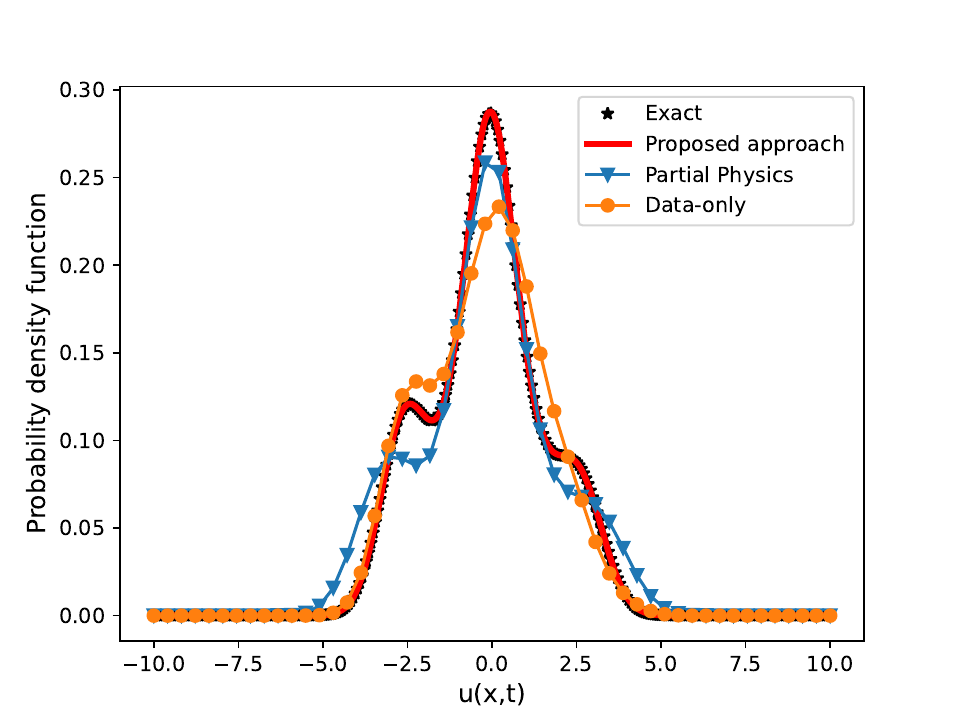}
      \caption{PDF when $t < t_{train}$}
    \end{subfigure} &
    \begin{subfigure}{0.48\textwidth}
      \includegraphics[width=\linewidth]{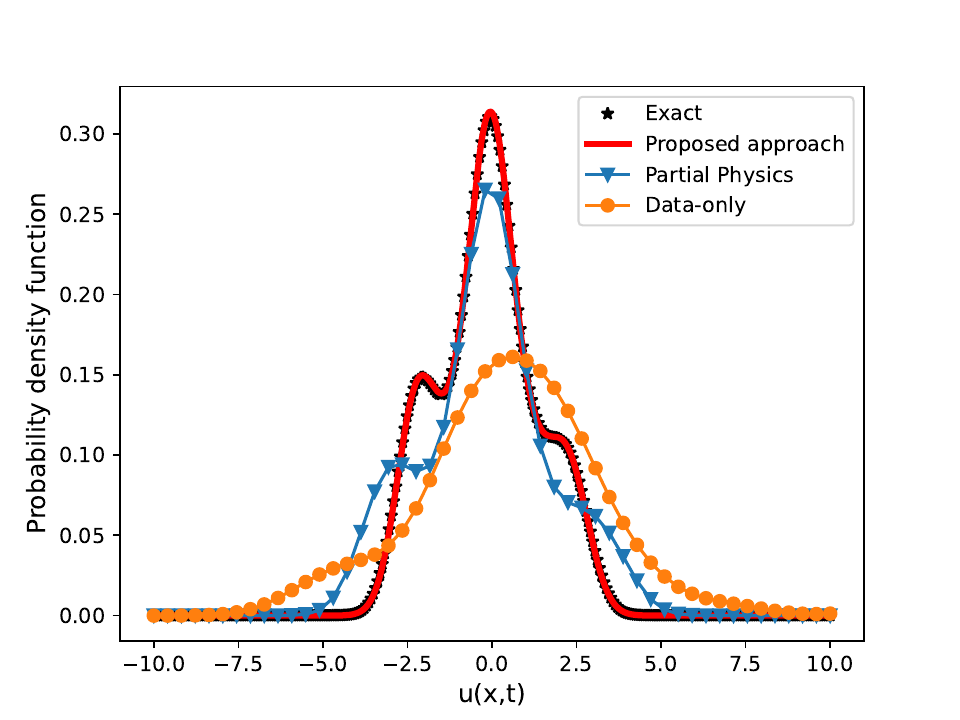}
      \caption{PDF when $t > t_{train}$}
    \end{subfigure}
  \end{tabular}
  \caption{Comparison of the probability density functions (PDFs) of the output $u(x=-0.7857,t=48\Delta t)$ (left) and $u(x=-0.7857,t=89\Delta t)$ (right) for the Allen Cahn equation obtained from four different models: DPA-WNO (Proposed), Known physics (Partial physics), data-based WNO (Data only), and the ground truth (Exact). The results correspond to the case where only diffusion term is present in the known physics}
  \label{fig:ACmcub_pdfs}
\end{figure}

\subsubsection{Missing diffusion in known physics} Next, we consider the case with missing diffusion in the known physics. The known physics in this case is represented as
\begin{equation}
  \begin{aligned}
 &\frac{\partial u}{\partial t} + 5u^3 - 5u = 0, \quad x \in (-1,1), t \in (0,T)
  \end{aligned}
\end{equation}
where other boundary and initial conditions are same as described above Eq.\ref{eq:AC_comp}. For this case, we have considered $\Gamma = 0.2$. The model is trained with 50 time steps with one step size of $\Delta t = 0.0003$ sec. During the training phase, a set of $N = 32$ sample solutions were generated, considering various initial conditions of the form $\bm{u}_0(x) = \alpha \cos(0.5\zeta \pi x)$. The parameter $\alpha$ was chosen from $\left\{-8,-7,-5,\ldots,-1,1,2,\ldots,8\right\}$, and $\zeta$ took the values of $\left\{3,5\right\}$.

In the testing phase, 100 initial conditions were selected for evaluation. These initial conditions followed the same form, $\bm{u}_0(x) = \alpha \cos(0.5\zeta \pi x)$, but with the parameter $\alpha$ randomly sampled from a uniform distribution $\mathcal{U}(-12,12)$, while $\zeta$ remained constant and had values $\left\{3,5\right\}$.
The remaining configurations are kept same as in the previous case of this example.

\begin{figure}[ht!]
\captionsetup[subfigure]{labelformat=empty}
  \centering
  \begin{tabular}{ccc}
    \begin{subfigure}{0.32\textwidth}
      \includegraphics[width=\linewidth]{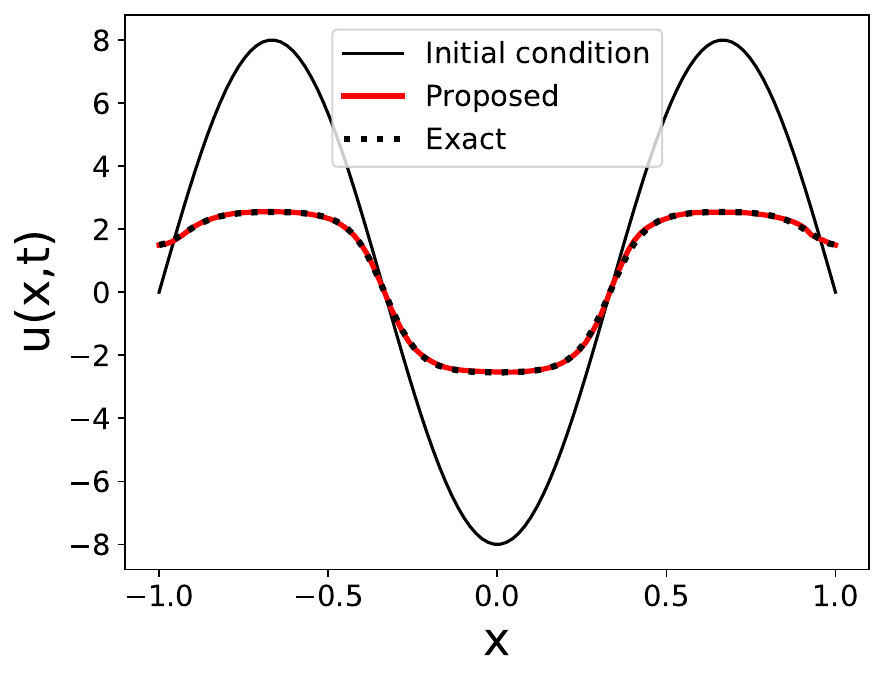}
      \caption{}
    \end{subfigure} &
    \begin{subfigure}{0.32\textwidth}
      \includegraphics[width=\linewidth]{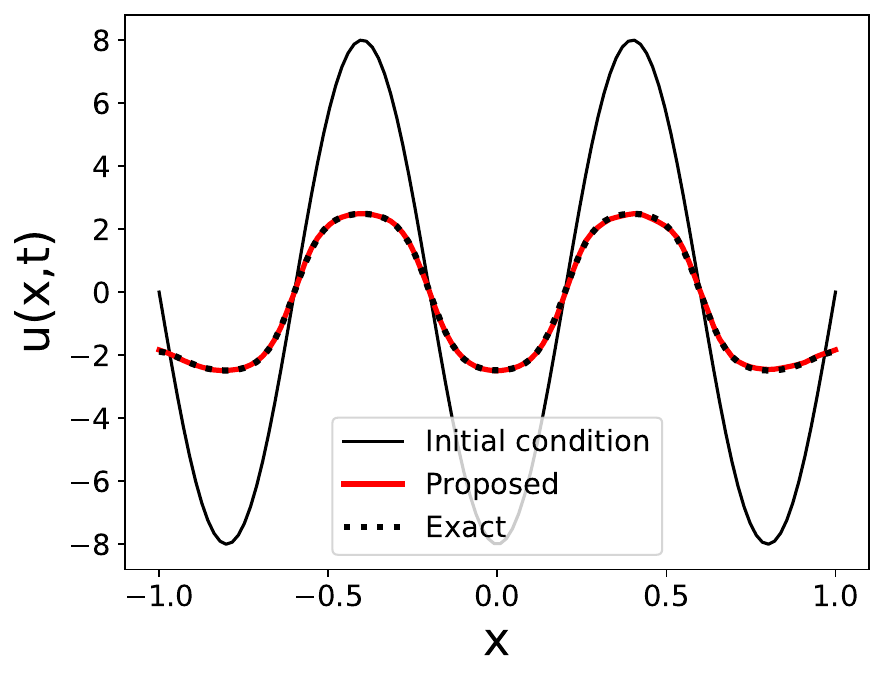}
      \caption{}
    \end{subfigure} &
    \begin{subfigure}{0.32\textwidth}
      \includegraphics[width=\linewidth]{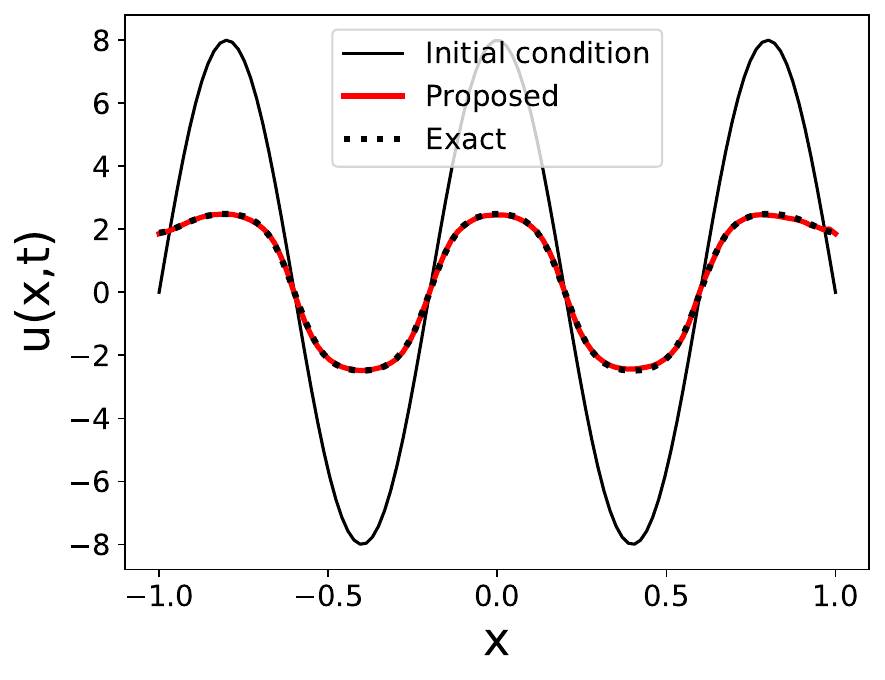}
      \caption{}
    \end{subfigure} \\
    
    \begin{subfigure}{0.32\textwidth}
      \includegraphics[width=\linewidth]{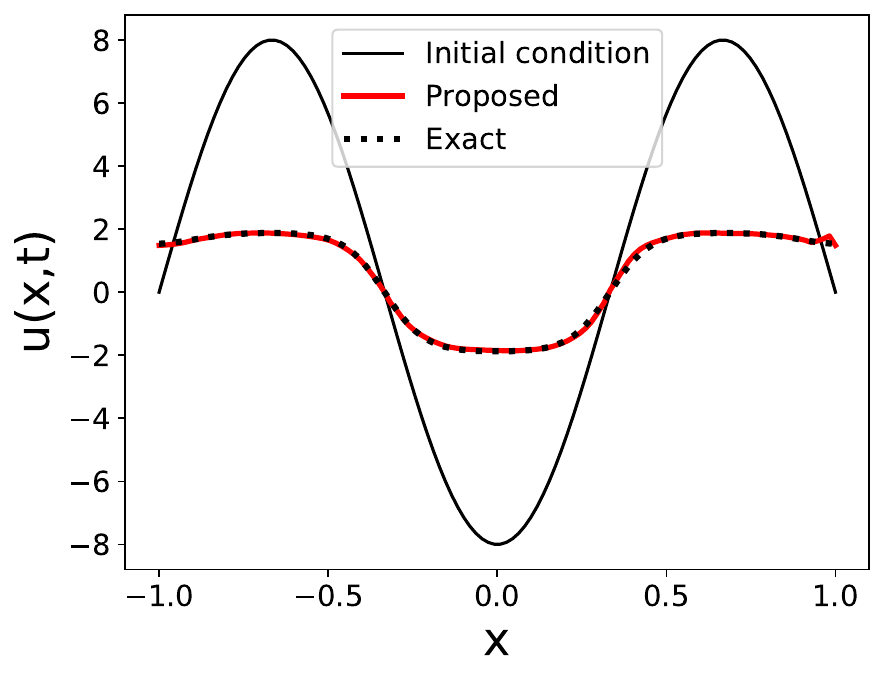}
      \caption{}
    \end{subfigure} &
    \begin{subfigure}{0.32\textwidth}
      \includegraphics[width=\linewidth]{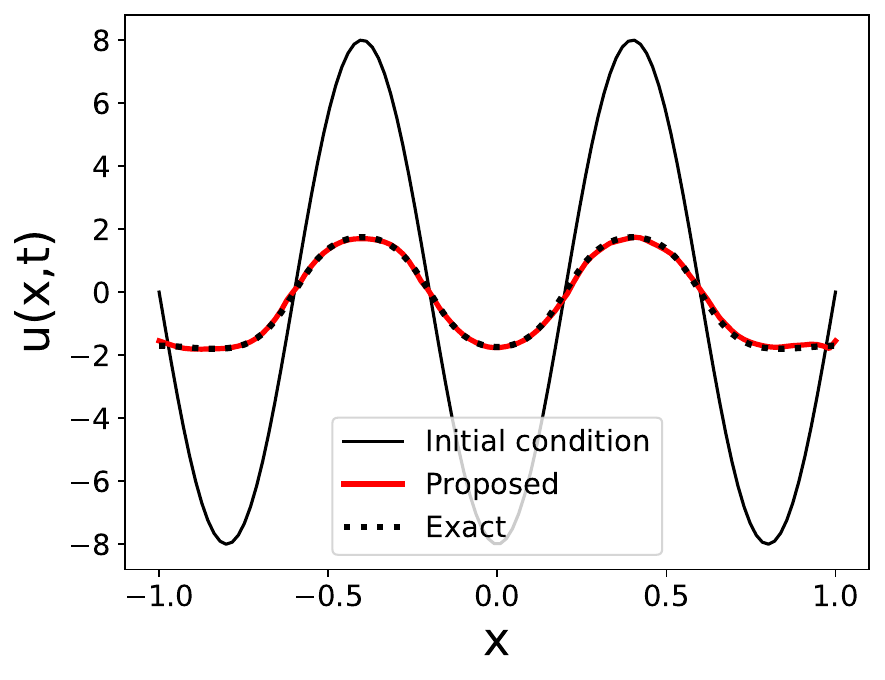}
      \caption{}
    \end{subfigure} &
    \begin{subfigure}{0.32\textwidth}
      \includegraphics[width=\linewidth]{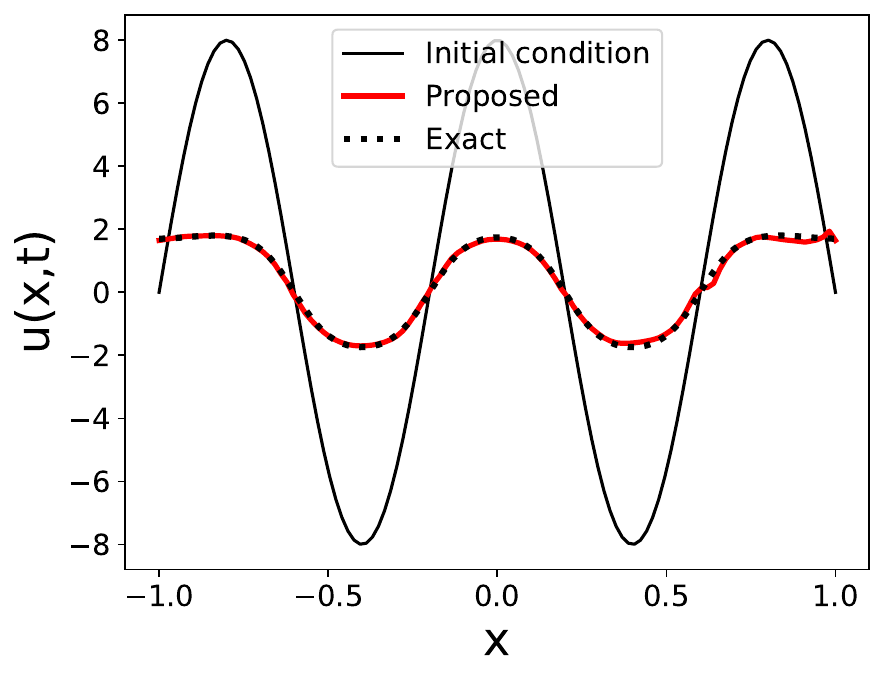}
      \caption{}
    \end{subfigure}
  \end{tabular}
  \caption{Comparison of predicted output $u(x,t)$ obtained using DPA-WNO, with the ground truth for Allen-Cahn equation (with missing diffusion term ). The comparison is shown for three different initial conditions taken from the trained samples. The top row corresponds to predictions within the training window ($t=48\Delta t$) while the bottom row corresponds to predictions outside the training window ($t=101\Delta t$)}
  \label{fig:ACmdiff_pred}
\end{figure}

\begin{figure}[ht!]
\captionsetup[subfigure]{labelformat=empty}
  \centering
  \begin{tabular}{ccc}
    \begin{subfigure}{0.32\textwidth}
      \includegraphics[width=\linewidth]{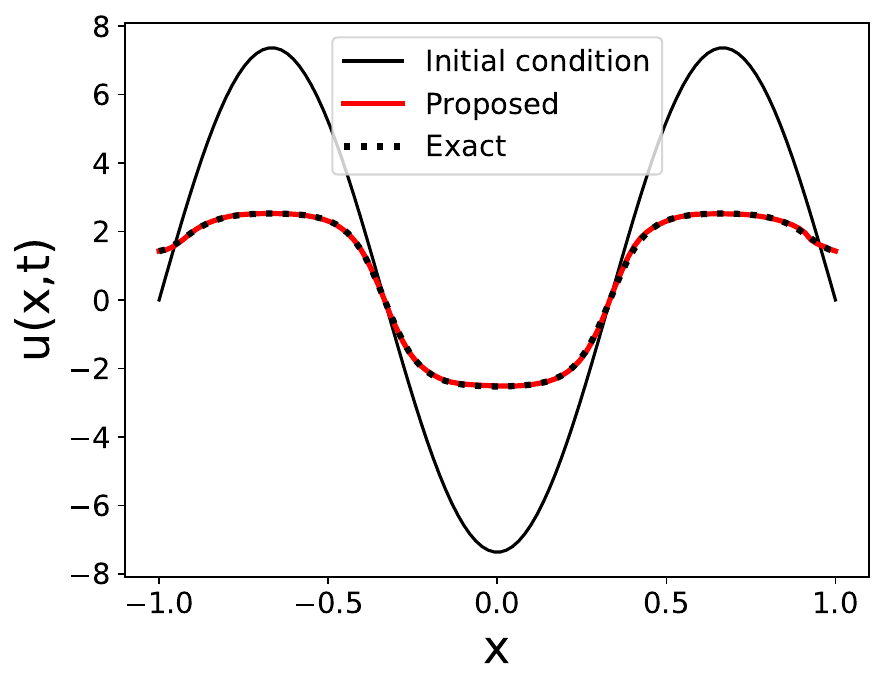}
      \caption{}
    \end{subfigure} &
    \begin{subfigure}{0.32\textwidth}
      \includegraphics[width=\linewidth]{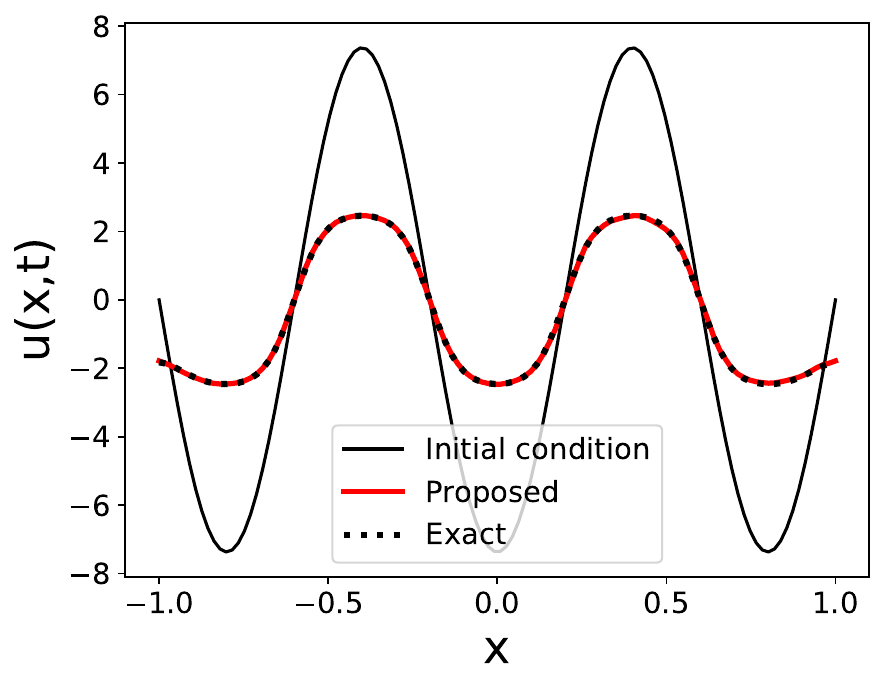}
      \caption{}
    \end{subfigure} &
    \begin{subfigure}{0.32\textwidth}
      \includegraphics[width=\linewidth]{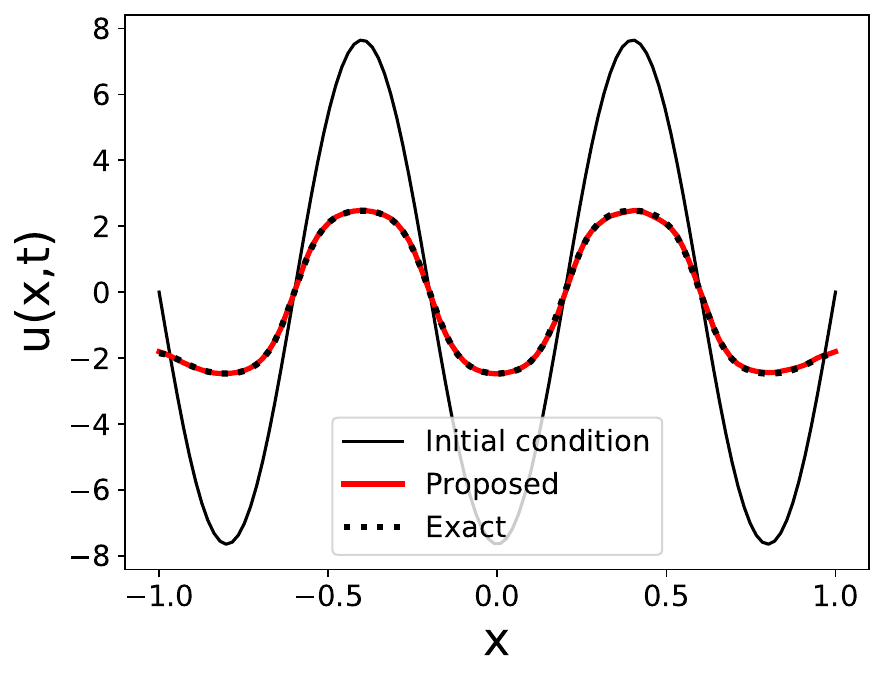}
      \caption{}
    \end{subfigure} \\
    
    \begin{subfigure}{0.32\textwidth}
      \includegraphics[width=\linewidth]{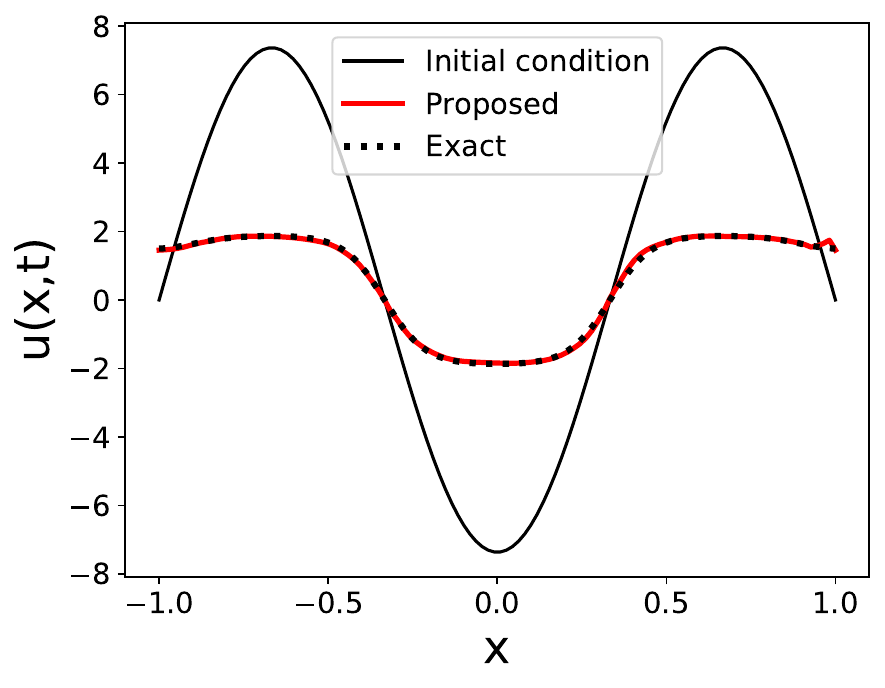}
      \caption{}
    \end{subfigure} &
    \begin{subfigure}{0.32\textwidth}
      \includegraphics[width=\linewidth]{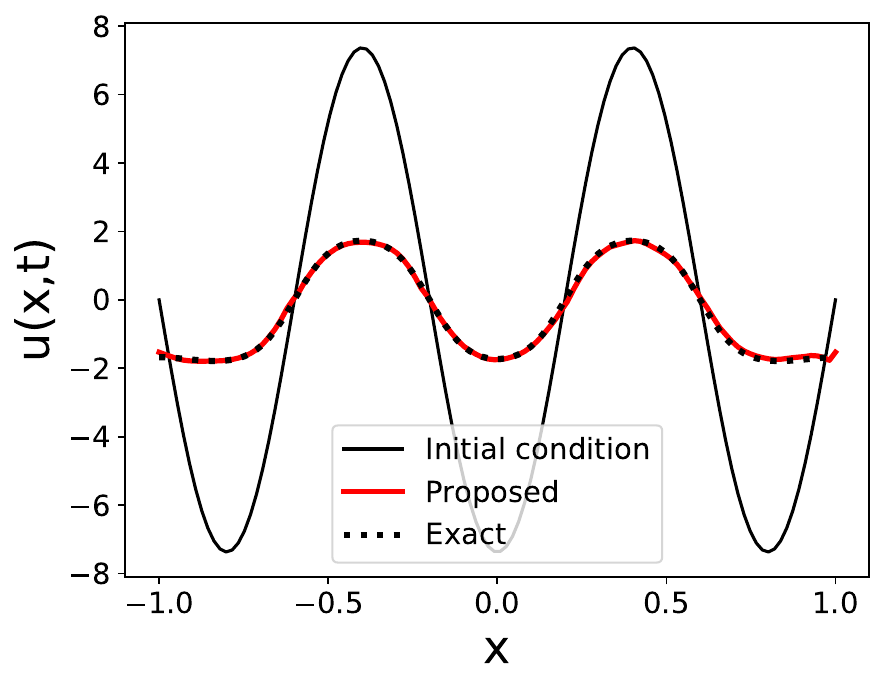}
      \caption{}
    \end{subfigure} &
    \begin{subfigure}{0.32\textwidth}
      \includegraphics[width=\linewidth]{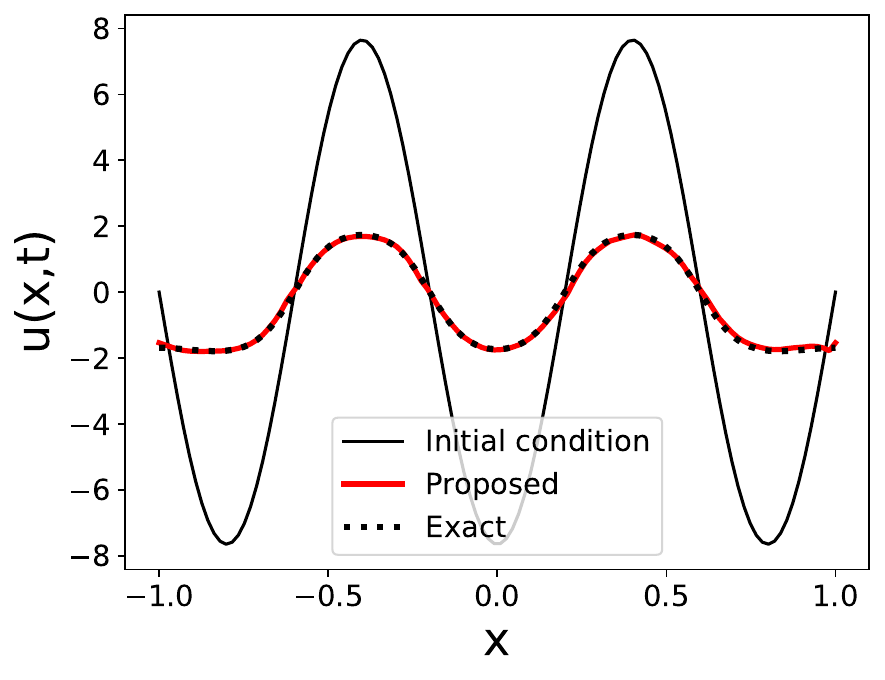}
      \caption{}
    \end{subfigure}
  \end{tabular}
  \caption{Comparison of predicted output $u(x,t)$ obtained using DPA-WNO, with the ground truth for Allen Cahn equation. The comparison is shown for three different initial conditions other than the trained samples. The top row corresponds to predictions within the training window ($t=48\Delta t$) while the bottom row corresponds to predictions outside the training window ($t=101\Delta t$). The results correspond to the case where the diffusion term is missing in the known-physics model}
  \label{fig:ACmdiff_gen}
\end{figure}

\begin{figure}[ht!]
\captionsetup[subfigure]{labelformat=empty}
  \centering
  \begin{tabular}{cc}
    \begin{subfigure}{0.48\textwidth}
      \includegraphics[width=\linewidth]{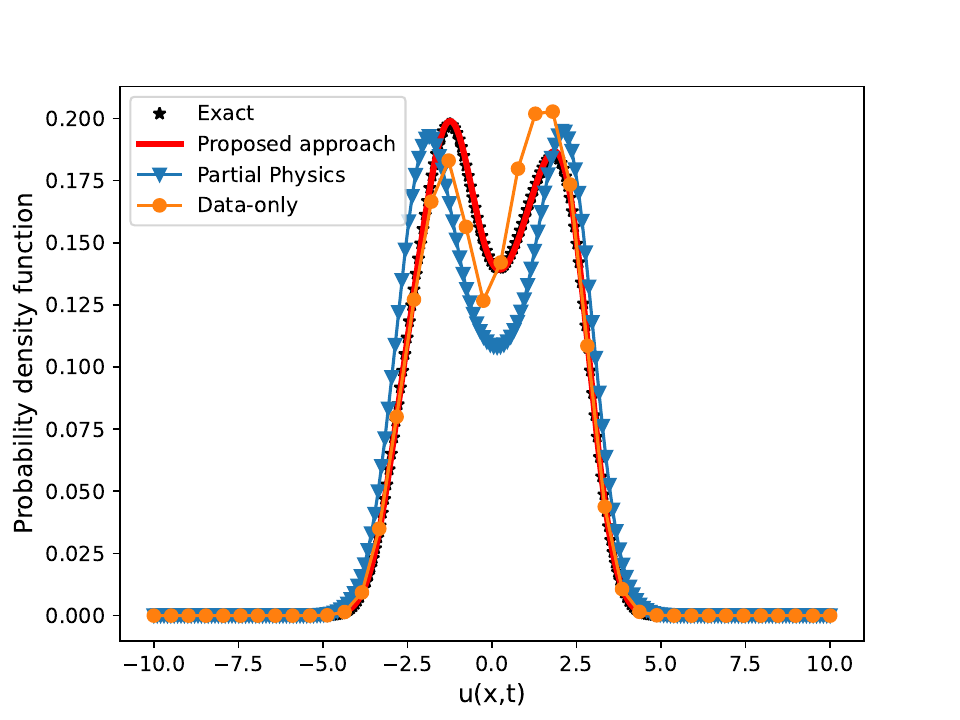}
      \caption{PDF when $t < t_{train}$}
      \label{fig:ACmdiff_pdfs_1}
    \end{subfigure} &
    \begin{subfigure}{0.48\textwidth}
      \includegraphics[width=\linewidth]{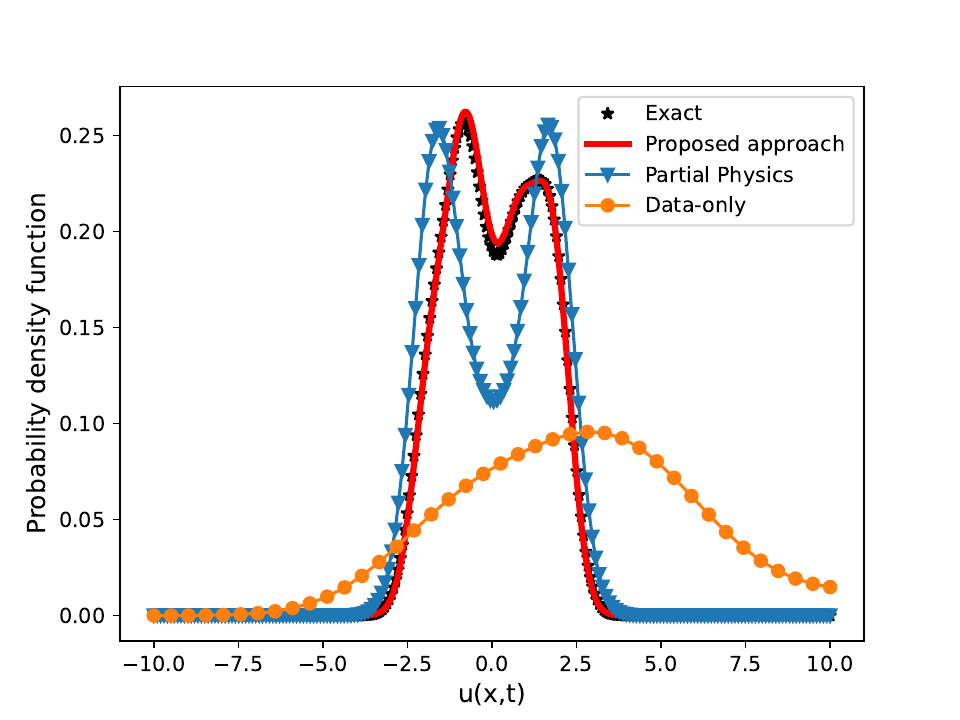}
      \caption{PDF when $t > t_{train}$}
      \label{fig:ACmdiff_pdfs_2}
    \end{subfigure}
  \end{tabular}
  \caption{Comparison of the probability density functions (PDFs) of the output $u(x=-0.8303,t=48\Delta t)$ (left) and $u(x=-0.8303,t=94\Delta t)$ (right) for Allen Cahn equation obtained from four different models: DPA-WNO (Proposed), Known physics (Partial physics), data-based WNO (Data only), and the ground truth (Exact). The results correspond to the case where the diffusion term is missing in the known-physics.}
  \label{fig:ACmdiff_pdfs}
\end{figure}

\begin{table}[t]
\centering
\begin{tabular}{|c|c|c|c|c|c|c|c|}
\hline
\multirow{2}{*}{\textbf{Method}} & \multirow{2}{*}{\textbf{Cases}} & \multicolumn{2}{c|}{\textbf{DPA-WNO}} &\multicolumn{2}{c|}{\textbf{WNO only}} & \multicolumn{2}{c|}{\textbf{Known physics}} \\
\cline{3-8}
& & \textbf{Er-1} & \textbf{Er-2} &\textbf{Er-1} & \textbf{Er-2} & \textbf{Er-1} & \textbf{Er-2} \\
\hline
\multirow{2}{*}{\textbf{Burgers}} & Only Diffusion & 0.4624 & 0.0659 & 11.0227 & 0.2867 & 2.916 & 0.5968\\
\cline{2-8}
& Missing Diffusion & 0.118 & 0.0141 & 15.667 & 0.5863 & 0.505 & 0.06855\\
\hline
\multirow{2}{*}{\textbf{Nagumo}} & Only Diffusion & 0.1873 & 0.056 & 35.685 & 0.5329 & 1.1312 & 0.2628 \\
\cline{2-8}
& Missing Diffusion & 0.0042 & 0.00105 & 34.2061 & 0.7299 & 0.2310 & 0.1411 \\
\hline
\multirow{2}{*}{\textbf{Allen Cahn}} & Only Diffusion & 0.1583 & 0.1994 & 72.033 & 1.4339 & 2.6089 & 1.3742 \\
\cline{2-8}
& Missing Diffusion & {0.0011} & 0.0007 & {28.6927} & 0.6894 & {0.1046} & 0.1763 \\
\hline
\multirow{1}{*}{\textbf{Burgers' 2D}} & Only Diffusion &  0.0010 & 0.023 & 0.1148 & 3.4394 & 0.2258 & 4.7962\\
\cline{2-8}
\hline
\end{tabular}
\caption{The prediction errors (Er-1: Mean square error (MSE) and Er-2: Mean Hellinger distance) for the different examples obtained using DPA-WNO (proposed approach), WNO only (data-driven WNO) and known physics. MSE is computed by taking all 100 test samples up to 100 time steps in all example cases. The Hellinger distance is computed by averaging over the first 100 time steps of all the 100 samples, with $x$ fixed as specified in the PDF plots of each example.}
\label{tab:MSEHD}
\end{table}

Fig. \ref{fig:ACmdiff_pred} shows the response obtained using the proposed approach. We observe that the proposed DPA-WNO yield accurate results even beyond the training window, highlighting the model's extrapolative ability. Fig. \ref{fig:ACmdiff_gen} demonstrates precise predictions for diverse initial conditions not included in the training data, indicating strong generalization capability. Fig.\ref{fig:ACmdiff_pdfs} illustrates the capability of the proposed approach in solving uncertainty propagation problems. We observe that the DPA-WNO model aligns well with the ground truth, while the known physics model deviates due to missing terms. The purely data-driven WNO model exhibits very limited accuracy in the region way outside the training window. The proposed model yield a low MSE and mean Hellinger distance, with values of $0.0011$ and $0.0007$ respectively, outperforming the purely data-driven and known physics models(Table \ref{tab:MSEHD}). These results demonstrate the excellent performance of the proposed DPA-WNO model.

For reliability analysis, we have assumed the initial condition to follow zero mean Gaussian random field with kernel function as shown in Eq. \ref{eq:kernel}. We have considered the following parameters: $\alpha = 4$, $l=0.5$, and $p=1$. The threshold for the maximum output magnitude was established at 7 units. Applying the proposed DPA-WNO method resulted in a reliability value of 99.54\%, which closely matches the ground truth reliability presented in Table \ref{tab:Reliability}

\begin{table}[h]
\centering
\begin{tabular}{|c|c|c|c|}
\hline
\multirow{2}{*}{\textbf{Method}} &{\textbf{Cases}}& \textbf{Reliability } & \textbf{Reliability} \\
&{}&{\textbf{(DPA-WNO)}} & {\textbf{(Exact)}}\\
\cline{2-4}
\hline
\multirow{2}{*}{\textbf{Burgers}} & {Only Diffusion}&{95.2 \%} & {95.72 \%}  \\
\cline{2-4}
&{Missing Diffusion} &{95.72 \%} & {95.72 \%} \\
\cline{2-4}
\hline
\multirow{2}{*}{\textbf{Nagumo}}&{Only Diffusion}  & {99.54\%} & {99.54\%} \\
\cline{2-4}
&{Missing Diffusion}& {99.40\%} & {99.40\%} \\
\cline{2-4}
\hline
\multirow{2}{*}{\textbf{Allen Cahn}} &{Only Diffusion}&{99.50 \%} & {99.50 \%}\\
\cline{2-4}
&{Missing Diffusion} & {99.54\%} & {99.54\%}\\
\cline{2-4}
\hline
\multirow{1}{*}{\textbf{Burgers-2D}}  &{Missing Diffusion}&{97.5\%} & {97.5\%} \\
\cline{2-4}
\hline
\end{tabular}

\caption{Reliability Analysis results: Investigating output solution considering 1000 samples of Initial Conditions from a Gaussian Random Field. The Limit State Function is determined by the maximum magnitude of the output, with a threshold set at 9 units for all the cases.}
\label{tab:Reliability}
\end{table}

\subsection{Example 4: 2-D Burgers' equation}\label{subsec:eg4}
As the last example, we have considered 2D Burgers' equation,
\begin{equation}\label{eq:2dcomp}
\begin{aligned}
    &\frac{\partial \mathbf{u}}{\partial t} + (\mathbf{u} \cdot \nabla)\mathbf{u} = \nu \nabla^2 \mathbf{u} \quad x \in (0,2), y \in (0,2), t \in (0,T)\\
    &u_1(x=0,y,t) = u_1(x=2,y,t) = u_1(x,y=0,t) = u_1(x,y=2,t)= 1,\\
    &u_2(x=0,y,t) = u_2(x=2,y,t) = u_2(x,y=0,t) = u_2(x,y=2,t)= 1
\end{aligned}
\end{equation}
where, $\mathbf{u} = (u_1,u_2)$ represent the velocities in $x$ and $y$ direction respectively, $t$ represents time, and $\nu=0.1/\pi$ represents the diffusion coefficient. To illustrate the applicability of the proposed approach, we have assumed that the diffusion in the $x-$direction is unknown, and hence, the known physics is of the following form:
\begin{equation}\label{eq:2dknown}
\begin{aligned}
    &\frac{\partial u_1}{\partial t} + u_1\frac{\partial u_1}{\partial x} + u_2\frac{\partial u_1}{\partial y} = 0  \\
    &\frac{\partial u_2}{\partial t} + u_1\frac{\partial u_2}{\partial x} + u_2\frac{\partial u_2}{\partial y} = \nu \left( \frac{\partial^2 v}{\partial x^2} + \frac{\partial^2 v}{\partial y^2} \right)\\
    &x \in (0,2), y \in (0,2), t \in (0,T)
\end{aligned}.
\end{equation}

\begin{figure}[p]
\captionsetup[subfigure]{labelformat=empty}
  \centering
  \begin{tabular}{ccc}
  \begin{subfigure}{0.32\textwidth}
      \includegraphics[width=\linewidth]{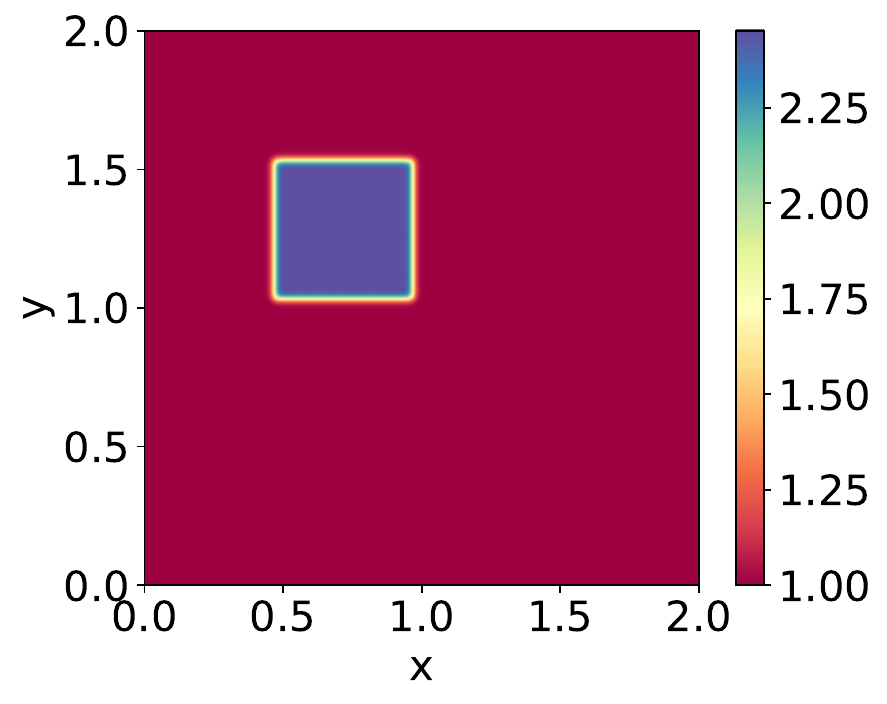}
      \caption{IC-1}
    \end{subfigure} &
    \begin{subfigure}{0.32\textwidth}
      \includegraphics[width=\linewidth]{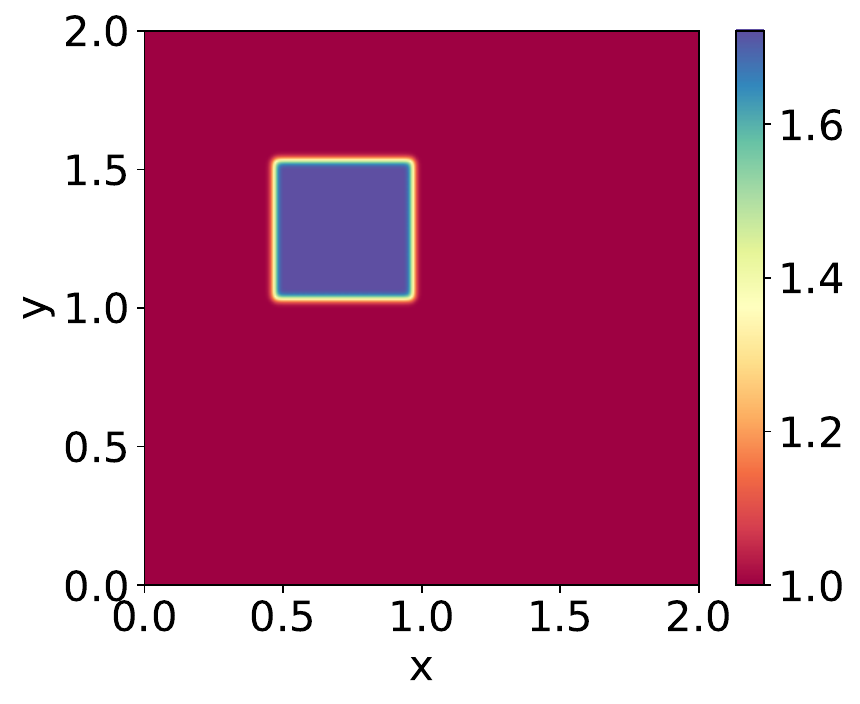}
      \caption{IC-2}
    \end{subfigure} &
    \begin{subfigure}{0.32\textwidth}
      \includegraphics[width=\linewidth]{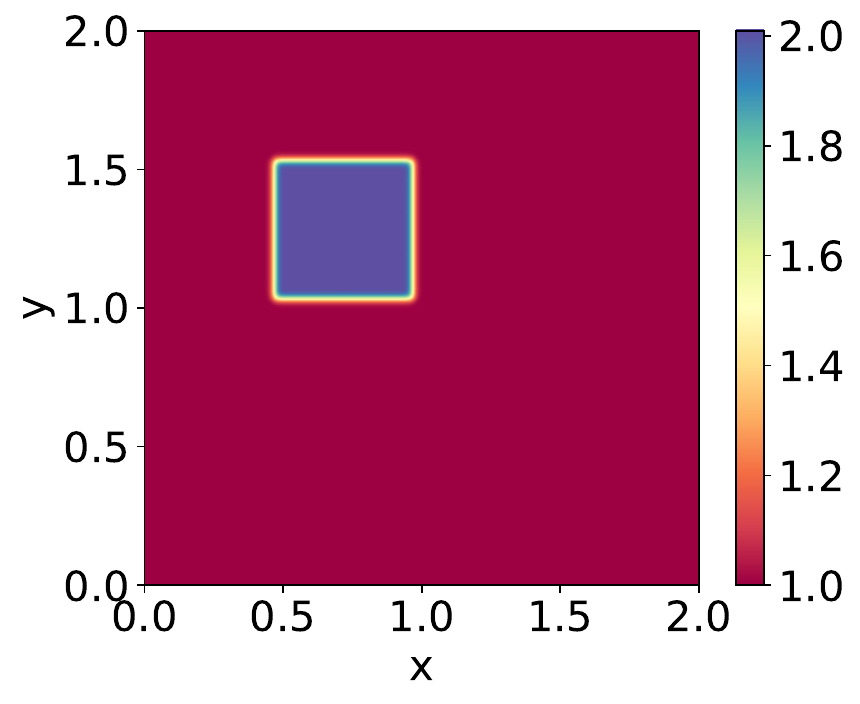}
      \caption{IC-3}
    \end{subfigure} \\

    \begin{subfigure}{0.32\textwidth}
      \includegraphics[width=\linewidth]{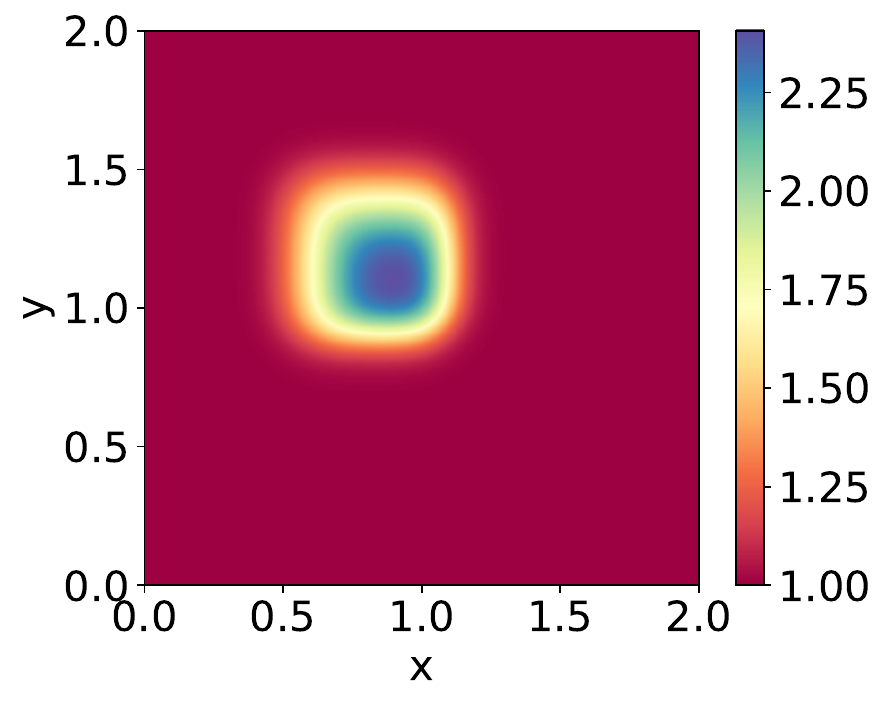}
      \caption{Exact-1}
    \end{subfigure} &
    \begin{subfigure}{0.32\textwidth}
      \includegraphics[width=\linewidth]{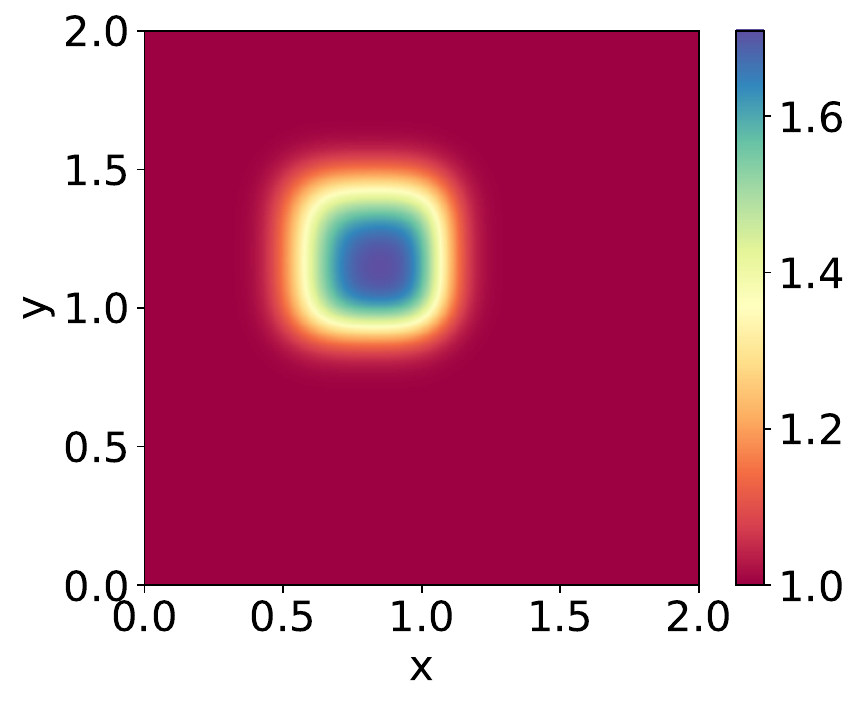}
      \caption{Exact-2}
    \end{subfigure} &
    \begin{subfigure}{0.32\textwidth}
      \includegraphics[width=\linewidth]{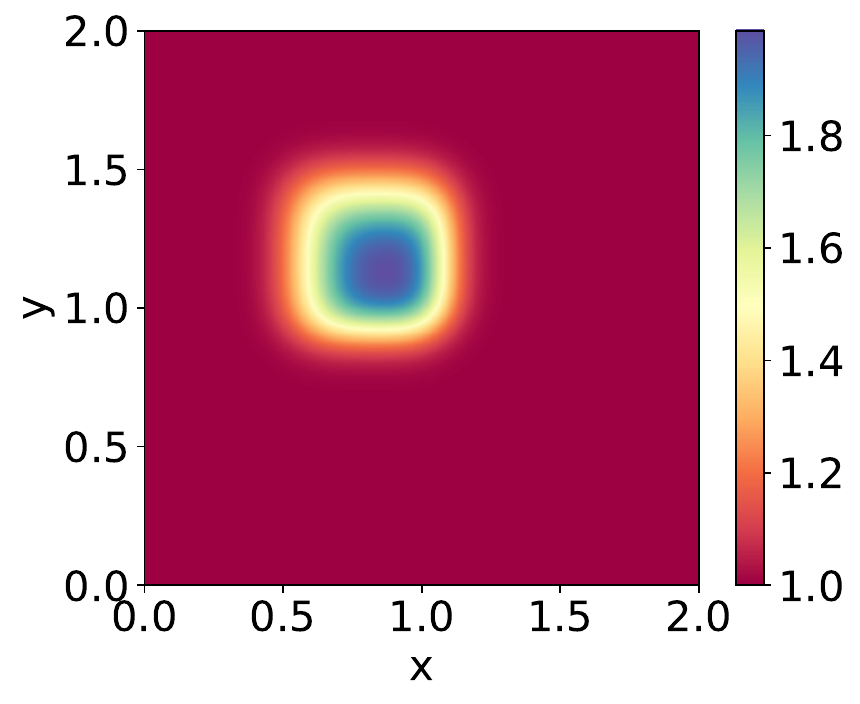}
      \caption{Exact-3}
    \end{subfigure} \\
    
    \begin{subfigure}{0.32\textwidth}
      \includegraphics[width=\linewidth]{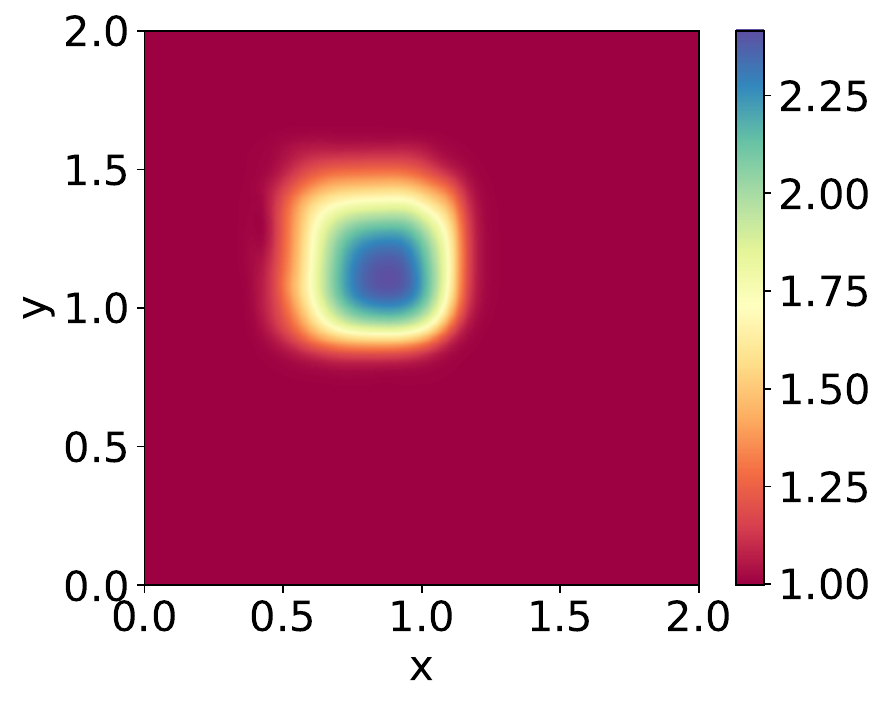}
      \caption{Predicted-1}
    \end{subfigure} &
    \begin{subfigure}{0.32\textwidth}
      \includegraphics[width=\linewidth]{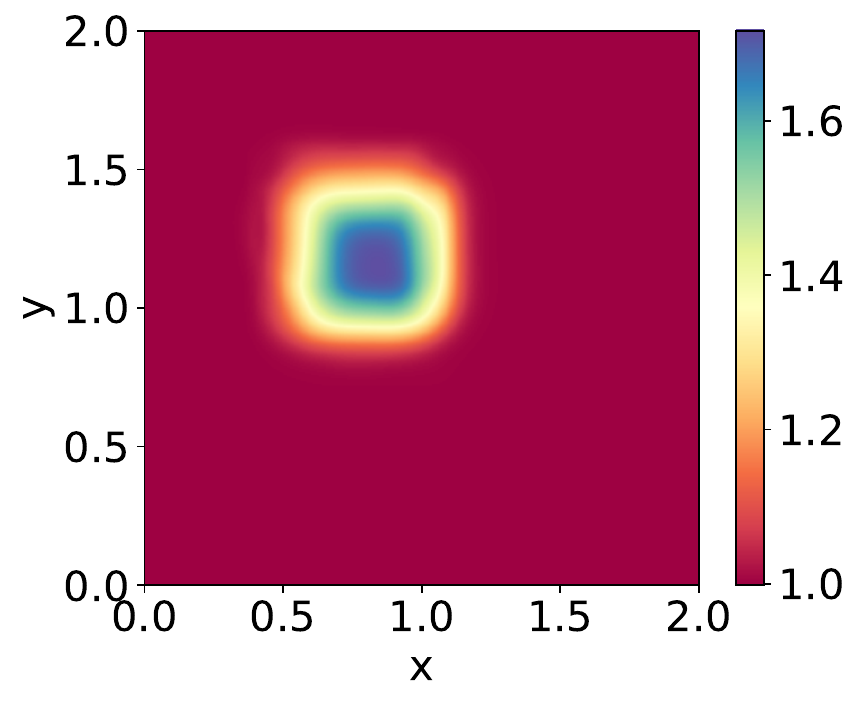}
      \caption{Predicted-2}
    \end{subfigure} &
    \begin{subfigure}{0.32\textwidth}
      \includegraphics[width=\linewidth]{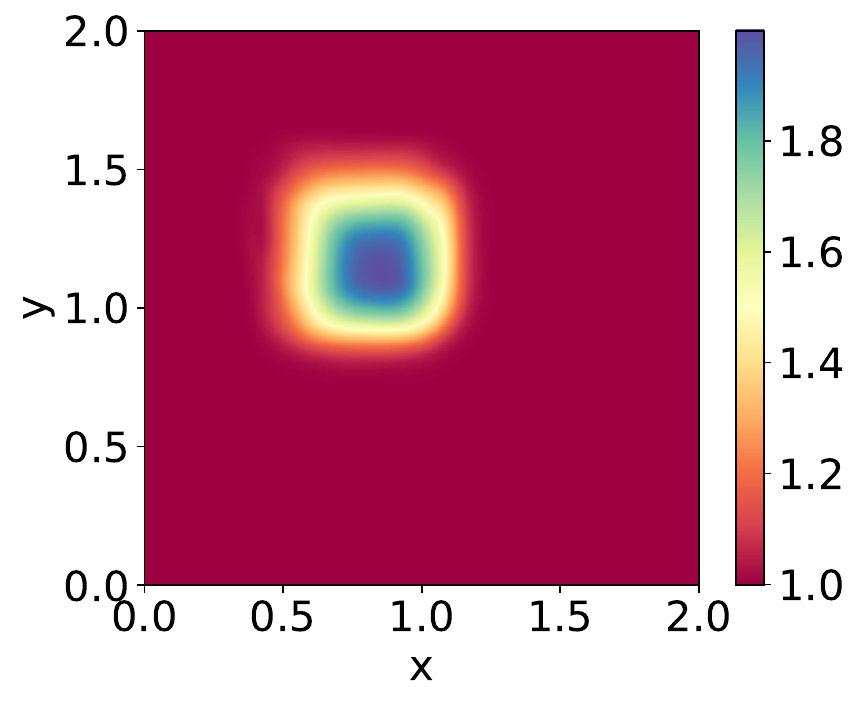}
      \caption{Predicted-3}
    \end{subfigure}\\

    \begin{subfigure}{0.32\textwidth}
      \includegraphics[width=\linewidth]{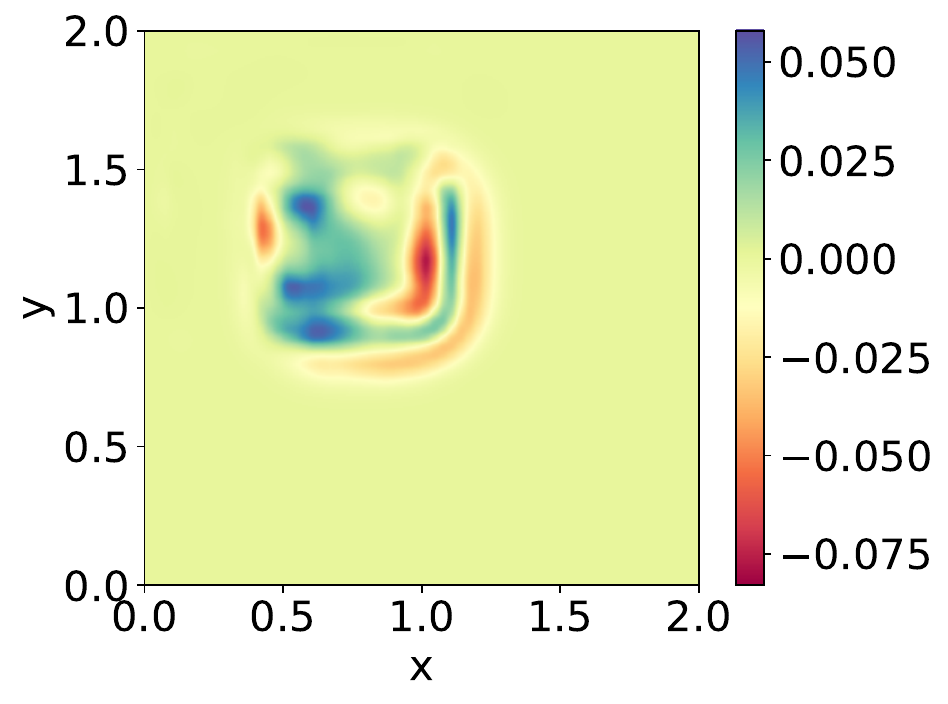}
      \caption{Error-1}
    \end{subfigure} &
    \begin{subfigure}{0.32\textwidth}
      \includegraphics[width=\linewidth]{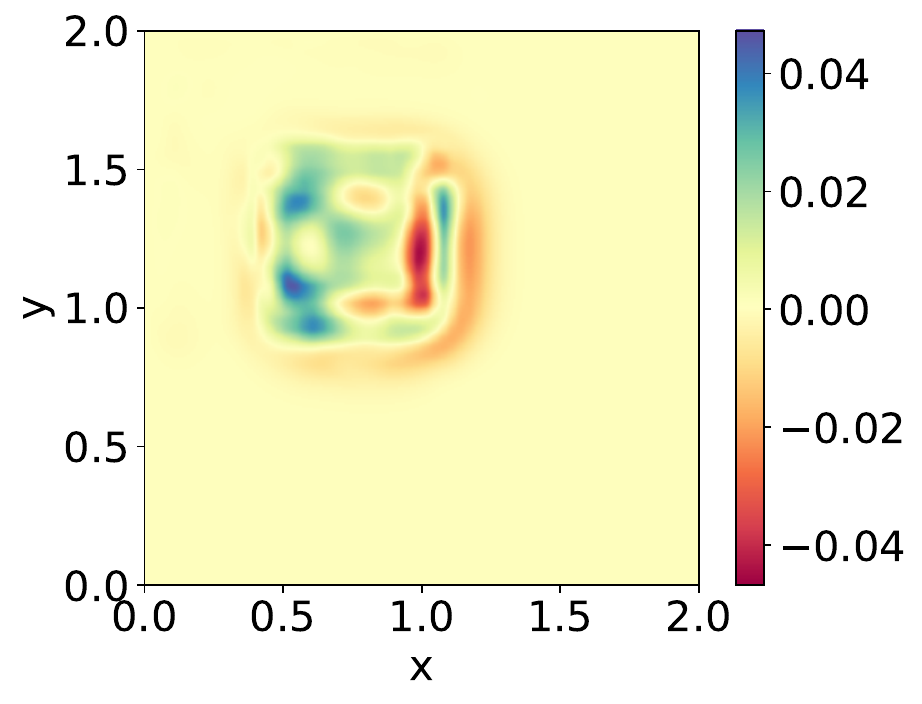}
      \caption{Error-2}
    \end{subfigure} &
    \begin{subfigure}{0.32\textwidth}
      \includegraphics[width=\linewidth]{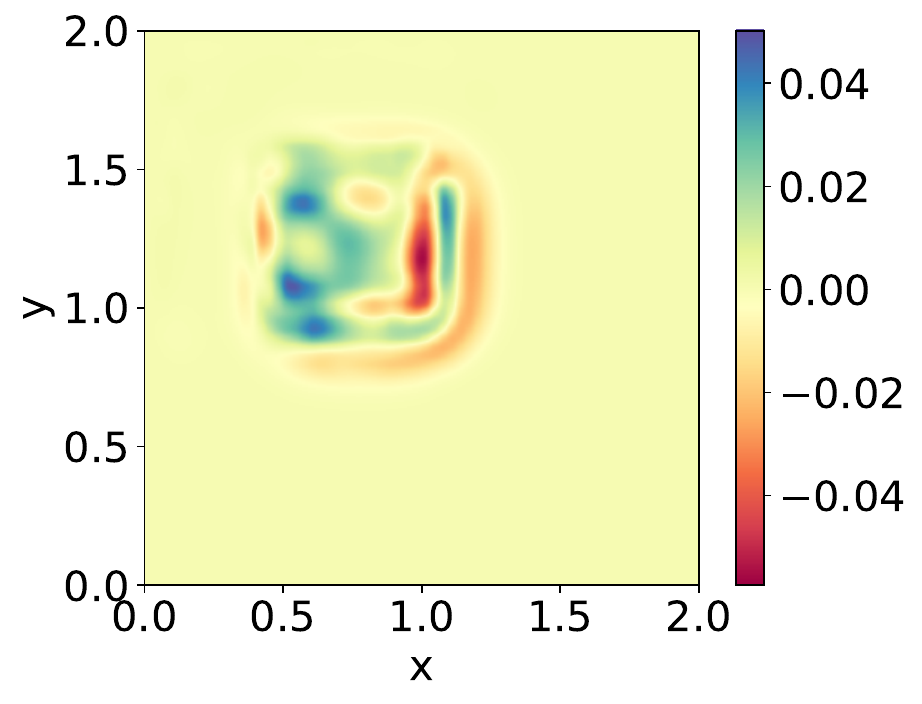}
      \caption{Error-3}
    \end{subfigure}\\
  \end{tabular}
  \caption{Comparison of predicted output $u(x,y,t=80\Delta t)$ obtained using DPA-WNO, with the ground truth for 2D Burgers equation. The comparison is shown for three different initial conditions from the training sample. This illustrates a weak generalization.}
  \label{fig:2DBurgmdiff_pred}
\end{figure}

To demonstrate the applicability of the model, we have synthetically generated the reference data $\mathcal{D} = \left\{\bm{u}_{0,1:N_x}^{(i)},\bm{u}^{(i)}_{1:N_t,1:N_x} \right\}_{i=1}^N$, by numerically solving the complete physics equation \ref{eq:2dcomp}. We have considered the square wave initial condition as defined below,
\begin{equation}\label{eq:ic2d}
  u_1(x, y, 0), u_2(x, y, 0) = \begin{cases}
u_0, & \text{if } 0.5 \leq x \leq 1.5 \text{ and } 0.5 \leq y \leq 1.5, \\
1, & \text{otherwise},
\end{cases}  
\end{equation}
The spatial domain is discretized into $64\times 64$ grid and the model is trained for $N_t = 40$ time steps, with $\Delta t =0.01$ seconds. We employ an adaptive training strategy discussed earlier in Sec. \ref{subsubsec:training}. The training utilizes the ADAM optimizer with a constant learning rate of $0.05$. A set of $N=32$ example solutions was produced, each contingent on distinct initial conditions of the form described by Eq.\ref{eq:ic2d}, with $\bm{u}_0 \sim \mathcal{U}(0,5)$. All other conditions are maintained in consistency with the original equation. With this setup, the objective is to employ the proposed approach as a surrogate model to learn the temporal evolution of the solution field and solve uncertainty quantification and reliability analysis problems by exploiting the approximate physics defined in Eq. \eqref{eq:2dknown} and the sparse data.

During the testing phase, we presented two distinct scenarios: one characterized by weak generalization and the other by strong generalization. In the case of weak generalization, we have considered 100 initial conditions with the same structure as described in Eq. \ref{eq:ic2d}. The values of the parameter $\bm{u}_0$ for these 100 instances have been randomly drawn from a uniform distribution, specifically $\bm{u}_0 \sim \mathcal{U}(0,10)$. All other conditions and configurations remain constant.
Conversely, in the context of the strong generalization scenario, we have employed three distinct types of initial conditions. Firstly, an initial condition encompassing a larger square region, where $u(x, y, 0) = v(x, y, 0) = \bm{u}_0$, and $\bm{u}_0$ values are sampled from a uniform distribution $\mathcal{U}(0,10)$ within that domain. Similarly, we have considered triangular and circular domains with $\bm{u}_0$ values drawn from the same distribution $\mathcal{U}(0,10)$. The remaining portion of the domain is maintained at a consistent value of one in all three cases, as depicted in Fig. \ref{fig:2DBurgmdiff_gen}

\begin{figure}[p]
\captionsetup[subfigure]{labelformat=empty}
  \centering
  \begin{tabular}{ccc}
  \begin{subfigure}{0.32\textwidth}
      \includegraphics[width=\linewidth]{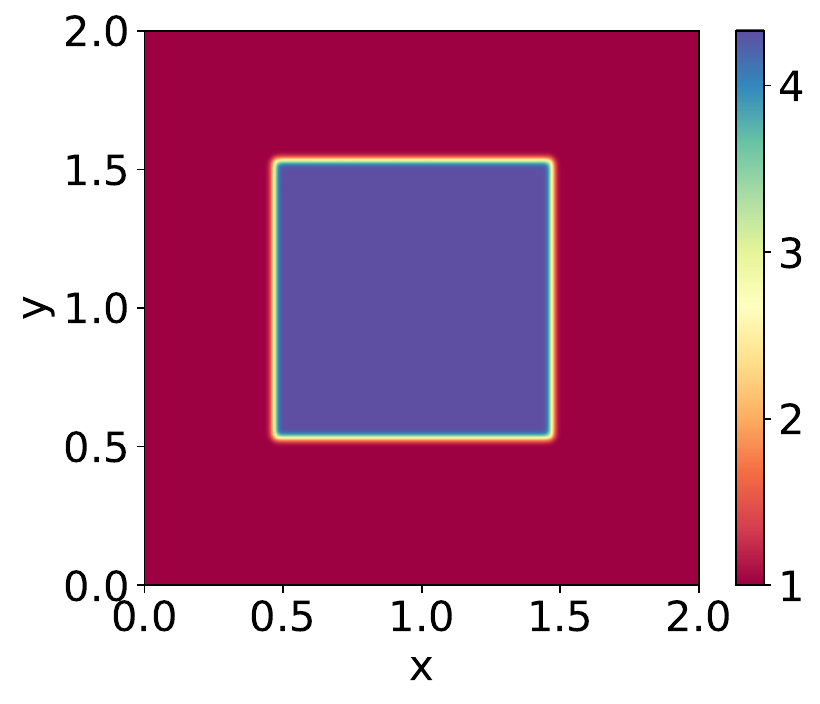}
      \caption{IC-1}
    \end{subfigure} &
    \begin{subfigure}{0.32\textwidth}
      \includegraphics[width=\linewidth]{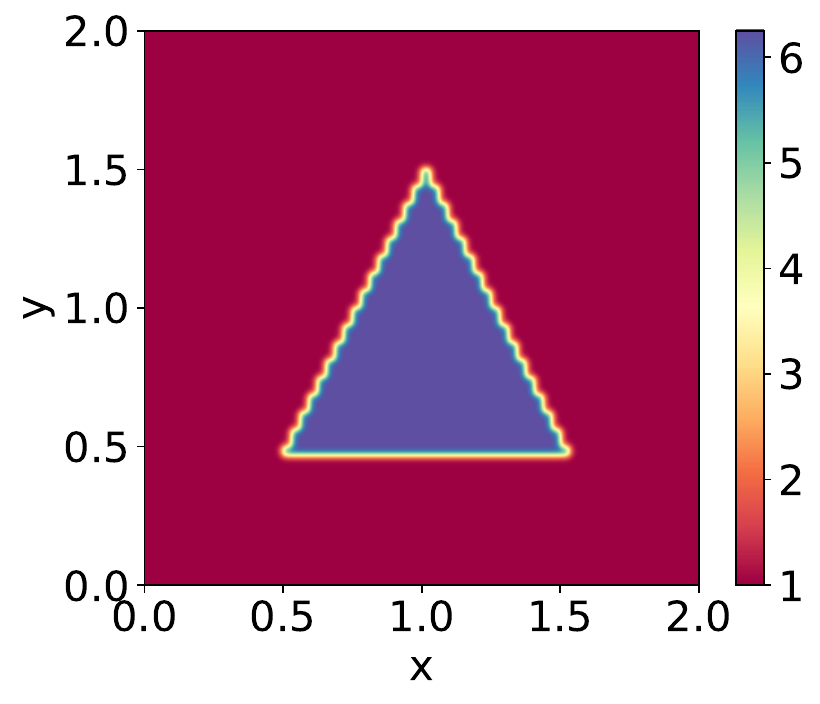}
      \caption{IC-2}
    \end{subfigure} &
    \begin{subfigure}{0.32\textwidth}
      \includegraphics[width=\linewidth]{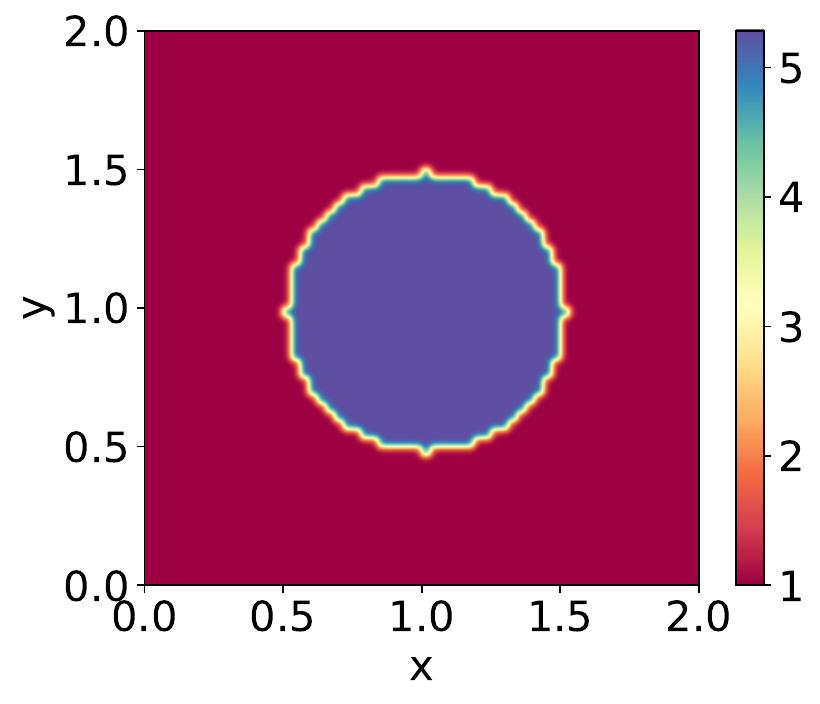}
      \caption{IC-3}
    \end{subfigure} \\
    \begin{subfigure}{0.32\textwidth}
      \includegraphics[width=\linewidth]{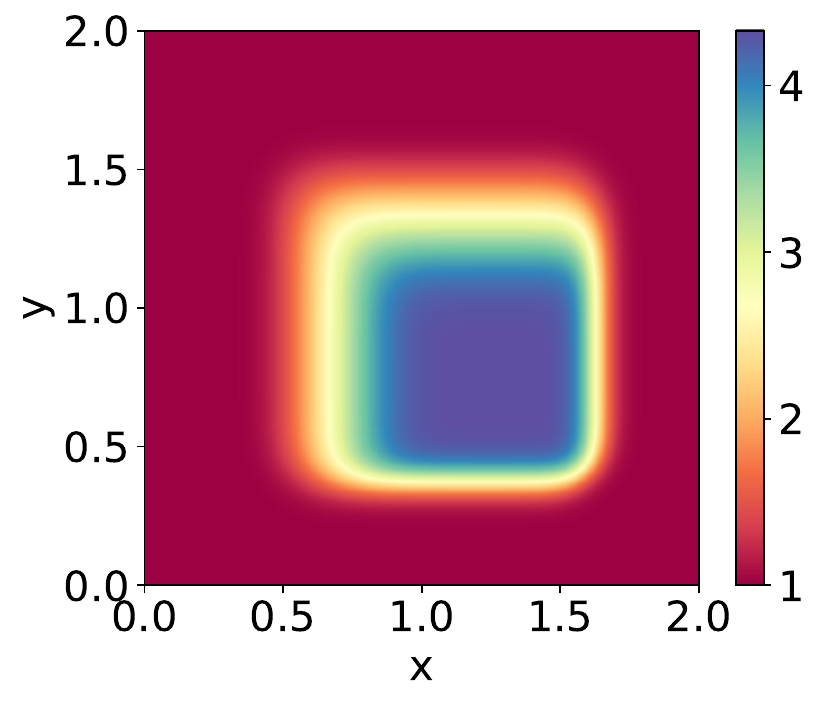}
      \caption{Exact-1}
    \end{subfigure} &
    \begin{subfigure}{0.32\textwidth}
      \includegraphics[width=\linewidth]{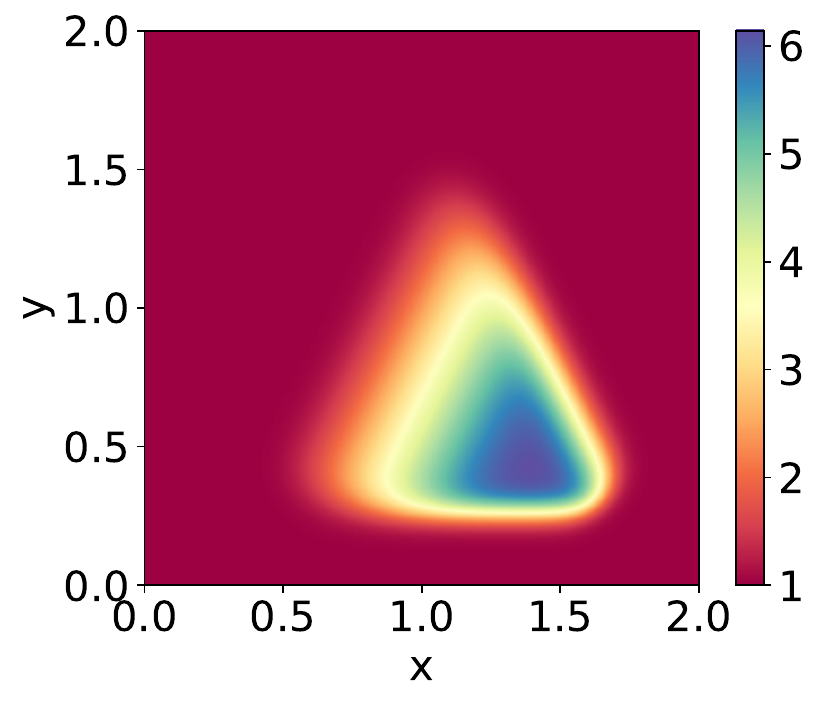}
      \caption{Exact-2}
    \end{subfigure} &
    \begin{subfigure}{0.32\textwidth}
      \includegraphics[width=\linewidth]{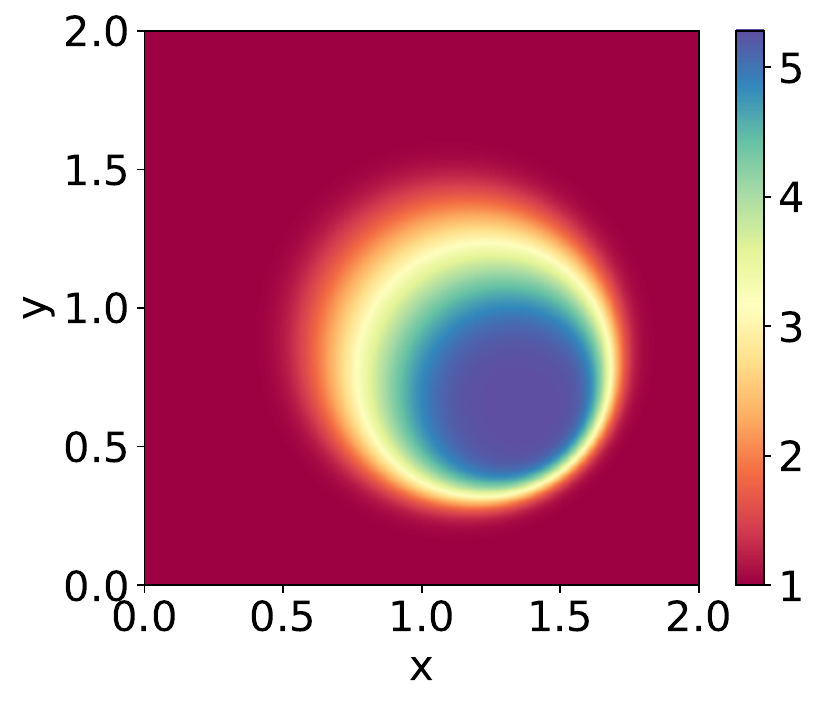}
      \caption{Exact-3}
    \end{subfigure} \\
    
    \begin{subfigure}{0.32\textwidth}
      \includegraphics[width=\linewidth]{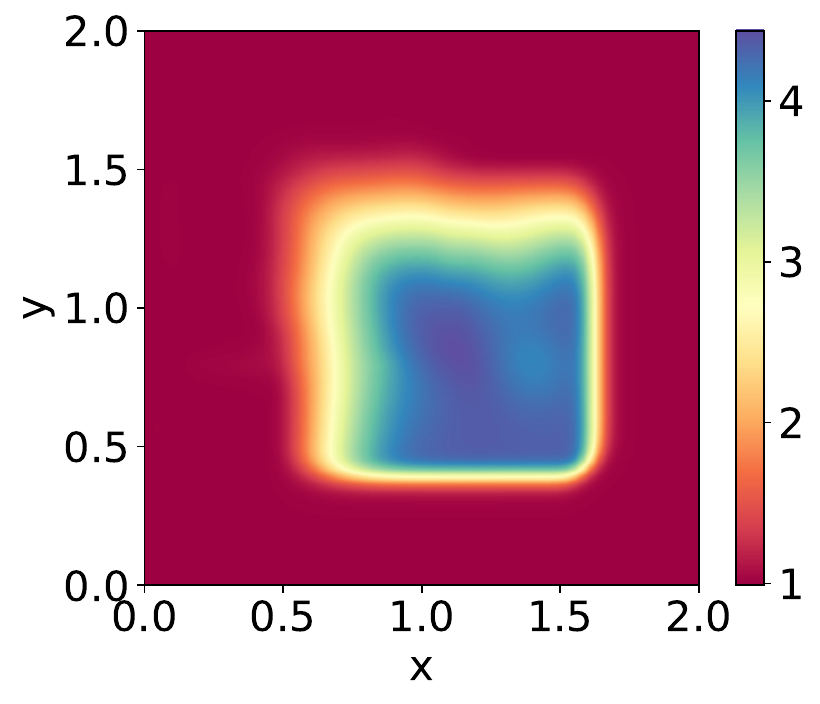}
      \caption{Predicted-1}
    \end{subfigure} &
    \begin{subfigure}{0.32\textwidth}
      \includegraphics[width=\linewidth]{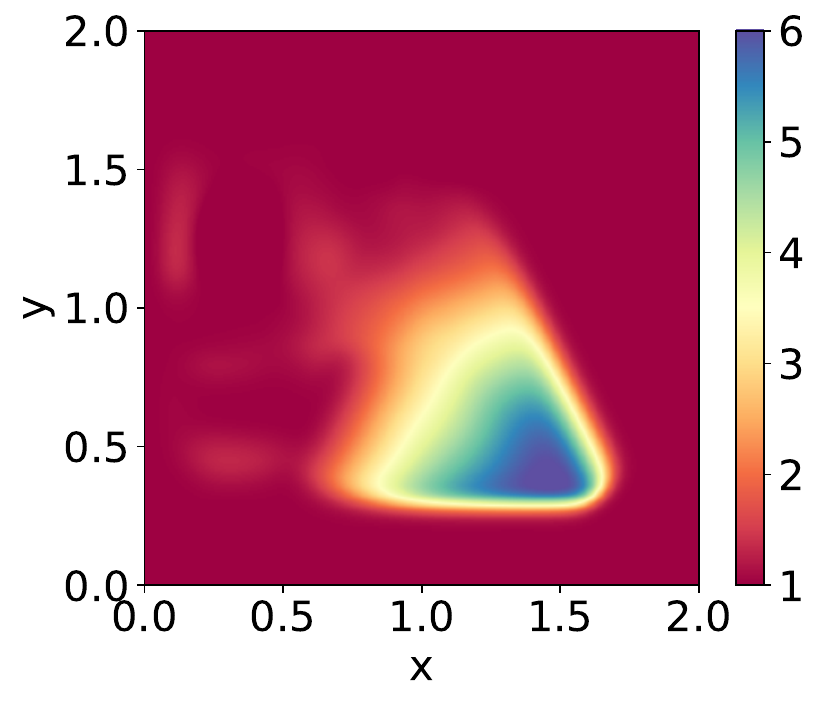}
      \caption{Predicted-2}
    \end{subfigure} &
    \begin{subfigure}{0.32\textwidth}
      \includegraphics[width=\linewidth]{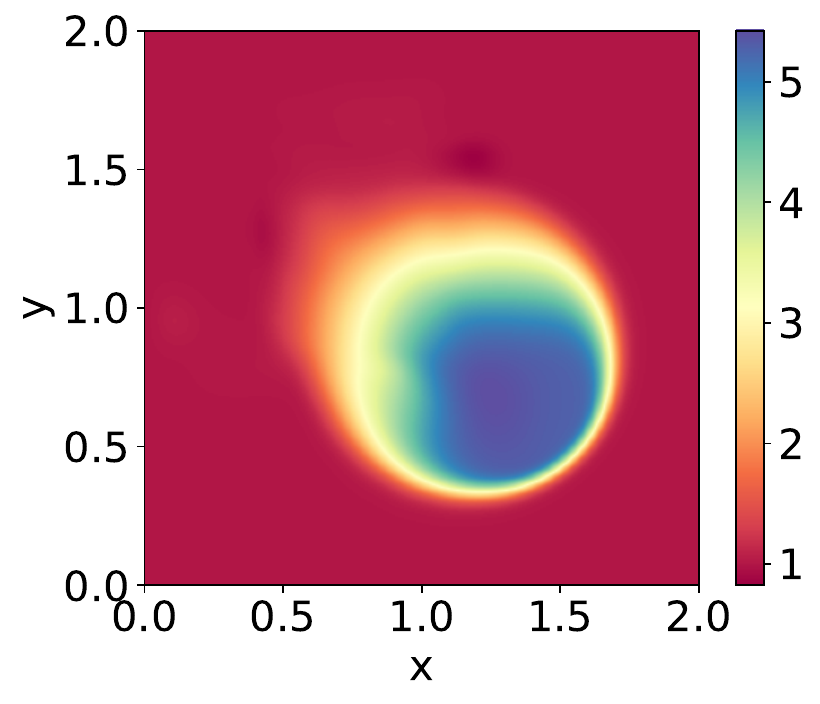}
      \caption{Predicted-3}
    \end{subfigure}\\

    \begin{subfigure}{0.32\textwidth}
      \includegraphics[width=\linewidth]{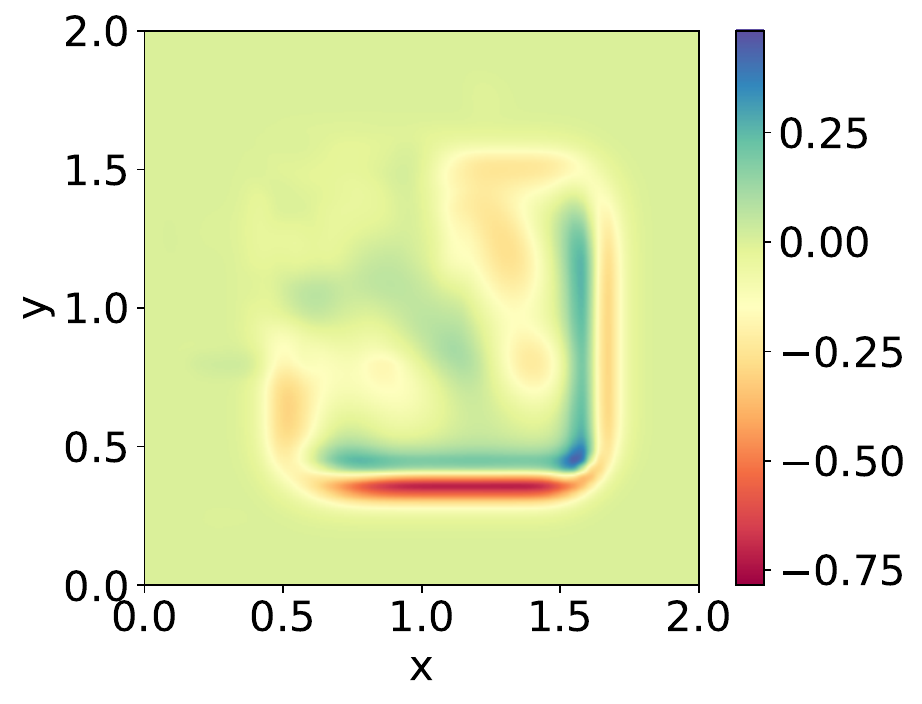}
      \caption{Error-1}
    \end{subfigure} &
    \begin{subfigure}{0.32\textwidth}
      \includegraphics[width=\linewidth]{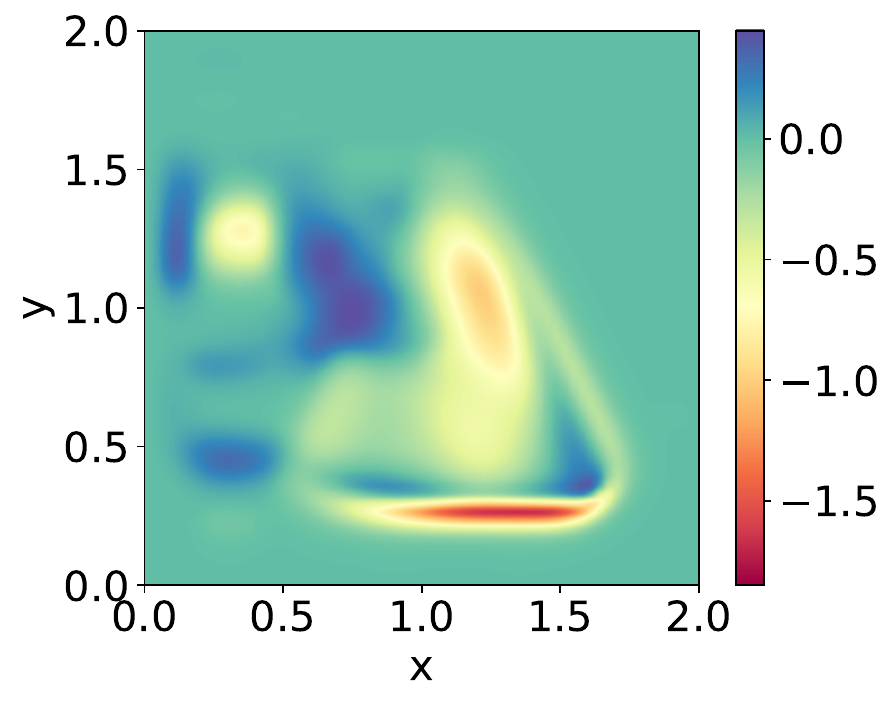}
      \caption{Error-2}
    \end{subfigure} &
    \begin{subfigure}{0.32\textwidth}
      \includegraphics[width=\linewidth]{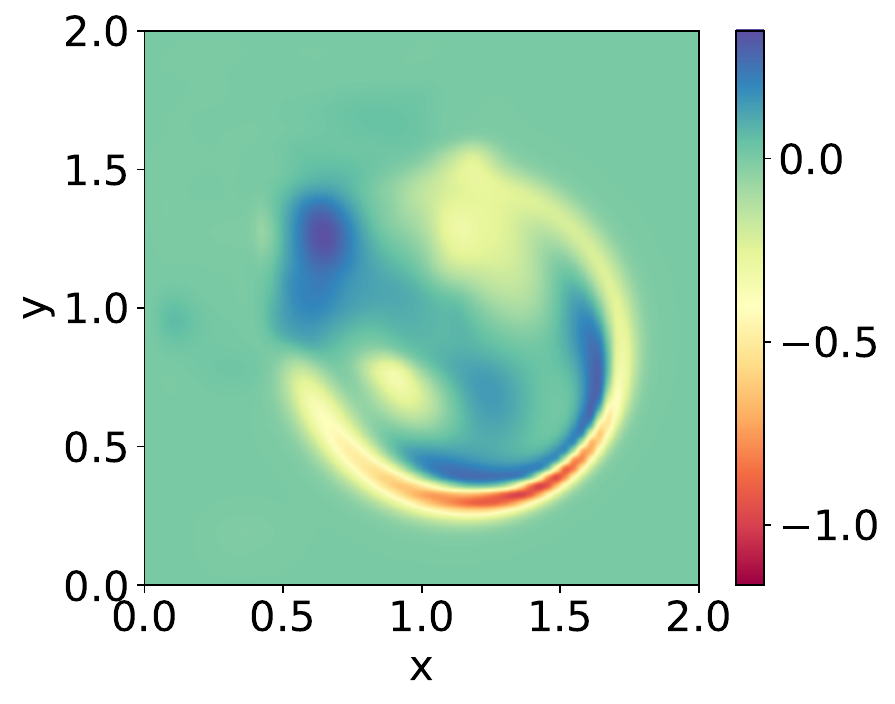}
      \caption{Error-3}
    \end{subfigure}\\
  \end{tabular}
  \caption{Comparison of predicted output $u(x, y, t=71\Delta t)$ obtained using DPA-WNO, with the ground truth for 2D Burgers equation. The comparison is shown for three different initial conditions outside the trained samples. This illustrates a strong generalization of the proposed model.}
  \label{fig:2DBurgmdiff_gen}
\end{figure}

\begin{figure}
\captionsetup[subfigure]{labelformat=empty}
  \centering
  \begin{tabular}{cc}
    \begin{subfigure}{0.48\textwidth}
      \includegraphics[width=\linewidth]{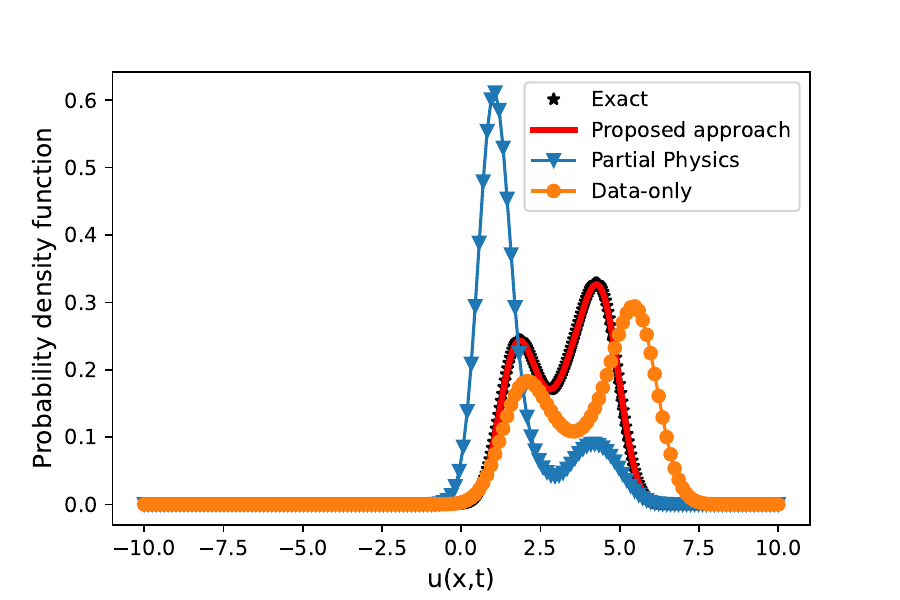}
      \caption{PDF when $t < t_{train}$}
    \end{subfigure} &
    \begin{subfigure}{0.48\textwidth}
      \includegraphics[width=\linewidth]{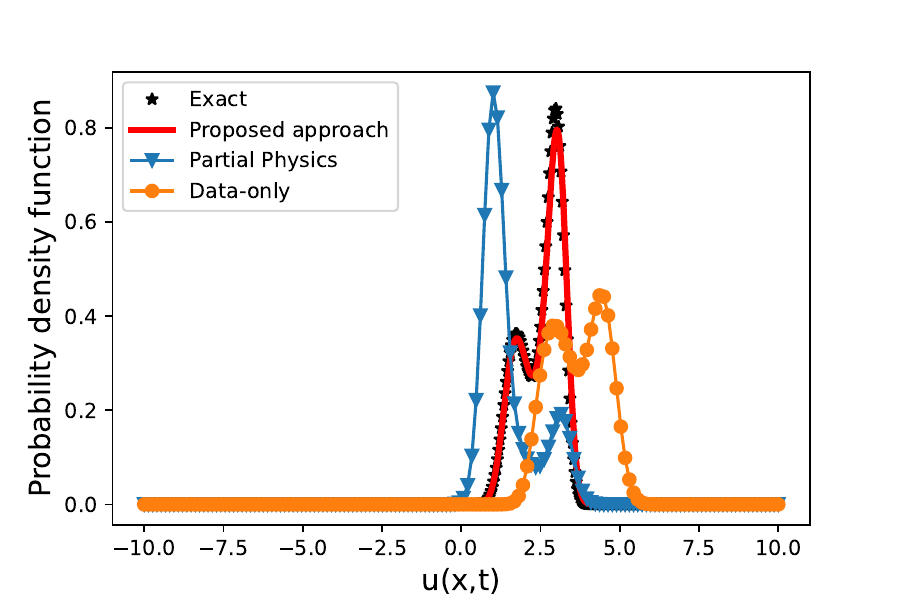}
      \caption{PDF when $t > t_{train}$}
    \end{subfigure}
  \end{tabular}
  \caption{Comparison of the probability density functions (PDFs) of the output $u(x=0.6875,y= 1.03125 ,t=40 \Delta t)$ (left) and $u(x=0.6875,y= 1.03125 ,t=74 \Delta t)$ (right) for the 2D Bargers equation obtained from four different models: DPA-WNO (Proposed), Known physics (Partial physics), data-based WNO (Data only), and the ground truth (Exact).}
  \label{fig:2DBurgers_pdfs}
\end{figure}

The results depicted in Fig. \ref{fig:2DBurgmdiff_pred} showcase precise predictions that extend beyond the training window in the case of weak generalization. Conversely, Fig. \ref{fig:2DBurgmdiff_gen} highlights robust predictions even for initial conditions outside the distribution of the training dataset, underscoring the model's ability to strongly generalize beyond the training regime scenario. The predictions shown in Fig\ref{fig:2DBurgmdiff_gen} also correspond to time outside the training time window.

In Fig. \ref{fig:2DBurgers_pdfs}, the output uncertainty depicted in the form of the response PDF is showcased.  The proposed DPA-WNO model displays strong concurrence with the ground truth, while the conventional physics-based model diverges due to omitted terms, and the purely data-driven WNO model exhibits limited accuracy, particularly beyond the training window. The DPA-WNO model achieves an exceptionally low Mean Squared Error (MSE) of $0.0010$, outperforming both the purely data-based and the physics-based models, as detailed in Table \ref{tab:MSEHD}. Moreover, it maintains a minimal mean Hellinger distance of $0.0023$, markedly surpassing the other two models, as depicted in Table \ref{tab:MSEHD}. These outcomes showcase the noteworthy prediction precision and generalization prowess of the proposed DPA-WNO model.

For the reliability analysis, we have chosen radial basis kernel function with $\alpha=4$ set at $7$ and a length scale of $0.9$. The threshold $g_t$ for the maximum output magnitude is configured at $9.1$ units. Employing the proposed DPA-WNO method yields a reliability score of 97.5\%, which accurately approximates the ground truth reliability of 97.5\%, as outlined in Table \ref{tab:Reliability}.

\section{Conclusion}\label{sec:concl}
In this paper, we have introduced a novel gray-box model referred to as the Differentiable Physics Augmented Wavelet Neural Operator (DPA-WNO) for stochastic mechanics problems. The proposed approach blends differentiable physics solver with the recently proposed wavelet neural operator. The basic premise is to allow the wavelet neural operator to learn and model-form error and incorporate it within the governing physics. We hypothesize that such a setup empowers the model to learn from data while retaining the interpretability and generalization capability of physics-based models. The proposed approach was employed to solve a class of stochastic mechanics problems where the stochasticity is due to uncertainty in the initial condition. The key observations of this paper are summarized below:
\begin{itemize}
    \item The proposed DPA-WNO demonstrates high accuracy when applied to all four benchmark example problems, carefully selected from diverse domains. The accuracy is maintained even for unseen environments, both in terms of initial conditions and temporal extrapolation. This showcases the robustness, generalization capacity, and broad applicability of the proposed method.
    \item The proposed approach works with the sparse dataset. This is evident from the fact that the model trained with only 32 training data yields accurate results. The data-driven WNO requires data in the order of a few hundred to yield satisfactory results.
    \item The proposed framework is semi-interpretable. This is due to the fact that the physics is augmented with WNO, within the form of governing PDE itself, which enhances its interpretability. 
    \item Due to the presence of WNO, the proposed approach is able to model the higher-order derivatives, just by providing functional value at the input. This is evident from all the cases, where WNO is able to model the diffusion term $\frac{{d^2 u(x,t)}}{{dx^2}}$ accurately, just with input $u(x,t)$. This makes the model independent of the order of derivative to be learned in the missing term.
    \item The proposed DPA-WNO showcases both strong and weak generalizations. This is illustrated for the 2D Burgers' equation where the model trained with square wave initial condition reasonably generalizes to triangular and circular waves. 
\end{itemize}

\section*{Acknowledgements}
SC acknowledges the financial support received from Science and Engineering Research Board (SERB) via grant no. SRG/2021/000467, Ministry of Port and Shipping via letter no. ST-14011/74/MT (356529), and seed grant received from IIT Delhi.

\section*{Code availability}
Upon acceptance, all the source codes to reproduce the results in this study will be made available to the public on GitHub by the corresponding author.

\section*{Competing interests} 
The authors declare no competing interests.

\end{document}